\documentclass[11pt]{article}

\usepackage[margin=1in]{geometry}
\usepackage{amsmath, amssymb}
\usepackage{hyperref}
\usepackage{enumitem}
\usepackage{xcolor}
\usepackage{xr}          % basic cross referencing
\usepackage{hyperref}    % optional
\usepackage{xr-hyper}    % hyperlinks for the external references
\usepackage{graphicx}
\usepackage{booktabs} 
\usepackage{soul}

\usepackage{algorithm}
\usepackage{algpseudocode}
\usepackage{tikz}
\usepackage{times}
\usepackage{latexsym}
\usepackage{amsthm}
\usepackage{amsmath}
\usepackage{amssymb}
\usepackage{multirow}
\usepackage{graphicx}
\usepackage{threeparttable}
\usepackage{algpseudocode}
\usepackage{color,soul}
\usepackage{booktabs}

\usepackage{caption}

\usepackage[labelfont=sf]{subcaption}

\newtheorem{theorem}{Theorem}
\newtheorem{definition}{Definition}
\newtheorem{lemma}{Lemma}

\newcommand{\ABSTRACT}[1]{%
  \begin{abstract}
  #1
  \end{abstract}
}

\newcommand{\KEYWORDS}[1]{%
  \vspace{0.5em}
  \noindent\textbf{Keywords:} #1
}

\usepackage{natbib}
 \bibpunct[, ]{(}{)}{,}{a}{}{,}%

\begin{document}

\title{ALARM: \underline{\textbf{A}}utomated M\underline{\textbf{L}}LM-Based \underline{\textbf{A}}nomaly Detection in Complex-Envi\underline{\textbf{R}}onment \underline{\textbf{M}}onitoring with Uncertainty Quantification}

% \author{}

% \author{
%   Congjing Zhang\textsuperscript{a}\thanks{Equal contribution.} \and
%   Feng Lin\textsuperscript{a}\footnotemark[1] \and
%   Xinyi Zhao\textsuperscript{a} \and
%   Pei Guo\textsuperscript{b} \and
%   Wei Li\textsuperscript{b} \and
%   Lin Chen\textsuperscript{b} \and 
%   Chaoyue Zhao\textsuperscript{a} \and
%   Shuai Huang\textsuperscript{a}\thanks{Corresponding author.} 
%    \\[4pt]
%   \textsuperscript{a}Department of Industrial and Systems Engineering, University of Washington
%   \textsuperscript{b}Wyze Labs, Inc.\\[4pt]
%   \texttt{\{congjing, ryanflin, xyzhao24, cyzhao, shuaih\}@uw.edu}\\
%   \texttt{\{pguo, wei.li, lchen\}@wyze.com}
% }

% \author{
%   Congjing Zhang\textsuperscript{a}\thanks{Equal contribution.} \quad
%   Feng Lin\textsuperscript{a}\footnotemark[1] \quad
%   Xinyi Zhao\textsuperscript{a} \quad
%   Pei Guo\textsuperscript{b} \\[4pt]
%   Wei Li\textsuperscript{b} \quad
%   Lin Chen\textsuperscript{b} \quad
%   Chaoyue Zhao\textsuperscript{a} \quad
%   Shuai Huang\textsuperscript{a}\thanks{Corresponding author.} \\[8pt]
%   \textsuperscript{a}Department of Industrial and Systems Engineering, University of Washington\\
%   \textsuperscript{b}Wyze Labs, Inc.\\[6pt]
%   \texttt{\{congjing, ryanflin, xyzhao24, cyzhao, shuaih\}@uw.edu}\\
%   \texttt{\{pguo, wei.li, lchen\}@wyze.com}
% }

\author{
  Congjing Zhang\textsuperscript{a}\thanks{Equal contribution.} \quad
  Feng Lin\textsuperscript{a}\footnotemark[1] \quad
  Xinyi Zhao\textsuperscript{a} \quad
  Pei Guo\textsuperscript{b} \quad
  Wei Li\textsuperscript{b} \quad
  Lin Chen\textsuperscript{b} \\[4pt]
  Chaoyue Zhao\textsuperscript{a} \quad
  Shuai Huang\textsuperscript{a}\thanks{Corresponding author.} \\[8pt]
  \textsuperscript{a}Department of Industrial and Systems Engineering, University of Washington\\
  \textsuperscript{b}Wyze Labs, Inc.\\[6pt]
  \texttt{\{congjing, ryanflin, xyzhao24, cyzhao, shuaih\}@uw.edu}\\
  \texttt{\{pguo, wei.li, lchen\}@wyze.com}
}

\date{}

\maketitle

\ABSTRACT{%
% Enter your abstract
% The reasoning capacity of LLMs can provide descriptive rationales to support a VAD model’s predictions, and offering rationales in layman’s language can also help users understand why the VAD algorithm flagged some cases as anomalies. 

The advance of Large Language Models (LLMs) has greatly stimulated research interest in developing multi-modal LLM (MLLM)-based visual anomaly detection (VAD) algorithms that can be deployed in complex environments. The challenge is that in these complex environments, the anomalies are sometimes highly contextual and also ambiguous, and thereby, uncertainty quantification (UQ) is a crucial capacity for an MLLM-based VAD system to succeed. In this paper, we introduce our UQ-supported MLLM-based VAD framework called ALARM. ALARM integrates UQ with quality-assurance techniques like reasoning chain, self-reflection, and MLLM ensemble for robust and accurate performance and is designed based on a rigorous probabilistic inference pipeline and computational process. Extensive empirical evaluations are conducted using the real-world smart-home benchmark data and wound image classification data, which shows ALARM's superior performance and its generic applicability across different domains for reliable decision-making.
}%

% \FUNDING{This research was supported by [grant number, funding agency].}

%Supplemental Material:
%Data Ethics & Reproducibility Note:

% Sample
%\KEYWORDS{Stochastic programming, Decision support,Uncertainty, Disaster response, Optimization}

% Fill in data. If unknown, outcomment the field
%\HISTORY{Received: Month DD, YYYY; Accepted: Month DD, YYYY; Published Online: Month DD, YYYY}

%%%%%%%%%%%%%%%%%%%%%%%%%%%%%%%%%%%%%%%%%%%%%%%%%%%%%%%%%%%%%%%%%%%%%%

% Text of your paper here

\section{Introduction}\label{sec:Intro}
The advance of Large Language Models (LLMs) has greatly stimulated research interest in developing multi-modal LLM (MLLM)-based visual anomaly detection (VAD) algorithms. VAD identifies unexpected events from videos or images to monitor and mitigate risks, thus improving security across diverse spaces, e.g., campuses, pedestrian zones, and crowded scenes \citep{gong2019memorizing, hasan2016learning, ionescu2019object}. Existing works in VAD have covered a range of application scenarios and methodological settings, from supervised to weakly-supervised, one-class classification, and unsupervised methods \citep{li2022self, tian2021weakly, wu2021learning, lv2021learning, zaheer2022generative}. The new interest in MLLM for VAD is largely inspired by MLLMs' reasoning capacity that can provide descriptive rationales to support a VAD model's predictions. For example, LAVAD \citep{zanella2024harnessing} focuses on detecting crimes and violent behaviors in a training-free LLM paradigm by generating a textual description for each video frame and using a prompting mechanism to aggregate these descriptions. AnomalyRuler \citep{yang2024follow} focuses on pedestrian anomalies related to biking or jumping by providing a clear rationale for its detections based on the induced rules from a few examples of normal video frames. Simply speaking, unlike machine learning-based VAD models that give black-box predictions \citep{kiran2018overview,nayak2021comprehensive}, MLLMs conduct anomaly detection by combining visual understanding with extensive world knowledge, contextual reasoning, and generate natural language explanations. Moreover, it is hoped that they are readily deployed in various domains without extensive training (i.e., with zero-shot or in-context learning) and make human-like decisions and articulate their reasoning in an interpretable way, fostering trust and enabling collaborative decision-making between AI systems and human users \citep{wang2024visiongpt,abshari2024llm}.

% Despite the promising prospects of LLM-based VAD, the challenges to materialize these promises stem from the same reasons that made LLM-based VAD attractive in the first place for many applications where the anomaly is highly 

Many existing works focus on conditions where anomalies, such as crimes, are apparent and well-defined. Existing paradigms of anomaly detection usually adopt statistical frameworks \citep{pang2021deep, adhikari2024recent} and fall short of addressing ambiguities (which lead to statistical uncertainties). For example, smart-home monitoring presents inherently contextual and ambiguous situations, such as protecting vulnerable residents (e.g., young children or elderly family members) and monitoring pets or wildlife \citep{ali2023real, ren2021deep, zhu2021video}. In these cases, what is deemed anomalous in one household may be entirely normal in another. The common-sense knowledge and reasoning capabilities of MLLMs offer great potential for capturing such contextual complexities, but these models lack an intrinsic capacity for uncertainty quantification (UQ), which is essential for managing ambiguity and ensuring reliable anomaly detection.

Another research gap is the lack of empirical evaluation of anomaly detection under ambiguous and complex conditions. Most existing studies focus primarily on assessing overall prediction performance \mbox{\citep{zhu2024llms, zhang2024holmes}}, without explicitly examining how models perform when contextual ambiguities are present. Consequently, uncertainty and robustness, i.e., two essential aspects for ensuring reliable anomaly detection in real-world environments, remain largely unquantified and underexplored. Recently, \mbox{\citet{Zhao_2025_CVPR}} introduced the SmartHome-Bench benchmark and revealed the impact of ambiguities on detection performance. An MLLM-based video anomaly detection framework called the Taxonomy-Driven Reflective LLM Chain (TRLC) is developed, which shows the promising prospect of MLLMs for VAD in smart-home monitoring and also helps us recognize that UQ is a crucial capacity for any LLM-based complex-environment monitoring system to address the challenges of context ambiguity and data limitation.

Therefore, in this paper, we propose ALARM, \textbf{A}utomated M\textbf{L}LM-Based \textbf{A}nomaly Detection in Complex-Envi\textbf{R}onment \textbf{M}onitoring with UQ, which integrates a range of quality-assurance mechanisms to enable robust and interpretable decision-making in complex environments. Specifically, we develop a novel, generic UQ methodology that decomposes an LLM-based decision-making pipeline into three sequential components, i.e., Data Comprehension, Analytical Thinking, and Reflection, quantifies uncertainty at each stage, and combines them with optimal weights. Notably, the ALARM framework is broadly applicable across diverse decision-making domains, as its three-stage reasoning process aligns closely with the cognitive structure of human decision-making. The input data can take various forms beyond visual information, such as tabular, textual, or sensor-based modalities. In this paper, we focus on visual data as representative examples and therefore employ MLLMs. For non-visual modalities, the same framework can be readily applied using standard LLMs. In various contexts, ranging from smart-home monitoring \mbox{\citep{Zhao_2025_CVPR}} to healthcare diagnostics \mbox{\citep{mehandru2025er}}, risk assessment \mbox{\citep{yuan2024r}}, and autonomous system control \mbox{\citep{vyas2025autonomous}}, decisions inherently follow the same sequence of understanding data, reasoning analytically, and reflecting upon prior judgments with auxiliary information. To show ALARM's generic applicability, we evaluate it using two real-world case studies, a real-world smart-home benchmark data and wound image classification data. Extensive empirical evaluations are conducted to show ALARM's superior performance and its generic applicability across different domains for reliable decision-making.

\section{Related Work}

\subsection{Complex-Environment Monitoring}

%\subsection{Complex-Environment Monitoring}

The pre-LLM era of complex-environment monitoring mostly focuses on multi-modal sensor integration or some special settings like senior homes. The research focus is to leverage sensing and analysis techniques, integrate ambient sensors, wearable devices, and existing computer vision algorithms to detect anomalies or health-related events \citep{park2024artificial, lopes2024covis, tian2024benefits}. For example, \cite{patel2009monitoring} develop a wearable sensor-based monitoring system to quantify motor fluctuations in Parkinson’s disease patients during daily life by using accelerometers placed on multiple body segments. \cite{lopes2021contactless} propose contactless health monitoring systems by using Internet of Things (IoT)-enabled multi-sensor platforms, including thermal cameras, depth sensors, and RGB imaging, to conduct non-invasive, automated screening for symptoms such as fever and shortness of breath. These systems reduce reliance on manual checks, mitigate risk in contexts like a pandemic, and provide continuous monitoring. Most smart-home video monitoring methods primarily rely on motion detection algorithms, statistical models, or basic machine learning techniques to detect unusual behaviors or patterns \citep{ren2021deep, yahaya2021towards, markovitz2020graph}. For example, \cite{withanage2016fall} use depth cuboid similarity features with RGB-D imaging to detect falls, aiming to support in-situ assistance for fall incidents in the context of independent living for the elderly. \cite{liu2021privacy} transform fall detection into a sparse recognition problem of the signal, incorporating visual shielding for enhanced privacy protection and recognition accuracy. \cite{yang2024fasteval} propose utilizing a dilated convolutional neural network to analyze finger-tapping videos and assess bradykinesia for the remote evaluation of Parkinsonian motor symptoms. 

The emergence of LLMs brings complex-environment monitoring to a new era. Recent advancements  focused on leveraging LLMs to create more intuitive and powerful user experiences \citep{bhat2025llm}. \cite{rivkin2024aiot} introduce an autonomous LLM agent framework to dynamically control a sequence of actions in response to a user's request. The framework constructs a dynamic tree of LLM-generated prompts to decide the next action, assess the outcome of each step, and determine when to terminate the process. \cite{duan2025home} utilize LLM-Agent technology to assist with home broadband installation and maintenance and help front-line personnel quickly diagnose issues, which can reduce service processing and fault recovery times. \cite{sun2024ai} apply LLM's reasoning capabilities to understand the context of the activity sequence from users, detect complex behaviors (e.g., forgetting medication), and interact with the user to provide reminders or assistance, aiming to improve the quality of life for elderly individuals.

Across these pre-LLM and LLM-based paradigms, a critical limitation is the lack of UQ capability. Most existing systems produce deterministic outputs without indicating confidence levels or supporting selective abstention when the AI model's confidence is low, increasing the risk of false alarms or missed detections. This gap undermines trust, constrains human-in-the-loop decision-making, and compromises user safety in safety-critical home environments.

\subsection{VAD via MLLMs}
MLLMs have been extensively applied in VAD recently. For instance, Holmes-VAD \citep{zhang2024holmes} processes untrimmed video with user prompts to produce anomaly scores and explanations. CALLM \citep{ntelopoulos2024callm} integrates a 3D autoencoder and a visual language model to predict anomalies. \cite{wang2024visiongpt} propose VisionGPT that combines an open-world object detection model (Yolo-World) with an MLLM for zero-shot anomaly detection. This system is designed to assist with visual navigation by identifying potential obstacles in real-time and providing audio-delivered descriptions of any anomalies to the user. To balance performance with computational efficiency, \cite{gao2025vagu} present a training-free approach that first performs a coarse-grained localization of potential anomalies and then uses more detailed analysis to refine the temporal boundaries and provide a semantic description. \cite{xu2024customizing} integrate multi-modal prompts for MLLMs, such as images, point clouds, and videos, into a 2D image format, which enables a single model to perform anomaly detection and reasoning across different data types. To address the limitations of existing MLLMs in handling the long-range context required for VAD, \cite{lv2024video} design a three-phase training method to improve the efficiency of fine-tuning video-based MLLMs, which minimizes the need for large amounts of VAD-specific data and reduces the cost of annotating instruction-tuning data. To evaluate existing VAD methods, \cite{bharadwaj2024vane} present a benchmark as a visual question-answering challenge on both real-world anomalies from existing datasets and synthetically generated videos with subtle inconsistencies.

Beyond videos and images, anomaly detection via LLMs has also explored other data modalities. \cite{park2024enhancing} uses a collaborative network of LLM agents, each with a specialized function such as data conversion, web research, or cross-checking, to detect anomalies in financial data. To detect logical anomalies in industrial settings, LogiCode \citep{zhang2024logicode} defines a set of logical rules based on normal examples and uses LLMs to create and execute code checking for inconsistencies in images. For semantic noise and ambiguity in log events, LLMeLog \citep{he2024llmelog} develops transformer-based anomaly detection model. 

\subsection{UQ for LLMs}\label{related_work_UQ}
UQ is a critical method to enhance the reliability and trustworthiness of LLMs. Input clarification ensembling \citep{hou2023decomposing} can be applied to any pre-trained LLM by generating a set of clarifications for a given input, feeding them to the LLM, and then ensembling the resulting predictions. By analyzing the disagreement among these predictions, the framework can distinguish between uncertainty that arises from ambiguous input (aleatoric) and uncertainty that stems from the model's lack of knowledge (epistemic). BSDETECTOR \citep{chen2023quantifying} estimates a confidence score for LLM-generated output by assessing the semantic similarity of multiple responses and prompting the LLM to evaluate the correctness of its own answer. Similarly, D-UE \citep{da2024llm} models the relationships between different LLM responses using a directional entailment graph and uses a Random Walk Laplacian to quantify the resulting uncertainty. However, the traditional dichotomy of aleatoric and epistemic uncertainty is too limited for the interactive and open-ended nature of LLM agents. To fully capture the complexities of uncertainty in LLMs, the uncertainty can be extended in different dimensions, including input, reasoning, parameter, and prediction \citep{liu2025uncertainty}. Regarding the phases of the interactions with LLMs, uncertainty can also be categorized into underspecification (when user prompts are incomplete), interactive learning (where the model can ask clarifying questions), and more expressive output uncertainties that go beyond simple numerical scores \citep{kirchhof2025position}. \cite{nikitin2024kernel} propose Kernel Language Entropy, which uses semantic kernels to encode the similarity between LLM outputs and quantifies uncertainty using von Neumann entropy. In a similar vein, Inv-Entropy \citep{song2025inv} quantifies uncertainty by evaluating the diversity of the input space. It is achieved through a dual random walk model and a perturbation algorithm that generates a diverse set of input samples.

Many of the existing works are generic and aim to uncover the theoretical nature of uncertainties in LLM decision-making or provide practical methods to evaluate the confidence of the LLM outputs for general usage. Our work differs from the existing works in the sense that we provide an in-depth UQ analysis of a multi-stage MLLM-based anomaly detection pipeline and the correspondingly optimized UQ score that considers multiple aspects of costs, utility, and uncertainty through an optimization framework to better operationalize an MLLM-based complex-environment monitoring system in the face of its self-recognized uncertainties. Nonetheless, there are some recent developments that we can use as baselines to compare our ALARM with, i.e., the Least Ambiguous Set-Valued Classifiers (LAC) (i.e., computing UQ scores regarding the softmax score corresponding to the true label \mbox{\citep{ye2024benchmarking, sadinle2019least}}), the Adaptive Prediction Sets (APS) (i.e., computing UQ scores by summing the ranked scores of each label \mbox{\citep{ye2024benchmarking,romano2020classification}}), the Epistemic Uncertainty (ICL-EU) and Aleatoric Uncertainty (ICL-AU) within in-context learning (i.e., computing UQ scores following the Bayesian formulation of predictive uncertainty related to provided demonstrations and model’s configurations within the in-context learning paradigm \mbox{\citep{ling2024uncertainty}}).

There is also a line of work that is related to ALARM which focus on selective classification. For instance, \mbox{\cite{geifman2017selective}} pair trained deep neural networks with a rejection function to guarantee a desired true error rate at test time, while SelectiveNet \mbox{\citep{geifman2019selectivenet}} integrates the reject option directly into the network architecture. \mbox{\cite{franc2023optimal}} provide a unified theory of reject-option classifiers, proving that cost-based, bounded-improvement, and bounded-abstention models all share the same optimal strategy, i.e., a Bayes classifier paired with a randomized selection function driven by a proper uncertainty score. \mbox{\cite{madras2018predict}} introduce learning to defer that jointly optimizes fairness and accuracy by allowing a model to dynamically decide whether to predict. \mbox{\cite{alves2025benchmarking}} benchmark on learning to defer, further emphasizing evaluation under realistic human-cost and human-error models. \mbox{\cite{ye2024benchmarking}} benchmark LLMs via calibration quality and selective-prediction metrics using LAC/APS-style scores. However, existing work does not decompose LLM uncertainty across different stages and therefore cannot offer fine-grained interpretability regarding where uncertainty originates. In contrast, our ALARM framework computes stage-wise uncertainty via the LLM reasoning chain and performs selective classification based on a unified UQ score derived from these decomposed signals.

\section{Theoretical Framework of ALARM}\label{section3}

% \subsection{Uncertainty quantification for LLM-assisted decision-making}

\begin{figure*}[!t]
\centering
\includegraphics[width=1\textwidth]{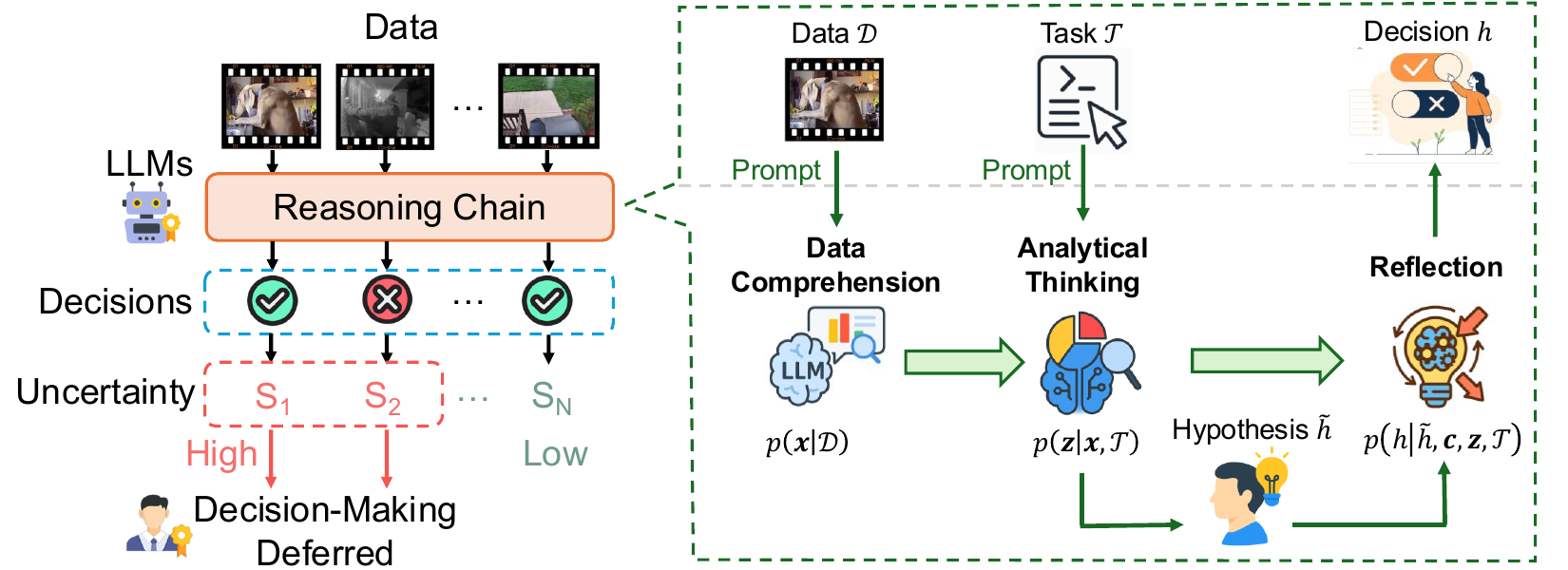}
\caption{Overview of the ALARM framework.}
\label{fig:uq}
%\vspace{-1em}
\end{figure*}

Figure \ref{fig:uq} presents an overview of ALARM and a typical MLLM-based decision-making pipeline that consists of Data Comprehension, Analytical Thinking (i.e., hypothesis generation), and Reflection. Here, hypothesis refers to a tentative decision since it is up for revision. Given a data instance, ALARM begins with understanding the data content (Data Comprehension), followed by analytical reasoning (Analytical Thinking), to generate a hypothesis about if the data instance is anomalous. In the absence of additional information, this hypothesis serves as the final justification for the decision. However, if there is side information such as additional knowledge provided by a human, a knowledge graph, or a set of rules describing common characteristics of anomalies, one can provide this side information such that the MLLM-based decision-making pipeline can reconsider the initial hypothesis and may revise the hypothesis, to reach a final decision (Reflection).

% Human decision-making under incomplete information typically follows a sequential reasoning process comprising comprehension, analytical thinking, and hypothesis generation. Upon the acquisition of side information about the task, individuals can engage in post-hoc reflective reasoning, which enables further evaluation and potential revision of initial hypotheses, thereby enhancing confidence in the resulting decisions. 

% The pipeline is illustrated in Figure \ref{fig:uq}. 

\subsection{Definition of the UQ Scores} \label{UQscore_define}

Analytically speaking, an MLLM is prompted with the task information $\mathcal{Q}=\{\mathcal{D}, \mathcal{T}\}$ where $\mathcal{D}$ is data (e.g., a video) and $\mathcal{T}$ is the task context which includes the goal and requirements of the task. The MLLM gives a comprehension of the data (Data Comprehension) and generates a description $\boldsymbol{x}$. As an MLLM is a probabilistic model, this generational process could be characterized as a probabilistic model $\boldsymbol{x}\sim p(\boldsymbol{x}|\mathcal{D})$. Given $\boldsymbol{x}$, MLLM reasons about the task and generates a detailed analysis (Analytical Thinking), denoted by $\boldsymbol{z}$. This process can be characterized as $\boldsymbol{z}\sim p(\boldsymbol{z}|\boldsymbol{x},\mathcal{T})$. Given $\boldsymbol{z}$, MLLM derives a task-related hypothesis $\tilde{h} = ext(\boldsymbol{z})$, where $ext(\cdot)$ represents an information-extraction mapping from $\boldsymbol{z}$ to $\tilde{h}$. Finally, the MLLM is provided with side information $\boldsymbol{c}$ and performs reflective reasoning (Reflection) to evaluate hypothesis $\tilde{h}$ and refine it through $h\sim p(h|\tilde{h},\boldsymbol{c},\boldsymbol{z},\mathcal{T})$. As illustrated in Figure \ref{fig:uq}, the whole pipeline builds on a reasoning chain: $\mathcal{D}\xrightarrow{} \boldsymbol{x}\xrightarrow{} \boldsymbol{z}\xrightarrow{} \tilde{h} \xrightarrow{} h$. This leads to a decomposition of the decision-making process as a probabilistic model:
\begin{align}
\begin{aligned}
p(h|\mathcal{Q})=&\int\underbrace{p(\boldsymbol{x}|\mathcal{D})}_{\text{data comprehension}}\times \underbrace{p(\boldsymbol{z},\tilde{h}|\boldsymbol{x}, \mathcal{T})}_{\text{analytical thinking}}\times\underbrace{p(h|\tilde{h},\boldsymbol{c},\boldsymbol{z},\mathcal{T})}_{\text{reflection}}d\boldsymbol{x}d\boldsymbol{z}.
\end{aligned}    
\label{decompose}
\end{align}

\begin{figure}
\centering
\includegraphics[width=\textwidth]{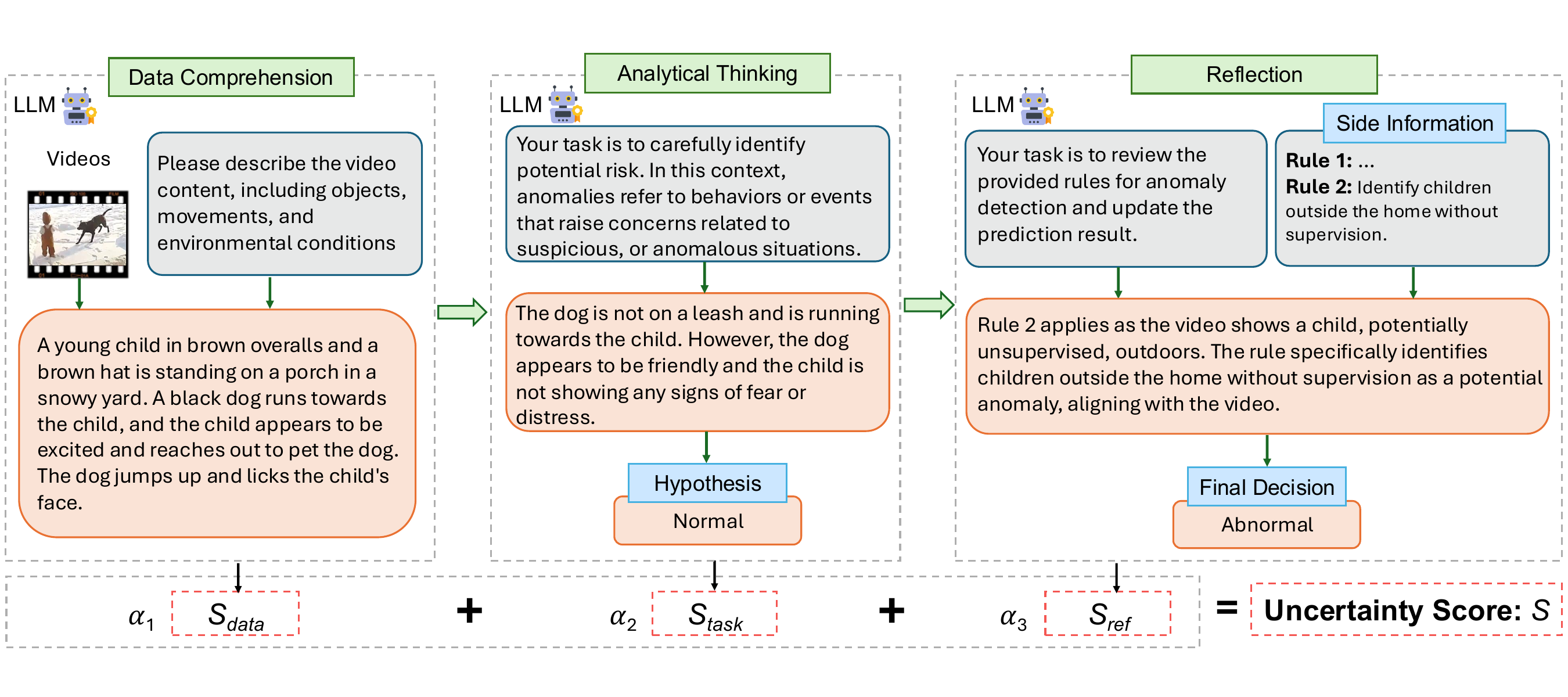}
%\vspace{-3mm}
\caption{Overview of LLM reasoning chain in ALARM for UQ in VAD.}\label{fig:overview_smart}
%\vspace{-2mm}
\end{figure}

This decomposition inspires the UQ score in ALARM as 
\begin{align}
    S=\alpha_1S_{data}+\alpha_2S_{task}+\alpha_3S_{ref},
\label{UQscore}
\end{align}
where $S$ is the UQ score for a task $\mathcal{Q}=\{\mathcal{D}, \mathcal{T}\}$, $S_{data}$, $S_{task}$, and $S_{ref}$ are the UQ scores calculated for Data Comprehension, Analytical Thinking, and Reflection in Eq. \eqref{decompose}, respectively, and $\alpha_1$, $\alpha_2$, and $\alpha_3$ are their corresponding weights. In Figure \mbox{\ref{fig:overview_smart}}, we present an overview of the LLM reasoning chain applied to a given video, visualizing the overall UQ process. The specific definitions of the individual UQ scores are shown in the following:
\begin{definition}[$S_{data}$]
$S_{data}$ is the semantic inconsistency among multiple LLMs when they describe the same data.
\end{definition}

\begin{definition}[$S_{task}$]
$S_{task}$ is the variation in reasoning outcomes that arises when LLMs analyze the data description $\boldsymbol{x}$ under the task context $\mathcal{T}$.
\end{definition}

\begin{definition}[$S_{ref}$]
$S_{ref}$ is the probability that LLMs change their initial hypothesis $\tilde{h}$ after reflecting with side information $\boldsymbol{c}$.
\end{definition}

The formulation to compute $S_{data}$, $S_{task}$, and $S_{ref}$ will be introduced in detail in Section \ref{PMFsec}. The basic idea is that, just like in statistics, we need multiple measurements to evaluate the uncertainty of a variable, here we employ multiple MLLMs (i.e., $M$ MLLMs) to replicate the task, and then develop a framework with probabilistic matrix factorization (PMF) to compute the UQ scores. The formulation to obtain optimal weights $\alpha_1$, $\alpha_2$, and $\alpha_3$ is shown in Section \ref{sec:ada_weights}. Before we dive into that, we need to first introduce the UQ-enabled quality assurance in Section \ref{sec5}.

\subsection{UQ-Enabled Quality Assurance in LLM's Decision-Makings}\label{sec5}

As shown in Figure \ref{fig:uq}, ALARM can abstain from making a prediction on a data instance when its UQ score is high. These highly uncertain instances could be sent to another decision-maker that could be either a gold-standard algorithm or human expert for evaluation. This design has been explored in the literature, for instance, \mbox{\cite{geifman2017selective}} pair trained deep neural networks with a rejection function to guarantee a desired true error rate at test time. \mbox{\cite{madras2018predict}} introduce learning to defer that jointly optimizes fairness and accuracy by allowing a model to dynamically decide whether to predict. Our uncertainty-based selective decision-making is also inspired by the line of work in learning with rejection \mbox{\citep{franc2023optimal, lin2018selective,  chow2003optimum}}. It is out of the scope of this paper, however, to further analytically model and optimize this dimension of human-AI collaboration that is possible with ALARM. Rather, in this paper, we need to first focus on the challenge of developing the UQ capacity, which also involves the concepts of cost and utility.

Denote the LLM decision-making function as $f(\cdot)$ and the human expert decision-making function as $f^*(\cdot)$. By a little abuse of the notation, we denote the uncertainty score function by $S(\cdot)$. Then, the final decision-making model $g(\cdot)$ in ALARM is as follows: 
\begin{align}
g(\mathcal{Q})=\left\{
\begin{aligned}
 f(\mathcal{\mathcal{Q}}),\quad &S(\mathcal{Q})\leq\tau\\
f^*(\mathcal{\mathcal{Q}}), \quad &S(\mathcal{Q})>\tau,
\end{aligned}\right.
\label{human_in_loop}
\end{align}
where $\tau$ is a threshold. Obviously, $\tau$ directly determines the ALARM coverage $\phi$ that is the percentage of data instances that it can make decisions (or equivalently, the rejection rate $P=1-\phi$). It is reasonable to assume that human experts incur higher expense than LLMs. A large $\phi$ means fewer queries for expensive human experts and thereby less extra cost. We can further define $\tau$ as the $\phi$-quantile of the uncertainty scores $S$, i.e., 
\begin{align*}
    P(S(\mathcal{Q})\leq\tau))\geq \phi, P(S(\mathcal{Q})>\tau)\geq P.
\end{align*}

Therefore, given $P$, we can configure the parameter $\tau$ in the decision-making rule of ALARM that is presented in Eq. \eqref{human_in_loop}. In Section \ref{sec:ada_weights} we will show how to obtain the optimal weights $\boldsymbol{\alpha}=(\alpha_1,\alpha_2,\alpha_3)$ and optimal $P$. In general, the implicit assumption behind the design of Eq. \eqref{human_in_loop} is that the human expert is more accurate than LLM, though it incurs more cost. The error of the human expert on any instance  $\boldsymbol{q}\in\mathcal{Q}$ with ground truth label $y\in\{-1, 1\}$ is $P(f^*(\boldsymbol{q})\neq y)=\delta$. The natural loss can be written as
\begin{align*}
\ell_g(\boldsymbol{q},y)=\boldsymbol{1}_{f(\boldsymbol{q})\neq y}\boldsymbol{1}_{S(\boldsymbol{q})\leq \tau}+\boldsymbol{1}_{f^*(\boldsymbol{q})\neq y}\boldsymbol{1}_{S(\boldsymbol{q})>\tau}.
\end{align*}
Then, we have the following property:
\begin{lemma}[Monotonicity]
For any instance $\boldsymbol{q}$, the value of the loss function $\ell_g(\boldsymbol{q},y)$ is a monotonic function of the uncertainty score $S(\boldsymbol{q})$.
\label{lemma0}
\end{lemma}

The proof of \textbf{Lemma 1} is in the Appendix \ref{proof_lemma1}. This lemma suggests an interesting fact that even a random UQ score $S(\boldsymbol{q})$ can improve $\ell_g(\boldsymbol{q},y)$, given that human expert is more accurate than LLM. But a better UQ score can improve $\ell_g(\boldsymbol{q},y)$ better. To see that, assuming the instances follow distribution $\mathcal{F}$, the expected risk is
\begin{align*}
&\mathcal{R}_g=\mathbb{E}_{(\boldsymbol{q},y)\sim\mathcal{F}}[\ell_g(\boldsymbol{q},y)]
=\mathbb{E}_{(\boldsymbol{q},y)\sim\mathcal{F}}[\{\boldsymbol{1}_{f(\boldsymbol{q})\neq y}-\boldsymbol{1}_{f^*(\boldsymbol{q})\neq y}\}\boldsymbol{1}_{S(\boldsymbol{q})\leq \tau}+\boldsymbol{1}_{f^*(\boldsymbol{q})\neq y}].
\end{align*}
As $S(\boldsymbol{q})$ and $f^*(\boldsymbol{q})$ are independent, and conditioned on $\boldsymbol{q}$, so are $S(\boldsymbol{q})$ and $f(\boldsymbol{q})$. We have
\begin{align*}
\mathcal{R}_g=& \mathbb{E}_{(\boldsymbol{q},y)\sim\mathcal{F}}[\boldsymbol{1}_{f(\boldsymbol{q})\neq y}\boldsymbol{1}_{S(\boldsymbol{q})\leq \tau}] -\delta P(S(\boldsymbol{q})\leq \tau)+\delta  . 
\end{align*}
The following theorem shows the effectiveness of the UQ-based selection strategy in Eq. \eqref{human_in_loop}:  

\begin{theorem}[Effectiveness]
Suppose the expected risk with a random selection strategy $S_r(\boldsymbol{q})\sim \text{Unif}(0,b)$, where $b>\tau$ is an upper bound, is denoted by $\mathcal{R}_r$. At the same coverage level $\phi$, we have $\mathcal{R}_g=\mathcal{R}_r+\text{Cov}(P(f(\boldsymbol{q})\neq y|\boldsymbol{q}),\boldsymbol{1}_{S(\boldsymbol{q})\leq \tau})$.
\label{thm1}
\end{theorem}

The proof of \textbf{Theorem 1} is in the Appendix \ref{proof_theorem1}. In Section \ref{Sec:exp} we will use random UQ score (i.e., by randomly dropping cases) as a baseline and show how much the optimal UQ score can outperform it. 

% \hl{According to \textbf{Theorem 1}, we know that the stronger the negative correlation between $\boldsymbol{1}_{S(\boldsymbol{q})\leq \tau}$ and the misclassification rates $P(f(\boldsymbol{q})\neq y|\boldsymbol{q})$, the greater risk reduction of $\mathcal{R}_g$ over $\mathcal{R}_r$.}

%as suggested by \cite{el2010foundations},
% To determine $\tau$,

% To ensure the high final detection accuracy, we only let MLLMs detect the videos with low uncertainty. The detected video set is denoted as $\mathcal{S}_{\text{low}} = \{ i \ | \ S_i \le \tau \}$. We select a threshold $\tau$ of uncertainty scores by keeping the top $(100-P)\%$ of results with the lowest uncertainty:
% \begin{equation}
% \tau = \text{Percentile}(\{S_i\}_{i=1}^N, 100-P)
% \end{equation}
% \textcolor{green}{Not sure if (100-P)\%=70\% coverage is a good number in the specific problem. If we use selective decision-making, I think it's good if we could draw a coverage-performance curve.} 

% \textcolor{red}{change US to trace of covariance} We consider the US as the variance of the learned posterior distribution for each video's latent vector $\mathbf{u}_i$. After we have approximate posterior $q(\mathbf{u}_i) = \mathcal{N}(\mathbf{u}_i | \boldsymbol{\mu}_{i}, \Sigma_{i})$, then the uncertainty for video $i$ is the trace of its posterior covariance matrix $\Sigma_{i}$.
% \begin{equation}
%     US_i = \text{Tr}(\Sigma_{i})
% \end{equation}

\subsection{Optimization of the UQ Weights} \label{sec:ada_weights}

\subsubsection{Optimal Weights for a Given \texorpdfstring{$P$}{P}.}  \label{subsec6.1} 

% \begin{figure*}[!b]
% \centering
% \includegraphics[width=0.4\textwidth]{fig/frame_sec3.3.pdf}
% \caption{Overview of the optimization of the UQ weights and rejection rate $P$.}
% \label{fig:frame_UQ_optimal}
% %\vspace{-1em}
% \end{figure*}

The basic principle is that the combined total score should, on average, outperform any individual score. The problem is how to evaluate the performance of the UQ score. Naturally, we can use accuracy as its performance metric, but note that the accuracy is not only a function of $\boldsymbol{\alpha}$ but also the rejection rate $P$, and the data sample $\mathcal{D}$. Thus, it can be denoted by $U(P, \boldsymbol{\alpha},\mathcal{D})$. With a given $P$, the optimal weights can be obtained on the training data by
\begin{align}
\boldsymbol{\alpha}=\mathop{\arg\max}_{\boldsymbol{\alpha}}\,\,U(P,\boldsymbol{\alpha},\mathcal{D}).
\label{umax}
\end{align} 
While the weights can be optimized by maximizing the detection accuracy $U(P,\boldsymbol{\alpha},\mathcal{D})$, the calculation of this objective might be affected by the noise in data, leading to unstable results. To enhance the reliability of the optimization, we instead optimize the weights in a stochastic manner:
 % \textcolor{red}{
 \begin{align}
\boldsymbol{\alpha}^*=\mathop{\arg\max}_{\boldsymbol{\alpha}}\mathbb{E}_{\mathcal{D}}[U(P,\boldsymbol{\alpha},\mathcal{D})].
\label{stoch}
\end{align}
% or
%  \begin{align*}
% \boldsymbol{\alpha}^*=\mathbb{E}_{\mathcal{D}}[\boldsymbol{\alpha}^{\mathcal{D}}]
% \end{align*}}

In practice, to obtain $\boldsymbol{\alpha}^*$ for a specific rejection rate $P$, we perform sample average approximation (SAA) for the optimization in Eq. \eqref{stoch}. Specifically, we divide the training data into $K$ folds $\mathcal{D}_1,\cdots,\mathcal{D}_K$. Each time $k=1,\cdots,K$, we use all but the $k$-th fold to learn the uncertainty model and finalize the individual uncertainty scores to calculate the combined score $S$ and the corresponding accuracy $U(P,\boldsymbol{\alpha},\mathcal{D})$ on the $k$-th fold, which generates a replicate. By taking an average over all the replicates, we approximate the optimal weights by 
 % \textcolor{red}{
 \begin{align}
\boldsymbol{\alpha}^*=\mathop{\arg\max}_{\boldsymbol{\alpha}}\frac{1}{K}\sum_{k=1}^KU(P,\boldsymbol{\alpha},\mathcal{D}_k).
\label{Kfold_opt_alpha}
\end{align}
% or
%  \begin{align*}
% \boldsymbol{\alpha}^*=\frac{1}{K}\sum_{k=1}^K\boldsymbol{\alpha}^{\mathcal{D}_k}
% \end{align*}}
For each value of $P$, we can obtain a unique set of the optimal $\boldsymbol{\alpha}^*$. With a slight abuse of notation, we denote the corresponding weights as $\boldsymbol{\alpha}^*(P)$. Ideally, a larger $P$ enables more accurate detection since more input from human experts can be incorporated. However, the increasing cost of experts usually makes the deployment with a high rejection rate $P$ impractical. As a result, it is necessary to determine the optimal $P$ under budget constraints.

\subsubsection{What is the Optimal \texorpdfstring{$P$}{P}?}\label{optimal_P_UQ}
As \textbf{Theorem \ref{thm1}} suggests, with human experts involved in the task, detection accuracy improves monotonically as $P$ increases. However, to reduce cost, $P$ should be as small as possible. Therefore, it is important to determine an optimal $P$ to balance the detection accuracy and the cost, which drives us to formulate the following optimization problem:
\begin{align}
\begin{aligned}
\min_{P} \quad& \lambda P+\mathbb{E}_{\mathcal{D}}[Q(P,\mathcal{D})]\\
s.t. \quad & P_l\leq P\leq P_u,
\end{aligned}
\label{optbudget}
\end{align}
where $\lambda$ is the cost of human labor, $P_l$ and $P_u$ are the lower/upper bounds of $P$, and $Q(P,\mathcal{D})$ is
\begin{align*}
Q(P,\mathcal{D})\triangleq \min_{\boldsymbol{\alpha}}\,C^{\mathcal{D}}(P;\boldsymbol{\alpha}).
\end{align*}

Here, $C^{\mathcal{D}}(P;\boldsymbol{\alpha})$ is the cost of potential wrong detection on data sample $\mathcal{D}$. To characterize the cost $C^{\mathcal{D}}(P;\boldsymbol{\alpha})$, let us first denote the accuracies, calculated using three uncertainty scores from single sources, $S_{data}$, $S_{task}$ and $S_{ref}$ with $\mathcal{D}$ at the rejection rate $P$, by $U_1(P,\mathcal{D})$, $U_2(P, \mathcal{D})$ and $U_3(P, \mathcal{D})$, respectively, for simplicity. We define the cost function as
\begin{align*}
    C^{\mathcal{D}}(P;\boldsymbol{\alpha})=\max_{i=1,2,3}\{U_i(P,\mathcal{D})-U(P, \boldsymbol{\alpha},\mathcal{D})\},
\end{align*}
which measures the regression in detection accuracy using the combined uncertainty score $S$ compared to using the individual uncertainty scores at a certain rejection rate $P$. It should be noted that, this definition of cost is different from the equivalent cost used in Eq. \eqref{umax} as it only measures the difference between the combined score and the individual scores. In practice, this definition is particularly useful for cost-saving, as it directly measures the relative advantage of the combined total score, which is of primary concern. The combined score needs not to be globally optimal; as long as it outperforms the individual scores, it sufficiently evidences the practical utility.

In practice, we solve the following approximation of Eq. \eqref{optbudget}
\begin{align}
\begin{aligned}
\min_{P} \quad& \lambda P+\frac{1}{K}\sum_{k=1}^KC^{\mathcal{D}_k}(P,\boldsymbol{\alpha^*})\\
s.t. \quad & P_l\leq P\leq P_u.    
\end{aligned}
\label{cost_opt}
\end{align}

A piecewise approximation solution will be employed to solve Eq. \eqref{cost_opt}, which is detailed in Section \ref{Sec:exp} that first samples a few levels of $P$ within the range of $P_l$ and $P_u$, solves for the optimal weights for each level of $P$, then smooths the trajectories of the optimal weights, and decides on the optimal $P$ which gives the minimum objective value of Eq. \eqref{cost_opt}.

\section{Computational Pipeline of the UQ in ALARM}\label{PMFsec}
Following the theoretical framework of ALARM presented in Section \ref{section3}, now we present the computational procedure to compute the UQ scores in Eq. \eqref{UQscore}. We will present details for each of the three components in Eq. \eqref{UQscore}, from Data Comprehension ($S_{data}$, Section \ref{sec_data}), to Analytical Thinking ($S_{task}$, Section \ref{sec_task}), to Reflection ($S_{ref}$, Section \ref{sec_ref}). In general, we assume that there is a training dataset of size $N$, $\mathcal{D}_{tr}$, while for each data instance we have the results from $M$ MLLMs, such that we have replications for uncertainty evaluation.

\subsection{Data Comprehension (\texorpdfstring{$S_{data}$}{S\_data}).} \label{sec_data}
To compute the uncertainty in $\boldsymbol{x}$, inspired by the recent work \citep{lingenerating, chen2025uncertainty}, we build a similarity matrix $\mathbf{W}_{\boldsymbol{x}}=[w_{i,(j,k)}]_{N\times L}$, where $L=\frac{M(M-1)}{2}$, to describe the semantic agreement among the $M$ MLLMs. Specifically, $w_{i,(j,k)}$ denotes the similarity between MLLMs $j$ and $k$ on the instance $i$. Here, $j,k = 1,2,\cdots, M, j\neq k$. For each instance $i$, denote the textual descriptions produced by the $M$ MLLMs by $\boldsymbol{x}_{i,j} (j=1,\cdots,M)$, which summarizes the events or features depicted in the instance $i$. We then propose to compute $w_{i,(j,k)}$ using the  pairwise cosine similarities: 
\begin{equation}
(\mathbf{W}_{\boldsymbol{x}})_{i,l}=w_{i,(j,k)} = \frac{\boldsymbol{x}_{i,j} \cdot \boldsymbol{x}_{i,k}}{\|\boldsymbol{x}_{i,j}\| \|\boldsymbol{x}_{i,k}\|}. 
\label{eq4}
\end{equation}

%\textcolor{red}{can we provide a proof or visual showing why this PMF makes sense as UQ}

%\vspace{-1mm}
In other words, each row of $\mathbf{W}_{\boldsymbol{x}}$ summarizes the agreement between $L$ pairs of descriptions for an instance. Intuitively, a high UQ score $S_{data}$ corresponds to more inconsistency among the $M$ MLLMs. An effective way to compute this inconsistency is through PMF \citep{mnih2007probabilistic,liu2012robust}, whereas we can measure the inconsistency among the $M$ MLLMs by the reconstruction error of the similarity matrix $\mathbf{W}_{\boldsymbol{x}}$ through matrix factorization. It starts with $\mathbf{W}_{\boldsymbol{x}}=\mathbf{U}_{\boldsymbol{x}}\mathbf{V}_{\boldsymbol{x}}^\top+\boldsymbol{\varepsilon}$, where $\mathbf{U}_{\boldsymbol{x}} \in \mathbb{R}^{N \times K}$ and $\mathbf{V}_{\boldsymbol{x}}\in \mathbb{R}^{L \times K}$ are low-rank matrices encoding the latent instance representations and the underlying structure of the pairwise consistency relationships. By assuming spherical Gaussian priors, we have the following distributions:
%\cite{tipping1999probabilistic, mnih2007probabilistic}
% \begin{align}
% &p(\mathbf{W}| \mathbf{U}, \mathbf{V}, \sigma^2) = \prod_{i=1}^{N} \prod_{\substack{j,k=1\\j\neq k}}^{M} \left[ \mathcal{N}(w_{i,(j,k)}| \mathbf{u}_i \mathbf{v}_l^{\top}, \sigma^2) \right]^{a_{il}} \nonumber\\
% &p(\mathbf{U} | \sigma_U^2) = \prod_{i=1}^{N} \mathcal{N}(\boldsymbol{u}_i | \boldsymbol{0}, \sigma_U^2 \mathbf{I}) \nonumber\\
% &p(\mathbf{V} | \sigma_V^2) = \prod_{\substack{j,k=1\\j\neq k}}^{M} \mathcal{N}(\boldsymbol{v}_l | \boldsymbol{0}, \sigma_V^2 \mathbf{I}) 
% \end{align}
% \begin{align*}
% &p(\mathbf{W}| \mathbf{U}, \mathbf{V}, \sigma^2) = \prod_{i=1}^{N} \prod_{l=1}^{L} \left[ \mathcal{N}(\mathbf{W}_{i,l}| \mathbf{u}_i \mathbf{v}_l^{\top}, \sigma^2) \right]^{a_{il}} \nonumber\\
% &p(\mathbf{U} | \sigma_U^2) = \prod_{i=1}^{N} \mathcal{N}(\boldsymbol{u}_i | \boldsymbol{0}, \sigma_U^2 \mathbf{I}) \nonumber\\
% &p(\mathbf{V} | \sigma_V^2) = \prod_{l=1}^{L} \mathcal{N}(\boldsymbol{v}_l | \boldsymbol{0}, \sigma_V^2 \mathbf{I}) 
% \end{align*}
% \allowdisplaybreaks  % Place this in the preamble
\begin{align*}
&p(\mathbf{W}_{\boldsymbol{x}}| \mathbf{U}_{\boldsymbol{x}}, \mathbf{V}_{\boldsymbol{x}}, \sigma^2) = \prod_{i=1}^{N} \prod_{l=1}^{L} \left[ \mathcal{N}((\mathbf{W}_{\boldsymbol{x}})_{i,l}| \mathbf{u}_i \mathbf{v}_l^{\top}, \sigma^2) \right]^{a_{il}} \nonumber\\
&p(\mathbf{U}_{\boldsymbol{x}} | \sigma_U^2) = \prod_{i=1}^{N} \mathcal{N}(\boldsymbol{u}_i | \boldsymbol{0}, \sigma_U^2 \mathbf{I}) \nonumber\\
&p(\mathbf{V}_{\boldsymbol{x}} | \sigma_V^2) = \prod_{l=1}^{L} \mathcal{N}(\boldsymbol{v}_l | \boldsymbol{0}, \sigma_V^2 \mathbf{I}), 
\end{align*}
% (where $l=j+k-1$) w_{i,(j,k)}
where $\mathbf{u}_i$ and $\mathbf{v}_l$ are corresponding row vectors of $\mathbf{U}_{\boldsymbol{x}}$ and $\mathbf{V}_{\boldsymbol{x}}$ respectively,  $a_{il}$ is an indicator of whether $(\mathbf{W}_{\boldsymbol{x}})_{i,l}$ exists (i.e., not missing due to technical issues and generation failure), and $\mathbf{I}$ is the identity matrix. We can obtain the optimal matrix factorization $\mathbf{U}_{\boldsymbol{x}}^*$ and $\mathbf{V}_{\boldsymbol{x}}^*$ through minimizing the expected reconstruction error, which is approximated by an empirical regularized reconstruction loss:
%\cite{mnih2007probabilistic}
\begin{align*}
&\mathcal{L} = \left\Vert \mathbf{A}\odot \left(\mathbf{W}_{\boldsymbol{x}} -\mathbf{U}_{\boldsymbol{x}}\mathbf{V}_{\boldsymbol{x}}^\top \right) \right\Vert_F^2 + \frac{\sigma^2}{\sigma^2_{U}} \sum_{i=1}^{N} \left\| \mathbf{U}_{\boldsymbol{x}}\right\|_{\mathrm{F}}^2 
+ \frac{\sigma^2}{\sigma^2_{V}} \sum_{l=1}^{L} \left\|\mathbf{V}_{\boldsymbol{x}}\right\|_{\mathrm{F}}^2\nonumber,
% &+ \frac{\sigma^2}{2\sigma^2_{U}} \sum_{i=1}^{N} \left\|  \boldsymbol{u}_i \right\|_{\mathrm{F}}^2 + \frac{\sigma^2}{2\sigma^2_{V}} \sum_{l=1}^{L} \left\|  \boldsymbol{v}_l\right\|_{\mathrm{F}}^2
\end{align*}
% \begin{align}
% \begin{aligned}
% \mathcal{L} &= \frac{1}{2} \sum_{i=1}^{N} \sum_{l=1}^{L} a_{il} \left(\mathbf{W}_{i,l} - \boldsymbol{u}_i^{\top} \boldsymbol{v}_l \right)^2 \\
% &+ \frac{\sigma^2}{2\sigma^2_{U}} \sum_{i=1}^{N} \left\| \mathbf{U}\right\|_{\mathrm{F}}^2 
% + \frac{\sigma^2}{2\sigma^2_{V}}\left\|\mathbf{V}\right\|_{\mathrm{F}}^2
% % &+ \frac{\sigma^2}{2\sigma^2_{U}} \sum_{i=1}^{N} \left\|  \boldsymbol{u}_i \right\|_{\mathrm{F}}^2 + \frac{\sigma^2}{2\sigma^2_{V}} \sum_{l=1}^{L} \left\|  \boldsymbol{v}_l\right\|_{\mathrm{F}}^2
% \end{aligned}    
% \end{align}
where $\mathbf{A}=[a_{il}]_{N\times L}$ and $\odot$ is the Hadamard product. Then, the uncertainty score for each instance $i$ can be calculated by the corresponding reconstruction error $\varepsilon_i=||(\mathbf{W}_{\boldsymbol{x}})_{i,\cdot}-\boldsymbol{u}_i\mathbf{V}_{\boldsymbol{x}}^\top||_2^2$. 

Note that we can estimate $\mathbf{V}_{\boldsymbol{x}}^*$ based on the training data $\mathcal{D}_{tr}$. Then, for any new data instance, we can compute its uncertainty score by solving a least-square problem:
\begin{align}
    S_{data}=\min_{\boldsymbol{\beta}}||\boldsymbol{w}_{\boldsymbol{x}}-\boldsymbol{\beta}\cdot{\mathbf{V}_{\boldsymbol{x}}^*}^\top||_2^2,
\label{S_data_cal}
\end{align}
where $\boldsymbol{w}_{\boldsymbol{x}}$ is the similarity vector of data description $\boldsymbol{x}$ of a new data instance using Eq. \eqref{eq4}.

\subsection{Analytical Thinking (\texorpdfstring{$S_{task}$}{S\_task}).}\label{sec_task}

\begin{figure*}[h]
\centering
\includegraphics[width=0.7\textwidth]{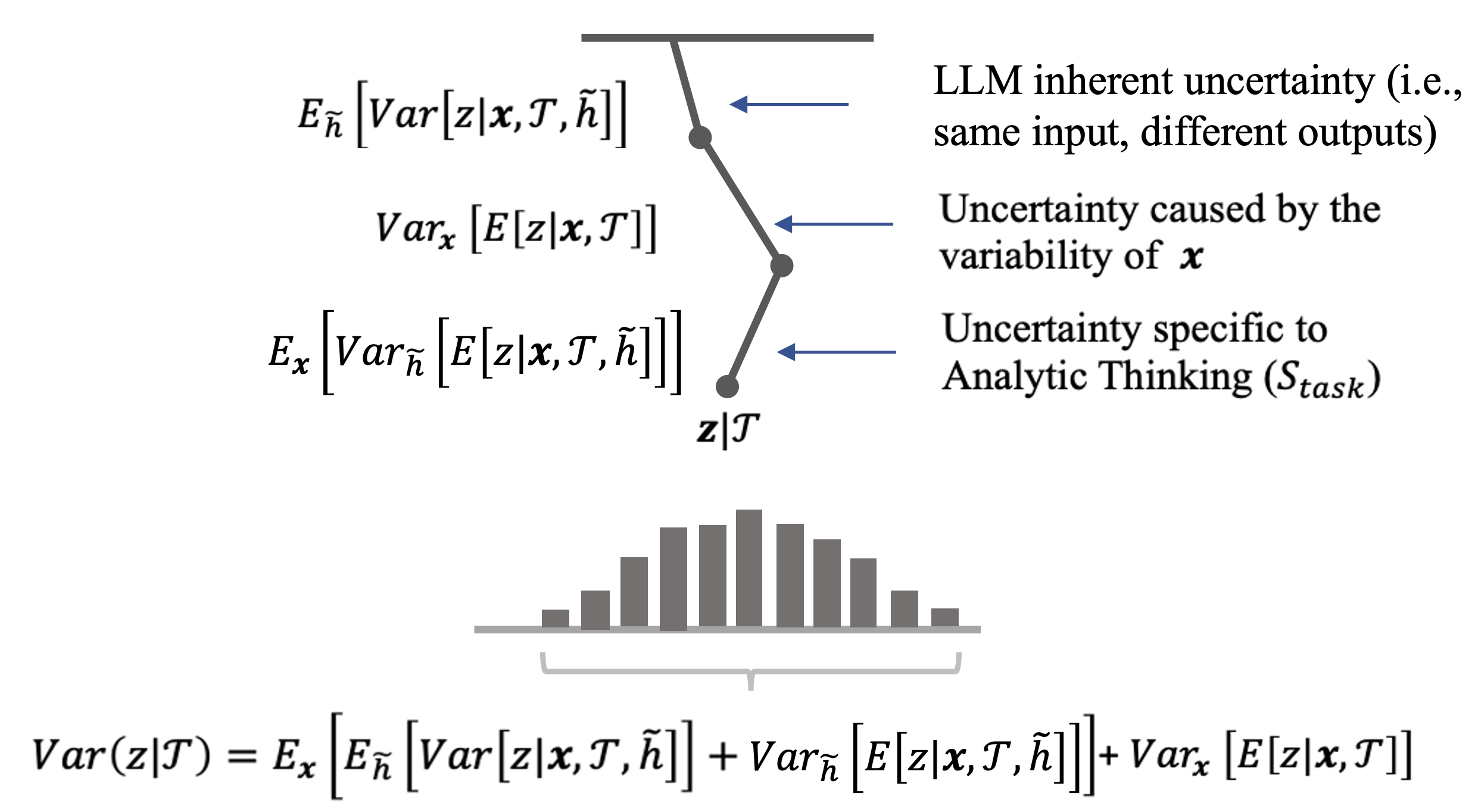}
\caption{Illustration of the variation analysis to extract $S_{task}$ from the total variability in $\text{Var}[\boldsymbol{z}|\mathcal{T}]$. }
\label{fig:vari_analysis_S_task}
%\vspace{-1em}
\end{figure*}

The next uncertainty score $S_{task}$ concerns the part when MLLMs perform Analytical Thinking, that is, to generate $\boldsymbol{z}$. As $\boldsymbol{z}$ is computed by MLLM based on the input $\boldsymbol{x}$ and $\mathcal{T}$, the total variability of $\boldsymbol{z}$ will necessarily include not only the uncertainty inherent in the Analytical Thinking part (which is the target of $S_{task}$) but also other uncertainties associated with $\boldsymbol{x}$ and MLLM itself. By the law of total variance, we can define $S_{task}$ as:
\begin{equation}
\begin{aligned}
S_{task}&=\text{Var}[\boldsymbol{z}|\mathcal{T}]=
\underbrace{\mathbb{E}_{\boldsymbol{x}}[\text{Var}[\boldsymbol{z}|\boldsymbol{x},\mathcal{T}]]}_{\substack{\text{expected variability}\\ \text{in analytic thinking}}}+\underbrace{\text{Var}_{\boldsymbol{x}}[\mathbb{E}[\boldsymbol{z}|\boldsymbol{x},\mathcal{T}]]}_{\text{variability from }\boldsymbol{x}}.
% $}
\label{eq:z_T}
\end{aligned}
\end{equation}
% \begin{align}
% \hlm{S_{task}=\text{Var}[\boldsymbol{z}|\mathcal{T}]=
% \underbrace{\mathbb{E}_{\boldsymbol{x}}[\text{Var}[\boldsymbol{z}|\boldsymbol{x},\mathcal{T}]]}_{\substack{\text{expected variability}\\ \text{in analytic thinking}}}+\underbrace{\text{Var}_{\boldsymbol{x}}[\mathbb{E}[\boldsymbol{z}|\boldsymbol{x},\mathcal{T}]]}_{\text{variability from }\boldsymbol{x}}.}
% % $}
% \label{eq:z_T}
% \end{align}
As shown in Figure \ref{fig:vari_analysis_S_task}, we need to break down the distinct components in the total variability $\boldsymbol{z}$ and remove the other components that are not related to Analytical Thinking, so we can properly compute $S_{task}$.

% Provided with descriptions $\boldsymbol{x}$, the LLMs will perform analytical thinking in which they generate task-related reasoning and a hypothesis $\tilde{h}$. Since the reasoning is conditioned on $\boldsymbol{x}$, the resulting uncertainty in $\tilde{h}$ can arise not only from the variability in the generated justifications, but also from the variations in the descriptions. To accurately capture the true uncertainty attributed to analytical thinking, we define a task-specific uncertainty score $S_{task}$ which isolates the contribution of reasoning by filtering out the impact of $\boldsymbol{x}$. 

% \begin{figure*}[!t]
% \centering
% \includegraphics[width=0.7\textwidth]{fig/vari_analysis_S_task.png}
% \caption{Illustration of the variation analysis to extract $S_{task}$ from the total variability in $\text{Var}[\boldsymbol{z}|\mathcal{T}]$. }
% \label{fig:vari_analysis_S_task}
% %\vspace{-1em}
% \end{figure*}

\begin{theorem}[]
Suppose the inherent variability in the LLMs is negligible, we have 
\begin{align*}
S_{task} = \mathbb{E}_{\tilde{h}}[R_{\tilde{h}}] - R,
\end{align*}
where $R_{\tilde{h}}=\mathbb{E}_{\boldsymbol{x}}\bigl[\mathbb{E}[\boldsymbol{z}|\boldsymbol{x},\mathcal{T},\tilde{h}]-\mathbb{E}[\boldsymbol{z}|\mathcal{T}]\bigr]^2$ and $R=\mathbb{E}_{\boldsymbol{x}}\bigl[\mathbb{E}[\boldsymbol{z}|\boldsymbol{x},\mathcal{T}]-\mathbb{E}[\boldsymbol{z}|\mathcal{T}]\bigr]^2$. 
\label{theorem_S_task}
\end{theorem}

The proof of \textbf{Theorem 2} is in the Appendix \ref{proof_theorem2} and its basic idea is shown in Figure \ref{fig:vari_analysis_S_task}. $S_{task}$ is the variation in reasoning outcomes that arises when LLMs analyze the data description $\boldsymbol{x}$ under the task context $\mathcal{T}$.

To characterize such variations in Analytical Thinking, with the same techniques in Section \ref{sec_data}, we construct the reasoning similarity matrix $\mathbf{W}_{\boldsymbol{z}}$ and apply PMF to estimate $R$ through the reconstruction error. To compute $R_{\tilde{h}}$, for different hypotheses $\tilde{h}$, we construct a set of hypothesis-dependent similarity matrices $\mathbf{W}^{\tilde{h}}_{\boldsymbol{z}}$ and repeat the same procedure for each hypothesis independently. It then gives $R = \displaystyle \min_{\boldsymbol{\beta}} \left\| \boldsymbol{w}_{\boldsymbol{z}} - \boldsymbol{\beta} \cdot \mathbf{V}^{*\top}_{\boldsymbol{z}} \right\|_2^2$ and $R_{\tilde{h}}=\displaystyle \min_{\boldsymbol{\beta}} \left\| \boldsymbol{w}_{\boldsymbol{z}}^{\tilde{h}} - \boldsymbol{\beta} \cdot \mathbf{V}^{*\top}_{\boldsymbol{z}} \right\|_2^2$, where $\mathbf{V}_{\boldsymbol{z}}^*$ is the basis learned from the reference dataset $\mathcal{D}_{tr}$ and remains identical across all $\tilde{h}$. As a result, we can define the uncertainty score $S_{task}$ as:
\begin{equation}
\begin{aligned}
S_{task}=\Delta R &= \mathbb{E}_{\tilde{h}}\left[\min_{\boldsymbol{\beta}} \left\| \boldsymbol{w}_{\boldsymbol{z}}^{\tilde{h}} - \boldsymbol{\beta} \cdot \mathbf{V}^{*\top}_{\boldsymbol{z}} \right\|_2^2\right]  - \min_{\boldsymbol{\beta}} \left\| \boldsymbol{w}_{\boldsymbol{z}} - \boldsymbol{\beta} \cdot \mathbf{V}^{*\top}_{\boldsymbol{z}} \right\|_2^2.
\end{aligned}
\label{S_task_cal}
\end{equation}

\subsection{Reflection (\texorpdfstring{$S_{ref}$}{S\_ref}).} \label{sec_ref}
During the Reflection stage, MLLMs reevaluate their initial hypotheses $\tilde{h}$ with task-specific side information $\boldsymbol{c}$. They are allowed to revise their reasoning $\boldsymbol{z}$ and potentially adjust their hypothesis $\tilde{h}$, leading to more reliable outcomes. To describe this adjustment, we define a binary indicator $y_{adj}$ that $y_{adj} = 1$ if $h \neq \tilde{h}$, otherwise $y_{adj} = 0$. When MLLMs are more uncertain, their hypotheses $\tilde{h}$ are subject to higher chance of alteration. Thus, the final decision $h$ will be different from the initial $\tilde{h}$, and the probability of $y_{adj}=1$ is expected to increase. As a result, the uncertainty in the decision-making of MLLMs can be quantified through the probability of such inconsistency conditioned on the side information $\boldsymbol{c}$, the reasoning $\boldsymbol{z}$ and the initial hypotehsis $\tilde{h}$, i.e., $p(y_{adj}=1|\boldsymbol{c},\boldsymbol{z},\tilde{h})$, which can be formulated by a binary classification model parametrized by $\boldsymbol{\theta}$:
\begin{align*}
&p(y_{adj}=1|\boldsymbol{c},\boldsymbol{z},\tilde{h}) = \sigma(e(\boldsymbol{c},\boldsymbol{z},\tilde{h});\boldsymbol{\theta}).
\end{align*}
Here, $\sigma(\cdot)$ denotes the sigmoid function and $e(\cdot)$ represents an embedding function, where $\boldsymbol{\theta}$ can be obtained by minimizing the binary cross-entropy:
\begin{align*}
&\mathcal{L} = -y_{adj} \log(p(y_{adj}=1|\boldsymbol{c},\boldsymbol{z},\tilde{h})) - (1 - y_{adj})  \log(1 - p(y_{adj}=1|\boldsymbol{c},\boldsymbol{z},\tilde{h})).
\end{align*}
Our Reflection uncertainty score $S_{ref}$ is then defined as the expected probability of a decision change. When $M$ LLM reasoning chains are involved, we assume a uniform distribution over them, and the uncertainty can be approximated by
\begin{align}
S_{ref}=\frac{1}{M}\sum_{j=1}^Mp_j(y_{adj}=1|\boldsymbol{c},\boldsymbol{z}_j,\tilde{h}_j),
\label{S_ref_cal}
\end{align}
where $\boldsymbol{z}_j$ and $\tilde{h}_j$ represent the reasoning and the initial hypothesis made by the $j$-th MLLM. A higher $S_{ref}$ indicates greater instability in the decision process caused by reflection, signaling lower confidence in the MLLMs' initial conclusions.

\section{Case Studies} \label{Sec:exp}

We thoroughly evaluate the performance of ALARM in two real-world applications. One is smart-home VAD and another one is wound classification. Both are known to be challenging problems \citep{stojkoska2017review, malone2022challenges} given the complexity and ambiguity of the problems and the limited data sizes for model development. We showcase that ALARM can achieve outstanding results on both cases and demonstrate how UQ plays an essential role to give flexibility and performance boost for MLLM-based decision-making and prediction pipelines.

\subsection{Experiment Setup}\label{exp_setup}

For both case studies we replicate the experiments and evaluation using multiple MLLMs, in order to derive robust model performance but also provides the necessary replications to compute the UQ scores (i.e., as elaborated in Section \ref{PMFsec}). For the smart-home application, we adopt five MLLMs that include Gemini-1.5-flash, Gemini-1.5-pro, GPT-4o, GPT-4o-mini, and Claude-3.5-sonnet. For the wound classification problem, we use Claude-3.7-sonnet, Claude-3.5-haiku, GPT-4o, GPT-4o-mini, and Claude-3.5-sonnet. The difference of the MLLMs used in both applications is because of the varying availability of the MLLMs at the time when the experiments were conducted. In our experiences, the quality of these MLLMs are consistently equal with each other and since our primary motivation for using multiple MLLMs is to increase replications of our results to compute the UQ, it is not necessary that we use the same set of MLLMs across different case studies. 

We compare the performance of ALARM against several baseline methods for VAD: zero-shot prompting (i.e., simply describing the tasks \citep{liu2023pre}), Chain-of-Thought prompting (i.e., supplying MLLMs with detailed instructions and prompting them to reason step by step \citep{wang2022self}), few-shot learning (i.e., extending Chain-of-Thought by additionally providing correctly predicted examples \citep{wei2022chain}), in-context learning (i.e., extending Chain-of-Thought by additionally providing reference rules related to the tasks \citep{inan2023llama}), LLM reasoning chain without UQ (i.e., the three-stage reasoning chain in our ALARM framework without incorporating UQ), and LLM reasoning chain with random drop (i.e., randomly abstaining from prediction on $P\%$ of the data in reasoning chain). We also compare ALARM with four additional UQ methods recently developed in the literature that are compatible with our framework and whose source codes are accessible for our implementation, i.e., LAC \mbox{\citep{ye2024benchmarking, sadinle2019least}}, APS \mbox{\citep{ye2024benchmarking,romano2020classification}}, ICL-EU and ICL-AU \mbox{\citep{ling2024uncertainty}}. We provide descriptions of these 4 UQ methods in Section \mbox{\ref{related_work_UQ}}. These UQ methods can be incorporated into our framework to conduct selective classification as described in Section \mbox{\ref{sec5}}, Eq. \mbox{\eqref{human_in_loop}}, i.e., by rejecting the top $P$\% of data instances with the highest UQ scores. We leverage the outputs from all the five MLLMs in three stages, i.e., Data Comprehension, Analytical Thinking, and Reflection, and use 5-fold cross-validation to compute average $S_{data}$ (Eq. \eqref{S_data_cal}), $S_{task}$ (Eq. \eqref{S_task_cal}), and $S_{ref}$ (Eq. \eqref{S_ref_cal}) and then derive the optimal weights optimized by Eq. \eqref{Kfold_opt_alpha} to integrate the normalized individual scores (i.e., $S_{data}, S_{task}, S_{ref}$) into the final uncertainty scores $S$ for each data instance. To ensure a fair comparison, the performance results related to selective classification reported in the following sections are computed solely on the ($100-P$)\% samples that are not deferred to human. These samples are classified by majority voting among the five MLLMs, without any human involvement. Without losing generality and avoiding presenting repetitive results, we set $P=5\%$ for ALARM unless otherwise specified. Additional experiment setup is provided in the Appendix \mbox{\ref{add_exp_setup}}.

\subsection{Smart-Home}

The smart-home dataset was collected in \cite{Zhao_2025_CVPR} that includes a set of 1,203 videos captured by both indoor and outdoor smart-home cameras with 554 normal videos and 649 abnormal videos (which includes 91 ambiguous videos that even humans have difficulty determining if they are normal or abnormal). As shown in Figure \ref{fig:overview_smart}, to implement ALARM, for any MLLM, when given a video, the MLLM is first prompted to perform Data Comprehension by generating a video description $\boldsymbol{x}$. Next, based on this description, the MLLM conducts Analytical Thinking and provides its reasoning result $\boldsymbol{z}$, leading to an initial hypothesis $\tilde{h}$ regarding the video’s anomaly label. To enhance this process, we incorporate side information $\boldsymbol{c}$ when we provide the prompts for MLLMs, i.e., here, it is a set of anomaly rules (as shown in Online Figure \ref{fig:taxonomy}) developed by domain experts. For example, one such rule states that children outside the home without supervision should be considered anomalous. Equipped with the anomaly rules $\boldsymbol{c}$, initial reasoning $\boldsymbol{z}$, and a task-specific prompt $\mathcal{T}$, the MLLM then engages in Reflection to reassess the hypothesis $\tilde{h}$ and produce the final prediction $h$. Both $\tilde{h}$ and $h$ are binary, while 1 indicates abnormal and 0 indicates normal. For instance, a video showing an unattended child outdoors may initially be misclassified as ``normal''. After incorporating $\boldsymbol{c}$ and reflective reasoning, MLLMs can correct such oversights, arriving at the more reliable final prediction $h$.

\subsubsection{Overall Evaluation}
As shown in Table \ref{tab:average_smart}, our ALARM framework  achieves the best scores on every performance evaluation metric. Rejecting even a small fraction of videos improves overall accuracy by 7.75\% over TRLC (the benchmark method developed in \cite{Zhao_2025_CVPR} for this smart-home dataset) and by 2.72\% over reasoning chain with random drop. Based on the metric recall we can see that ALARM improves even more markedly than the other approaches (i.e., 9.16\% over TRLC). For the especially difficult ambiguous class (i.e., the 91 ambiguous data instances, denoted as $D_{Ambiguity}$), ALARM delivers a 9.65\% gain over TRLC, indicating that the UQ score of ALARM is more effective in detecting challenging cases for MLLMs.

\begin{table*}[htbp]
  \centering
  \caption{Performance of anomaly detection in smart homes across five MLLMs in different methods.}
  \small % Reduce font size
  \begin{tabular}{lcccc}
    \toprule
    \textbf{Methods} & \textbf{Overall Accuracy} & \textbf{Recall} & \textbf{F1 Scores} & \textbf{Accuracy on $D_{Ambiguity}$} \\
    \midrule
    Zero Shot & 64.99  & 51.83 & 58.82 & 43.73 \\
    Chain-of-Thought & 71.39 & 66.78 & 71.78 & 44.21 \\
    Few Shot & 71.49  & 66.90 & 71.89 & 43.16 \\
    In-Context Learning & 71.94  & 58.77 & 69.49 & 31.79 \\
    TRLC & 76.59  & 81.20 & 78.98 & 61.54 \\
    Reasoning Chain w/ Random Drop & 81.62  & 86.38 & 83.51 & 61.83\\
    Reasoning Chain w/ LAC & 83.53  & 89.69 & 84.96 & 67.94\\
        Reasoning Chain w/ APS & 81.61  & 86.13 & 83.41 & 61.44\\
        Reasoning Chain w/ ICL-EU & 83.45  & 86.95 & 85.11 & 63.64\\
        Reasoning Chain w/ ICL-AU & 81.78  & 86.10 & 83.82 & 61.63\\
    ALARM (\textbf{ours}) & \textbf{84.34}  & \textbf{90.36} & \textbf{86.17} & \textbf{71.19} \\
    \bottomrule
  \end{tabular}
  %\vspace{-4mm}
  \label{tab:average_smart}
\end{table*}

\begin{figure*}[!htbp]
    \centering
    \begin{subfigure}{0.4\textwidth} 
        \centering
        \includegraphics[width=\linewidth]{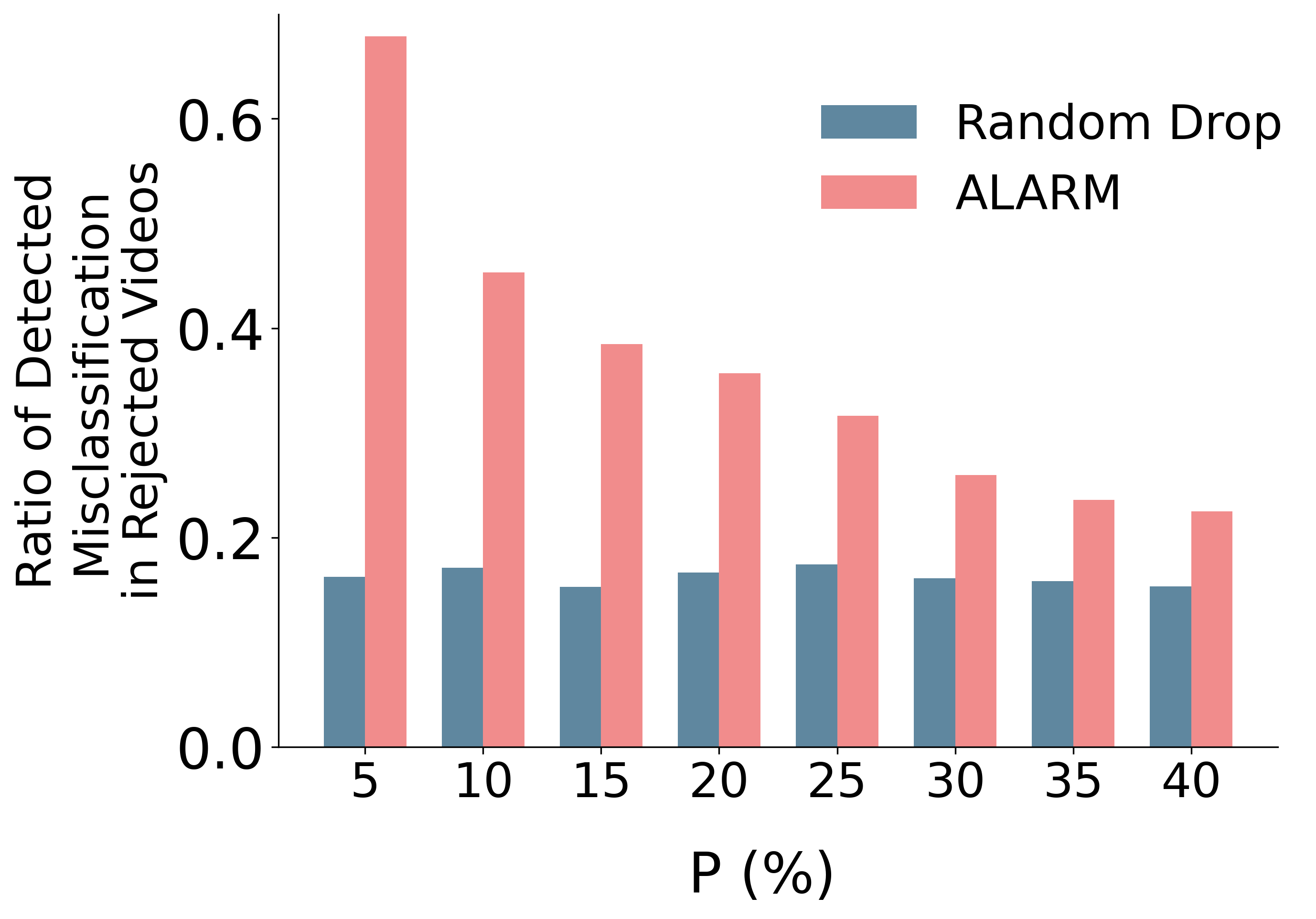}
        \caption{Smart home.}
        \label{fig:missclass_ratio_smart}
    \end{subfigure}%
    %\hfill
    \hspace{0.001\textwidth}
    \begin{subfigure}{0.4\textwidth} 
        \centering
        \includegraphics[width=\linewidth]{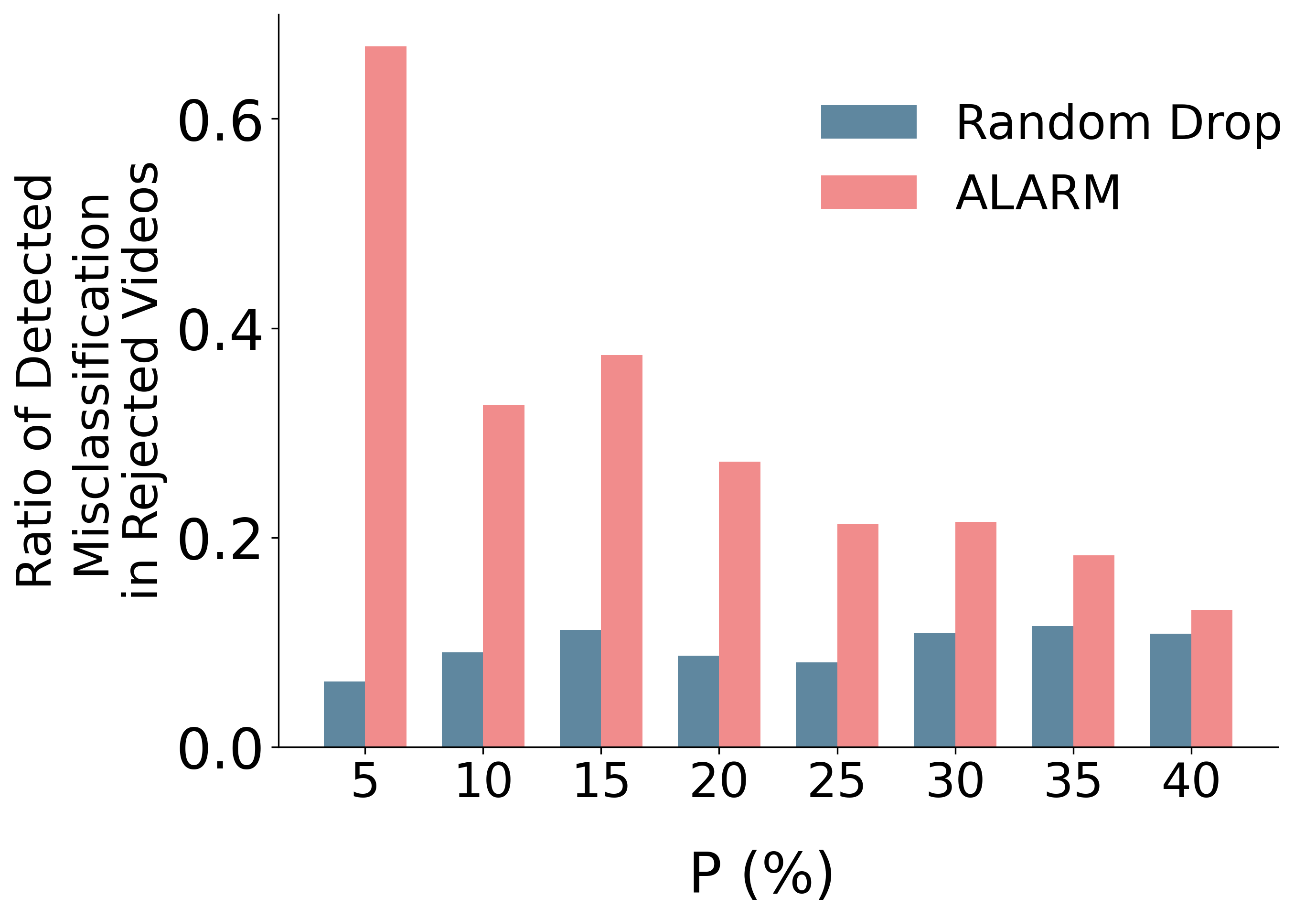}
        \caption{Wound classification.}
        \label{fig:missclass_ratio_wound}
    \end{subfigure}%
    \caption{The ratio of detected misclassification in the rejected cases.}
    \label{fig:missclass_both}
\end{figure*}

One may have the impression that ALARM (with the UQ) only marginally improves on the LLM reasoning chain with Random Drop as the accuracy improvement is only 2.72\%. This is a tempting misimpression because the 2.72\% is computed based on the whole testing dataset. The base effect of the sample size of the testing dataset diluted the contribution of the UQ. To better appreciate the power of UQ, here, we show in Figure \ref{fig:missclass_ratio_smart} the ratio of detected misclassification in the rejected cases by both UQ and Random Drop, i.e., if out of 10 rejected cases, 2 are truly misclassified by the MLLM, this ratio is 0.2. One can see in Figure \ref{fig:missclass_ratio_smart} that our proposed UQ method is very effective since most of its rejected cases are truly misclassified by MLLM when P is small. When P increases, this ratio of detected misclassification by UQ also gradually decreases, while still outperforming Random Drop. Meanwhile, the ratio of detected misclassification by Random Drop remains the same. This is expected since Random Drop blindly rejects cases. We further defer the readers to see Figure \ref{fig:metric_smart} and the corresponding discussion that demonstrate the ALARM (with the UQ) significantly and consistently improves on the LLM reasoning chain with Random Drop across all the levels of P and in terms of both accuracy and recall.

\subsubsection{The Impact of Rejection Rate \texorpdfstring{$P$}{P}}
\begin{figure*}[!htbp]
    \centering
    \begin{subfigure}{0.35\textwidth} 
        \centering
        \includegraphics[width=\linewidth]{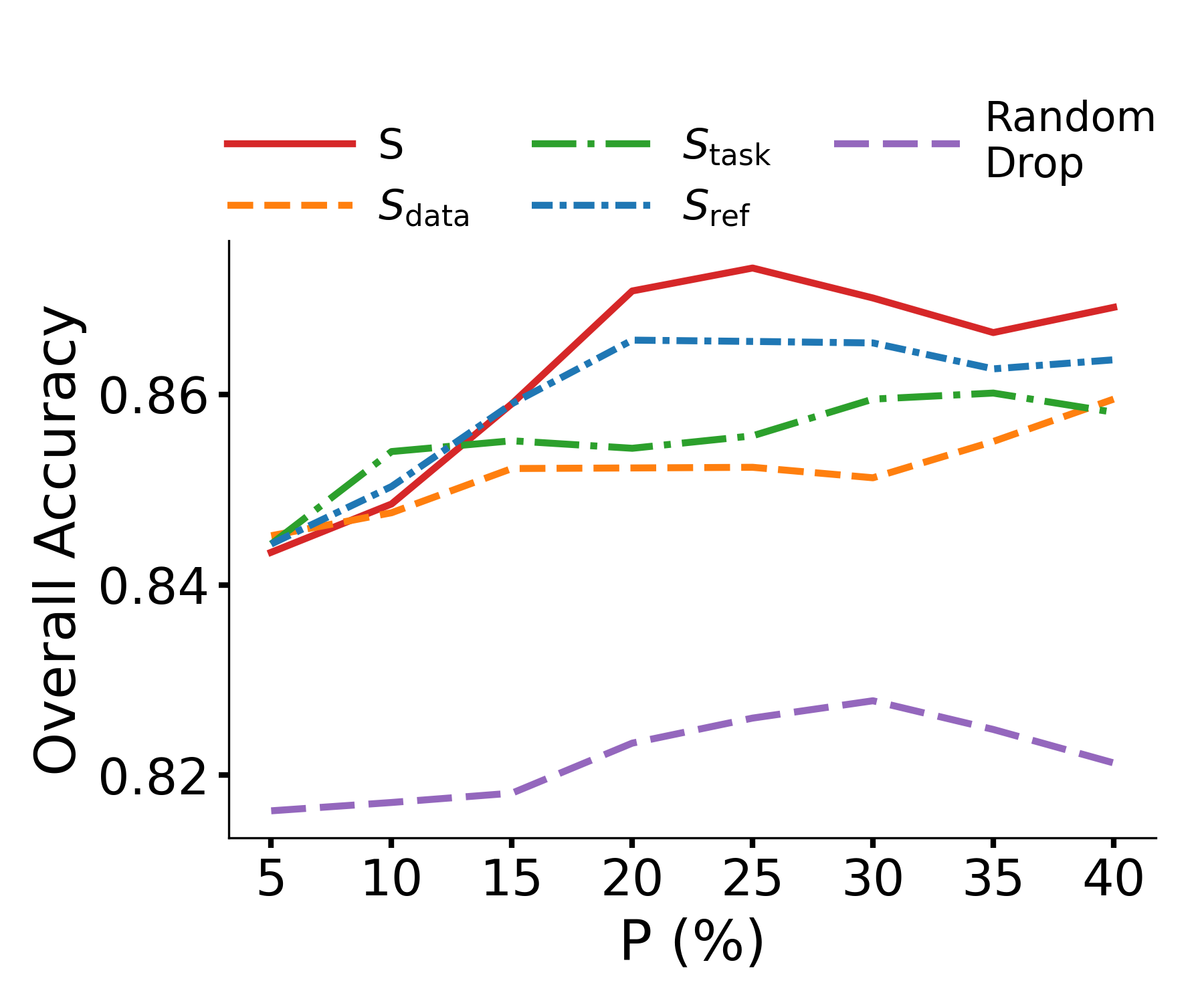}
        \caption{Overall Accuracy.}
        \label{fig:sub_offensiveness}
    \end{subfigure}%
    %\hfill
    \hspace{0.001\textwidth}
    \begin{subfigure}{0.35\textwidth} 
        \centering
        \includegraphics[width=\linewidth]{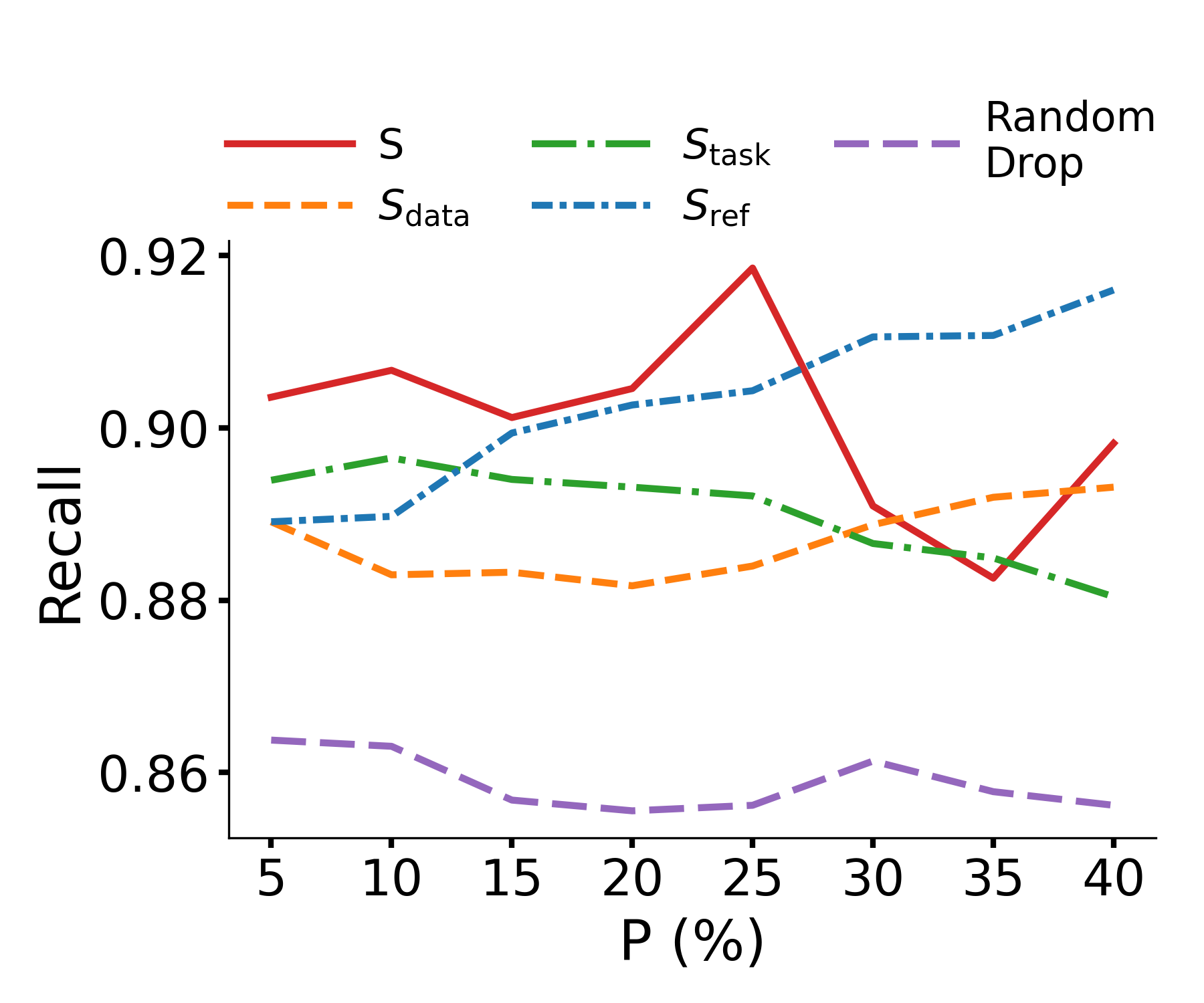}
        \caption{Recall.}
        \label{fig:sub_politeness}
    \end{subfigure}%
    \caption{Metric trends of different uncertainty scores in ALARM as the rejection rate $P$ varies in smart home.}
    \label{fig:metric_smart}
\end{figure*}

Figure \ref{fig:metric_smart} presents the overall accuracy (a) and recall (b) of ALARM as the rejection rate $P$ varies from 5\% to 40\%. Recall that a lower $P$ means more videos are assigned by ALARM for MLLMs to make the final decisions $h$. It can be seen that ALARM with the uncertainty score $S$ consistently outperforms baseline Random Drop in terms of both metrics across the range of $P$. And the individual scores $S_{data}$, $S_{task}$, and $S_{ref}$ also outperform the naive Random Drop. The accuracy reaches its maximum when $P$ = 25\%. When $P$ becomes too large, some correctly classified videos are also removed because of the high rejection rate, leading to a plateau in accuracy improvement. Recall shows a similar trend. One thing we can observe is that the recall of $S$ exhibits fluctuations. This is because the optimal weights of $S$ are optimized for accuracy rather than recall. But in general, even though the weights are not optimized for recall, at most values of $P$, the recall of $S$ still surpasses that of any individual uncertainty source. Last but not least, we can also observe that at small $P$ values (i.e., $5\%$ and $10\%$), the accuracies of $S$ and the three individual uncertainty scores are very close. This is because of the limited rejected cases (when $P$ is small) that are insufficient to separate the different methods.

\subsubsection{The Trajectory of the Optimal Weights \texorpdfstring{$\boldsymbol{\alpha}$}{alpha}}
\begin{figure*}[!htbp]
    \centering
    \begin{subfigure}{0.32\textwidth} 
        \centering
        \includegraphics[width=\linewidth]{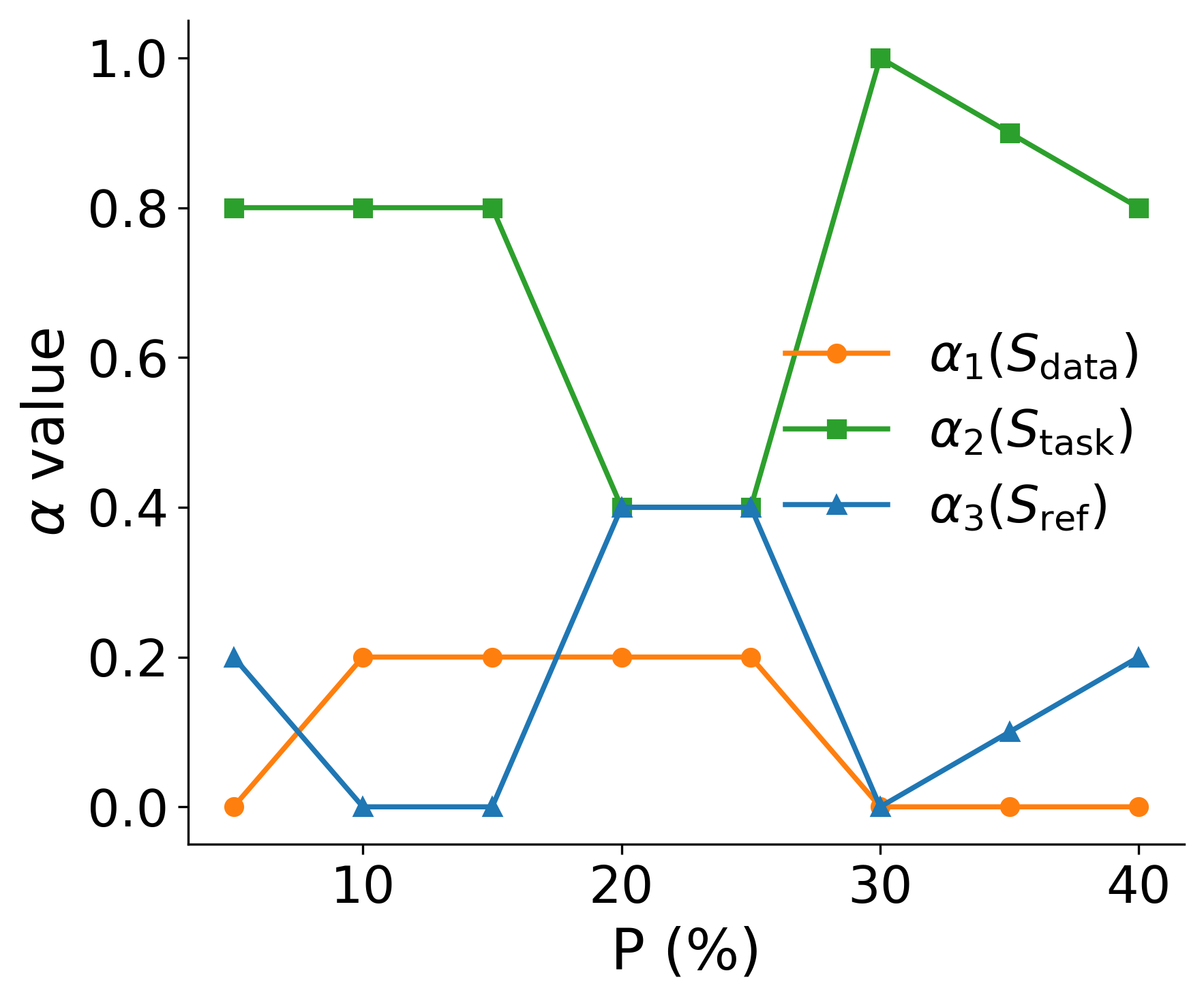}
        \caption{Optimal weights.}
        \label{fig:optimal_smart}
    \end{subfigure}%
    \hfill
        \begin{subfigure}{0.32\textwidth} 
        \centering
        \includegraphics[width=\linewidth]{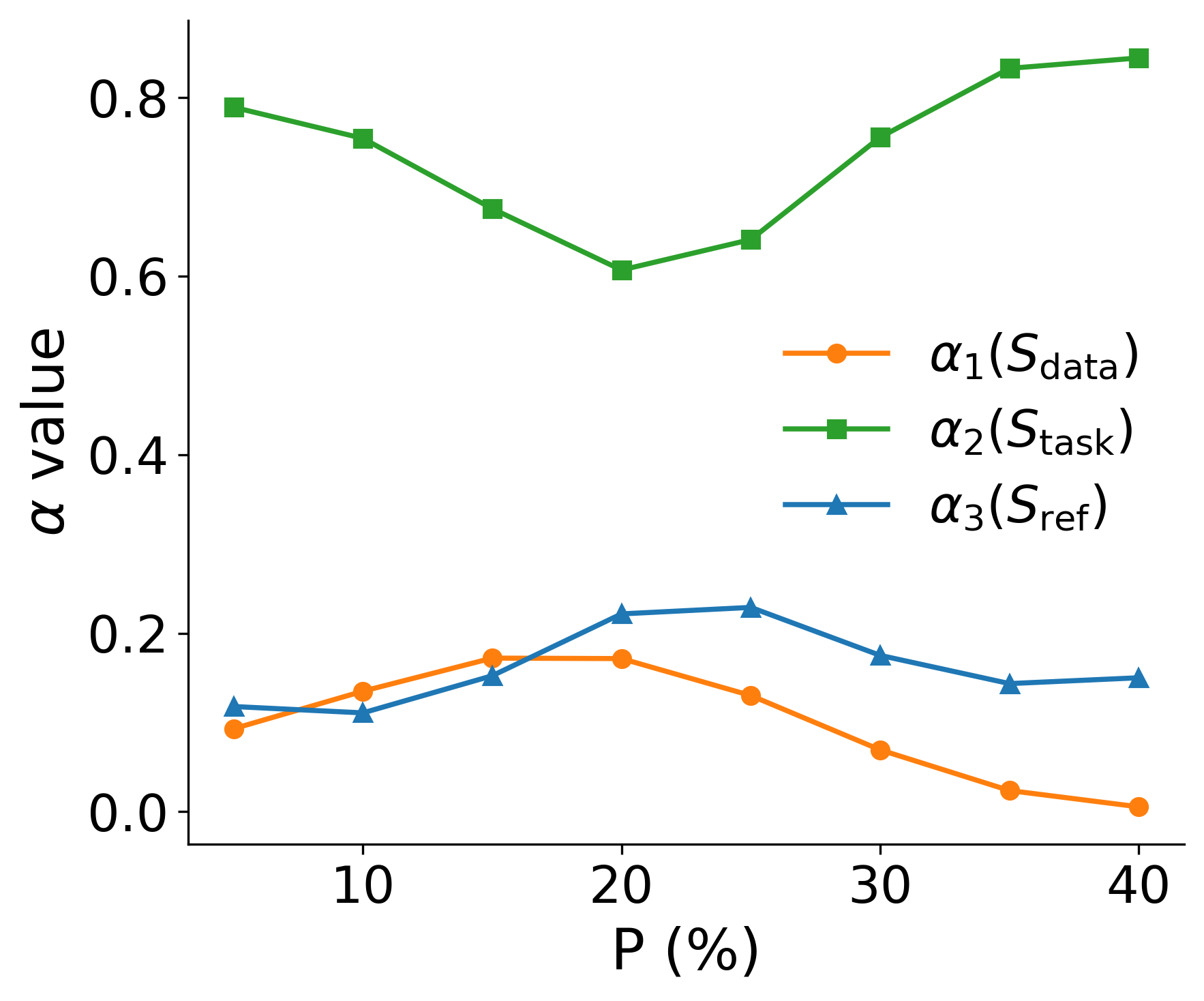}
        \caption{Smooth weights.}
        \label{fig:smooth_smart}
    \end{subfigure}%
    \hfill
    \begin{subfigure}{0.32\textwidth} 
        \centering
        \includegraphics[width=\linewidth]{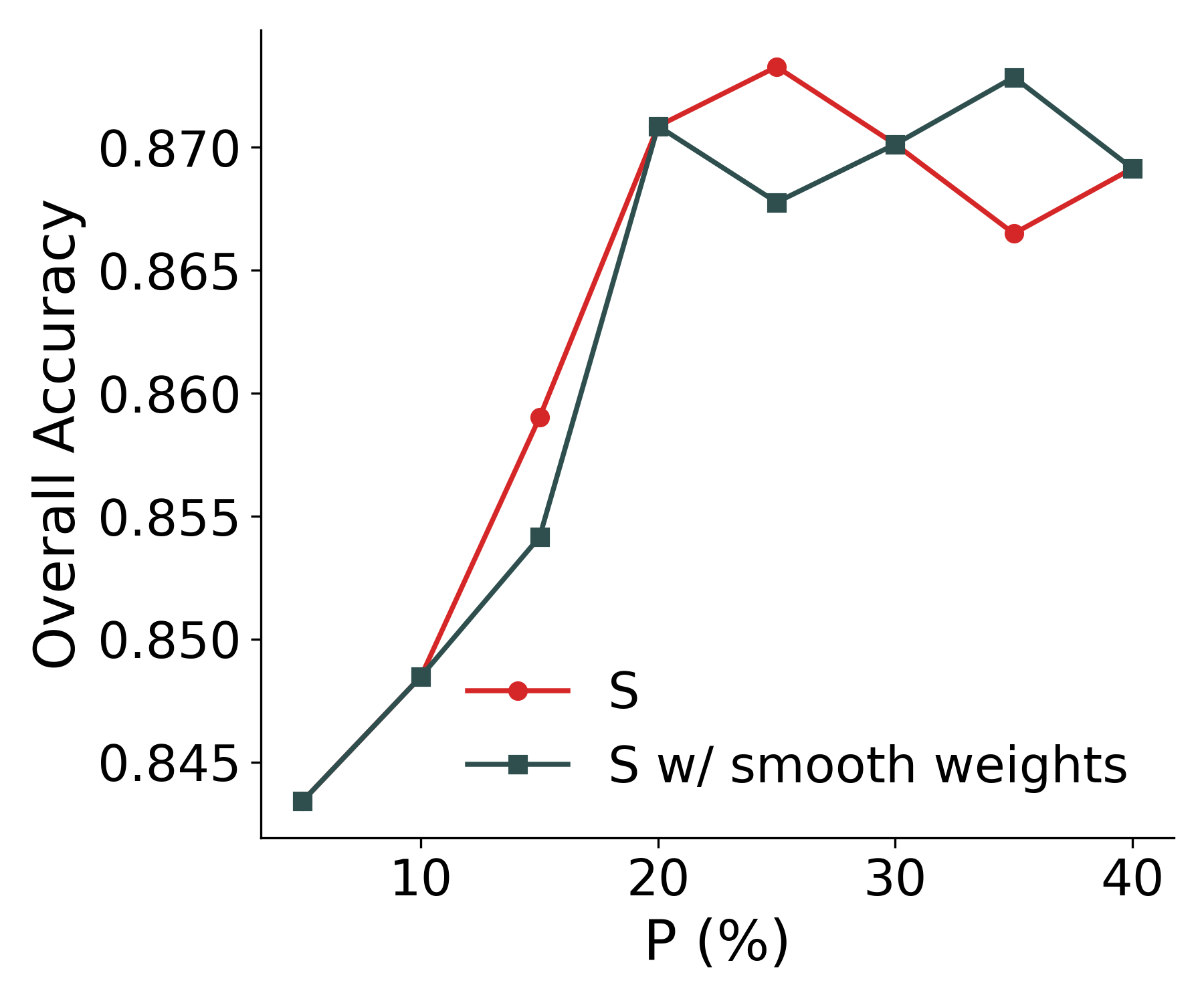}
        \caption{Overall accuracy.}
        \label{fig:acc_comp_smooth_smart}
    \end{subfigure}%
    \caption{Overall accuracy by integrating individual uncertainty scores using optimal weights and smooth weights in smart home.}
    \label{fig:acc_optimal_smooth_smart}
\end{figure*}

Section \ref{subsec6.1} introduces the optimization framework to obtain optimal $\boldsymbol{\alpha}$ for any given $P$. In practice, it is reasonable to assume that close levels of $P$ should share a similar $\boldsymbol{\alpha}$. Therefore, we can solve the optimization problem for a few levels of $P$ and obtain a few sampled points, then fit the whole trajectory of the optimal weights by smoothing. To test this idea, we set the levels of $P$ from 5\% to 40\%. Figure \ref{fig:optimal_smart} shows the optimal weights $\alpha_1$, $\alpha_2$, and $\alpha_3$ obtained for each $P$ via our cross-validation based optimization procedure. We then apply Gaussian kernel smoothing to these raw trajectories to smooth out the sampling variation and noise caused by the cross-validation in the optimization procedure and obtain the smoothed weights in Figure \ref{fig:smooth_smart}. Figure \ref{fig:acc_comp_smooth_smart} compares the overall accuracy of ALARM when using the unsmoothed optimal weights versus the smoothed weights. Across all $P$ values, the two curves are closely aligned, with only marginal differences—typically within 0.006 in accuracy, indicating that smoothed optimal weights as shown in Figure \ref{fig:smooth_smart} does not degrade performance. In fact, at small $P$, smoothing slightly improves accuracy, likely due to reduced overfitting of the optimal weights to fold-specific noise. Smooth weights offer a more robust and interpretable weighting scheme, and ALARM’s accuracy is largely insensitive to minor weight perturbations, underscoring the robustness of our UQ integration method. This experiment shows that in practice one can easily obtain all the optimal weights for a range of levels of $P$ and save much computational cost. 

\subsubsection{The Impact of the Number of MLLMs \texorpdfstring{$M$}{M}}
\begin{figure*}[!htbp]
    \centering
    \begin{subfigure}{0.32\textwidth} 
        \centering
        \includegraphics[width=\linewidth]{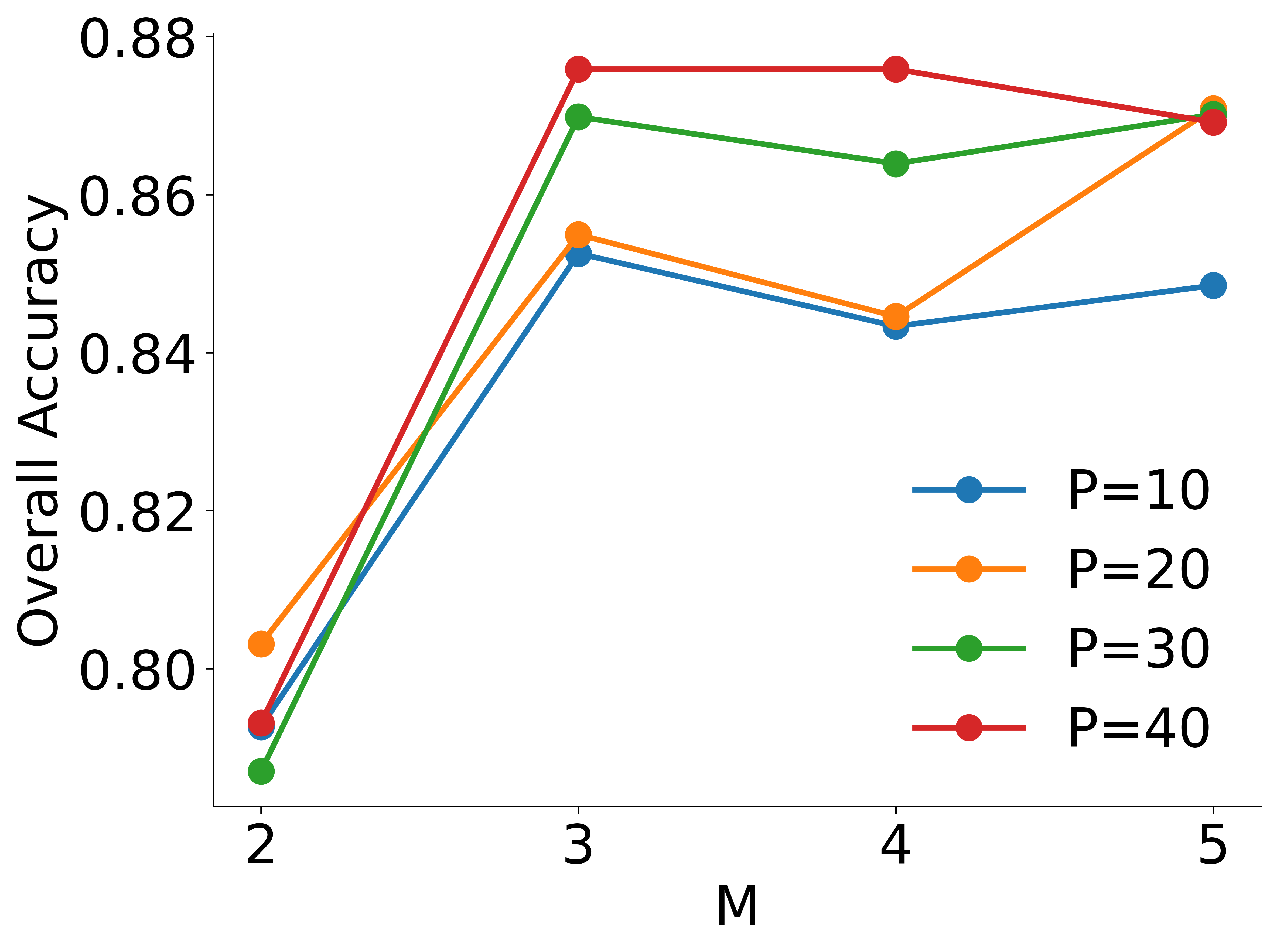}
        \caption{Smart home.}
        \label{fig:M_P_smart}
    \end{subfigure}%
    \hspace{0.001\textwidth}
    \begin{subfigure}{0.32\textwidth} 
        \centering
        \includegraphics[width=\linewidth]{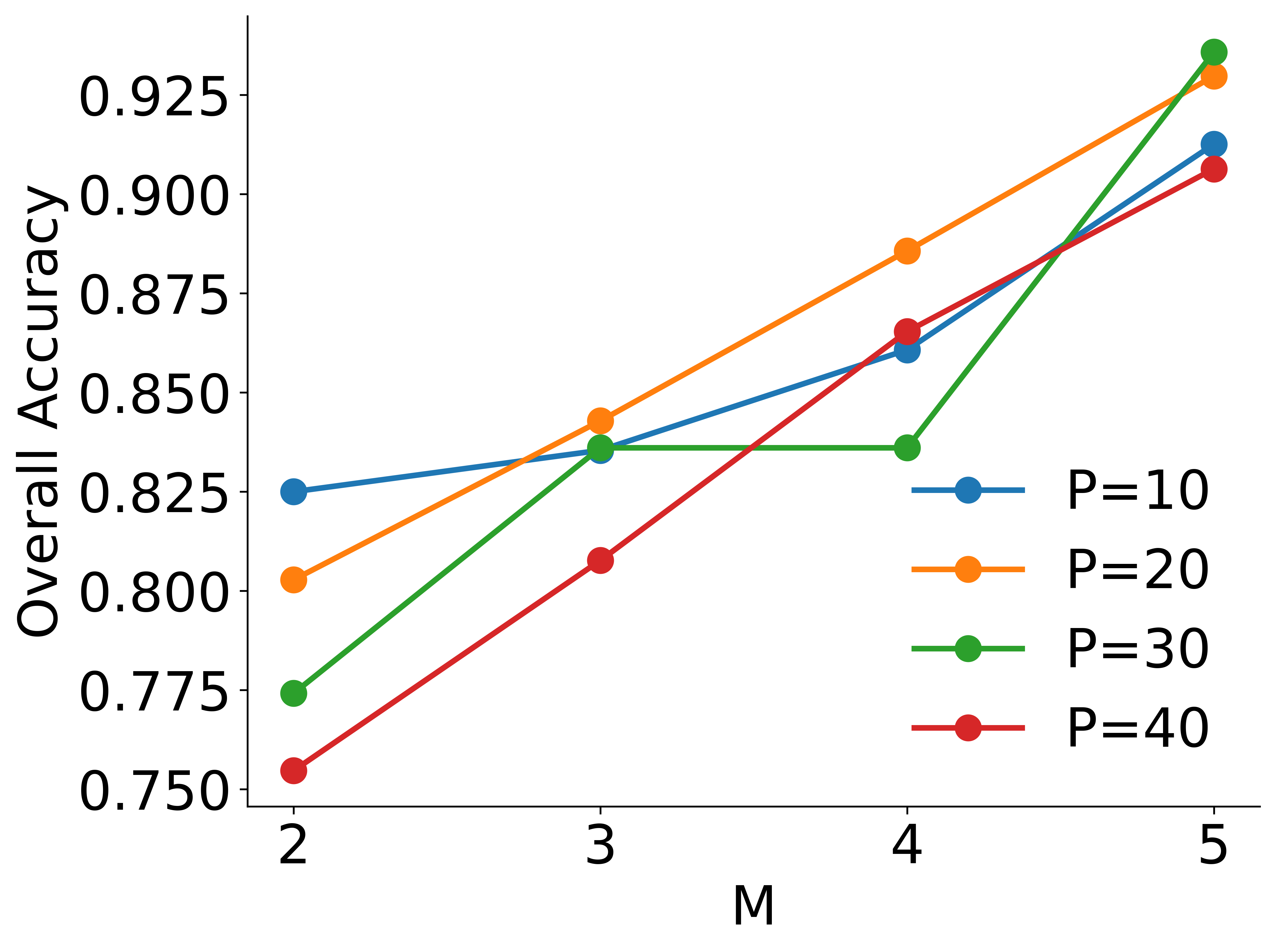}
        \caption{Wound classification.}
        \label{fig:M_P_wound}
    \end{subfigure}%
    \caption{Overall accuracy across varying rejection rates $P$ under different numbers of MLLMs $M$.}
    \label{fig:M_P}
\end{figure*}

In addition to the full MLLM ensemble with $M = 5$ described in Section \mbox{\ref{exp_setup}}, we vary $M$ to examine its effect on UQ quality and prediction performance. For the smart-home VAD, we evaluate MLLM ensembles of $M=2$ (Claude-3.5-sonnet and GPT-4o), $M=3$ (+Gemini-1.5-pro), $M=4$ (+Gemini-1.5-flash), and $M=5$ that includes all the 5 LLMs. We strategically design MLLM ensembles by including models from three major families (GPT, Claude, and Gemini) to ensure architectural diversity and reduce the risk of correlated errors. Figure \mbox{\ref{fig:M_P_smart}} shows the overall accuracy across varying $P$ under different values of $M$. Regardless of the rejection rate $P$, accuracy increases by about 6-8\% from $M=2$ to $M=3$, after which performance reaches a plateau. This suggests that using at least three MLLMs is essential for ALARM’s UQ mechanism to operate effectively, as this configuration captures most of the uncertainty information in the smart-home domain, while additional MLLMs provide marginal gains.

% \subsubsection{Determine the Optimal $P$ and the Impact of Unit Cost $\lambda$}\label{optimal_P_unit_cost}
\subsubsection{Determine the Optimal \texorpdfstring{$P$}{P} and the Impact of Unit Cost \texorpdfstring{$\lambda$}{lambda}}
\label{optimal_P_unit_cost}

\begin{figure*}[!htbp]
    \centering
    \begin{subfigure}{0.32\textwidth} 
        \centering
        \includegraphics[width=\linewidth]{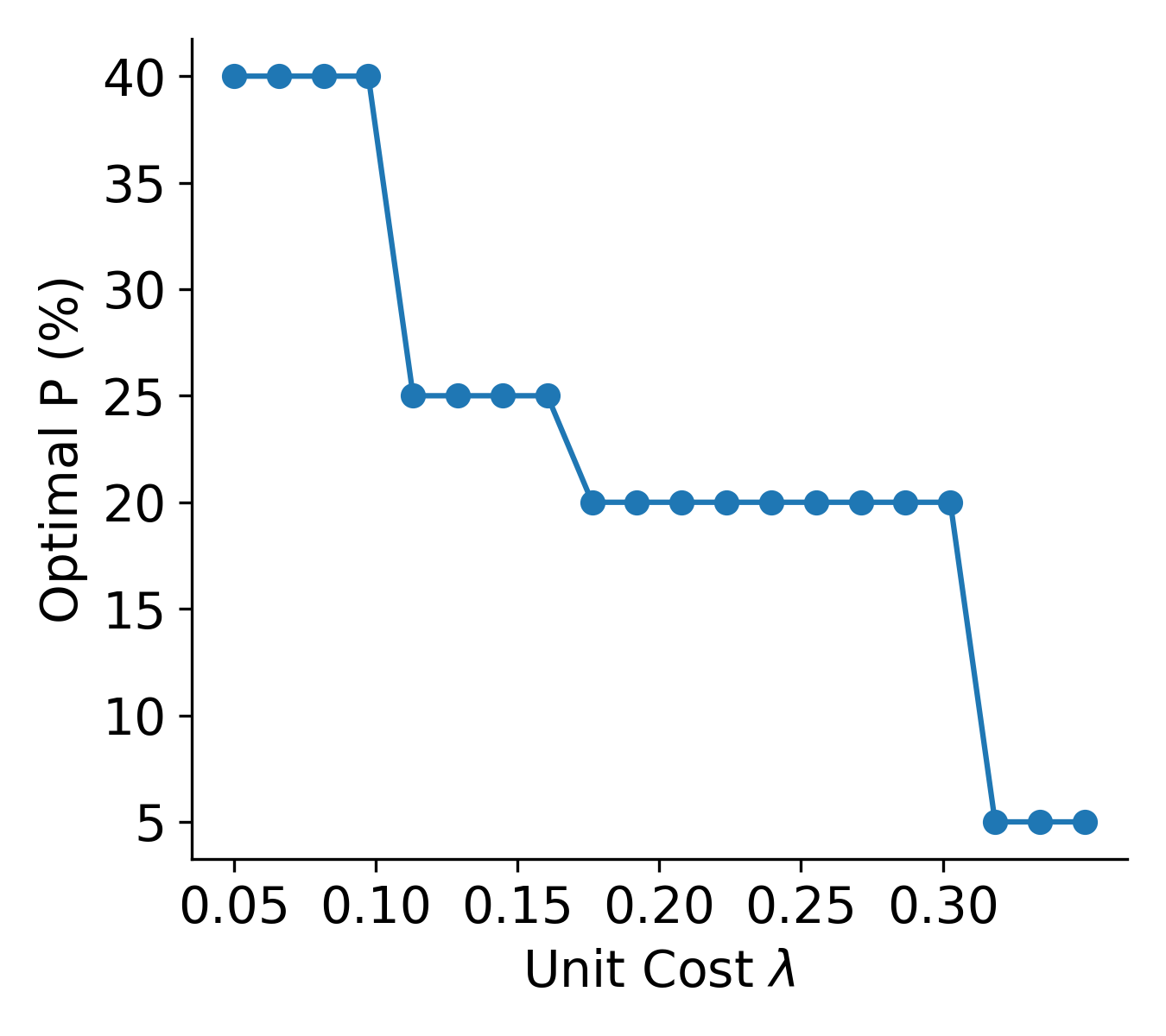}
        \caption{Smart home.}
        \label{fig:best_P_vs_c_smart}
    \end{subfigure}%
    \hspace{0.001\textwidth}
    \begin{subfigure}{0.32\textwidth} 
        \centering
        \includegraphics[width=\linewidth]{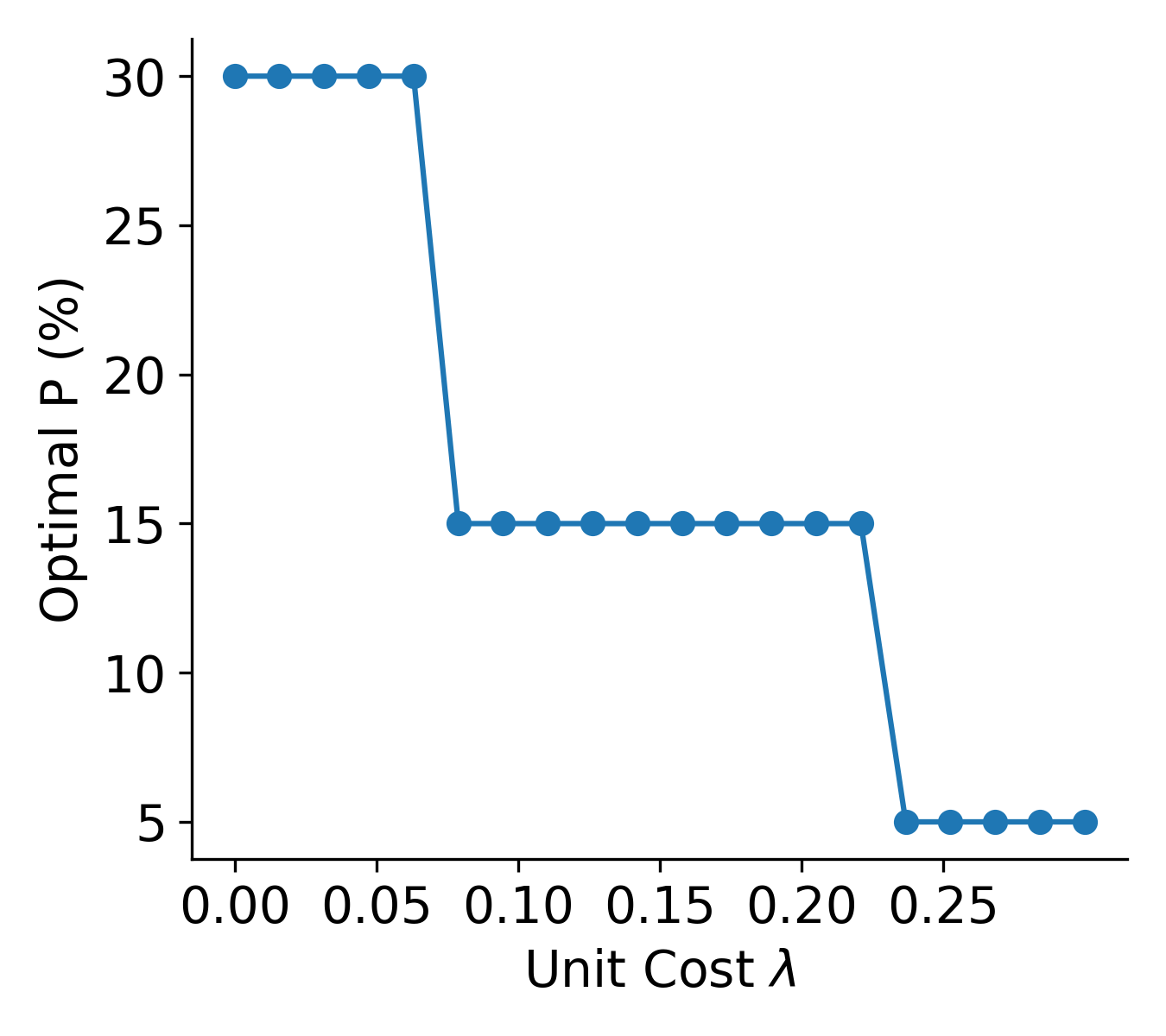}
        \caption{Wound classification.}
        \label{fig:P_cost_wound}
    \end{subfigure}%
    \caption{Optimal $P$ values under different unit cost $\lambda$ of human labor.}
    \label{fig:P_cost_both}
\end{figure*}

Figure \ref{fig:best_P_vs_c_smart} illustrates how the optimal rejection rate $P$ changes according to changes of cost ($\lambda$) of human expert, computed based on the optimization framework developed in Eq. \eqref{cost_opt}. The overall pattern is consistent with our hypothesis that higher cost leads to smaller $P$.

\subsubsection{Manual Inspection and Additional Insights}

\begin{figure}
\centering
\includegraphics[width=\textwidth]{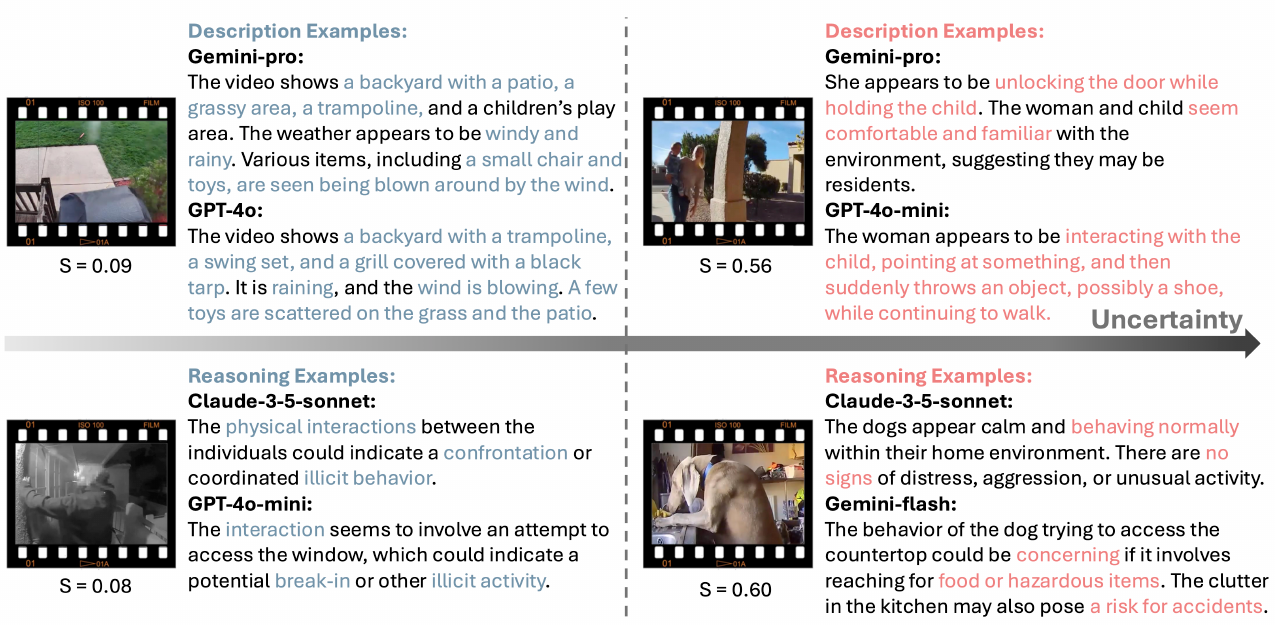}
\caption{Examples of video descriptions and reasoning for videos with low and high uncertainty scores.}
\label{fig:example_video_smart}
%\vspace{-1.5em}
\end{figure}

We further conduct manual inspection to better understand how our UQ scored the cases. Figure \ref{fig:example_video_smart} illustrates how Eq. \eqref{UQscore} provides a good measure of the unpredictability of a case. Low-uncertainty examples (left column) show that the five MLLMs give quite aligned descriptions and reasonings, i.e., consistent mention of ``trampoline'', ``windy'', or ``illicit behavior'', yielding a small uncertainty score $S$ = 0.09 and 0.08, respectively. High-uncertainty videos (right column) show divergent focuses (unlocking a door vs. interacting with a child) or contradictory rationales (normal calm dogs vs. potential hazard) among the MLLMs, driving $S>0.5$. Visually, the errant cases often involve ambiguous intent, causing divergence among the MLLMs in their decision-makings which can be effectively quantified by our UQ score.

\subsection{Wound Classification}

\subsubsection{Problem Setup}
\begin{figure}
\centering
\includegraphics[width=\textwidth]{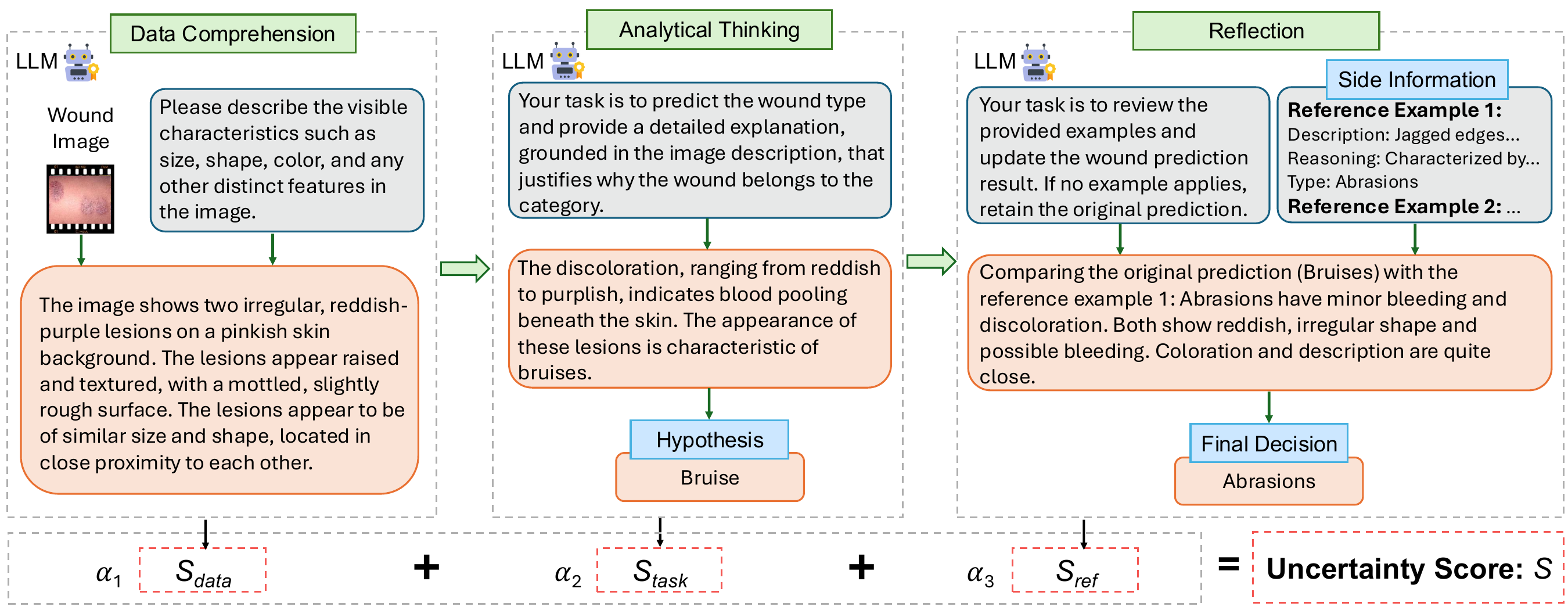}
%\vspace{-3mm}
\caption{Overview of LLM reasoning chain in ALARM for UQ in wound classification.}\label{fig:overview_wound}
%\vspace{-2mm}
\end{figure}

To demonstrate the generalizability of the ALARM framework, we further apply it to wound type classification that is known as a challenging problem in healthcare. The dataset\footnote{https://www.kaggle.com/datasets/yasinpratomo/wound-dataset} contains 432 wound images labeled with seven categories, including burns, abrasions, bruises, lacerations, cuts, stab wounds, and ingrown nails. Figure \ref{fig:overview_wound} illustrates how ALARM's three-stage reasoning chain, i.e., Data Comprehension, Analytical Thinking, and Reflection, is applied to this problem, with uncertainty quantified at each stage and integrated into a final score. Due to the space limit, we refer readers to more details of this case study in the Appendix.

\begin{table}[!htbp]
  \centering
  \caption{Performance of wound classification across five MLLMs in different methods.}
  \small % Reduce font size
  \begin{tabular}{lcccc}
    \toprule
    \textbf{Methods} & \textbf{Overall Accuracy} \\
    \midrule
    Zero Shot &  75.50 \\
    Chain-of-Thought & 79.44\\
    Few Shot &  79.49\\
    In-Context Learning & 79.48 \\
    Reasoning Chain w/o UQ & 79.72  \\
   Reasoning Chain w/ Random Drop & 88.53 \\
      Reasoning Chain w/ LAC & 89.60 \\
         Reasoning Chain w/ APS & 84.40 \\

               Reasoning Chain w/ ICL-EU & 88.10 \\
         Reasoning Chain w/ ICL-AU & 86.67 \\
    ALARM (\textbf{ours}) & \textbf{91.72} \\
    \bottomrule
  \end{tabular}
  %\vspace{-4mm}
  \label{tab:overall_perf_wound}
\end{table}

\subsubsection{Overall Evaluation}

\begin{figure}
\centering
\includegraphics[width=0.4\textwidth]{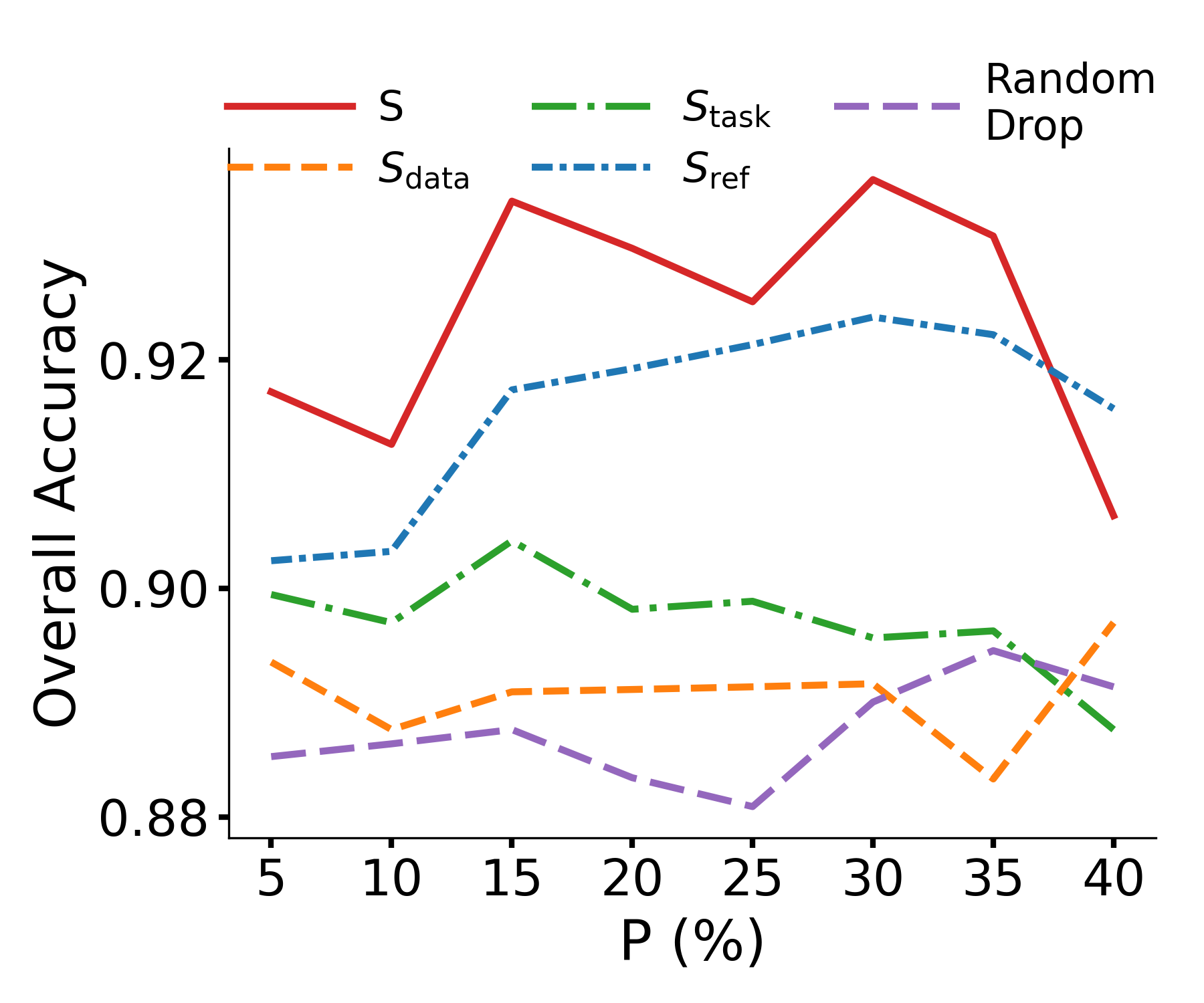}
%\vspace{-3mm}
\caption{Overall accuracy of different uncertainty scores in ALARM and random drop as the rejection rate $P$ varies in wound classification.}\label{fig:acc_P_wound}
%\vspace{-2mm}
\end{figure}

Results are shown in Table \ref{tab:overall_perf_wound}. The overall patterns we observe here are consistent with what we observed in the case study of smart-home dataset. Figure \ref{fig:acc_P_wound} also shows across most values of $P$, $S$ consistently achieves the highest accuracy, demonstrating the benefit of integrating multiple uncertainty sources. One notable difference from the smart-home case is that, here, Random Drop also yields a comparable performance with $S_{data}$, and is also close to $S_{task}$. This indicates a different structure of the two problems. We suspect that this actually means $S_{data}$ and $S_{task}$ are not as informative as $S_{ref}$, rather than saying Random Drop is magically as effective as an uncertainty score. Either case, we can see that the combined uncertainty score $S$ exhibits the strongest performance, indicating that though $S_{data}$ and $S_{task}$ are weak information individually, together they can add unique value to $S_{ref}$ and lead to a joint score $S$ that is better than its parts.

Figure \ref{fig:missclass_ratio_wound} indicates that at low rejection rates, ALARM’s UQ mechanism is highly effective, with the majority of rejected cases being truly misclassifications. In contrast, Random Drop rejects cases indiscriminately, resulting in a flat detection ratio across all levels of $P$ and substantially lower error-detection precision than ALARM. As $P$ increases, ALARM’s ratio of detected misclassifications declines gradually. This is expected: after the most uncertain cases are rejected, the remaining cases sent for rejection are progressively less likely to be wrong.

% \begin{figure}
% \centering
% \includegraphics[width=0.4\textwidth]{fig/missclass_ratio_stacked_wound.png}
% %\vspace{-3mm}
% \caption{The ratio of detected misclassification videos in the rejected videos in wound classification. \textcolor{red}{to save space you can also merge this figure with FIg 5}
% }\label{fig:missclass_ratio_wound}
% %\vspace{-2mm}
% \end{figure}  

\subsubsection{The Trajectory of the Optimal Weights \texorpdfstring{$\boldsymbol{\alpha}$}{alpha}}

\begin{figure}[!htbp]
    \centering
    \begin{subfigure}{0.32\textwidth} 
        \centering
        \includegraphics[width=\linewidth]{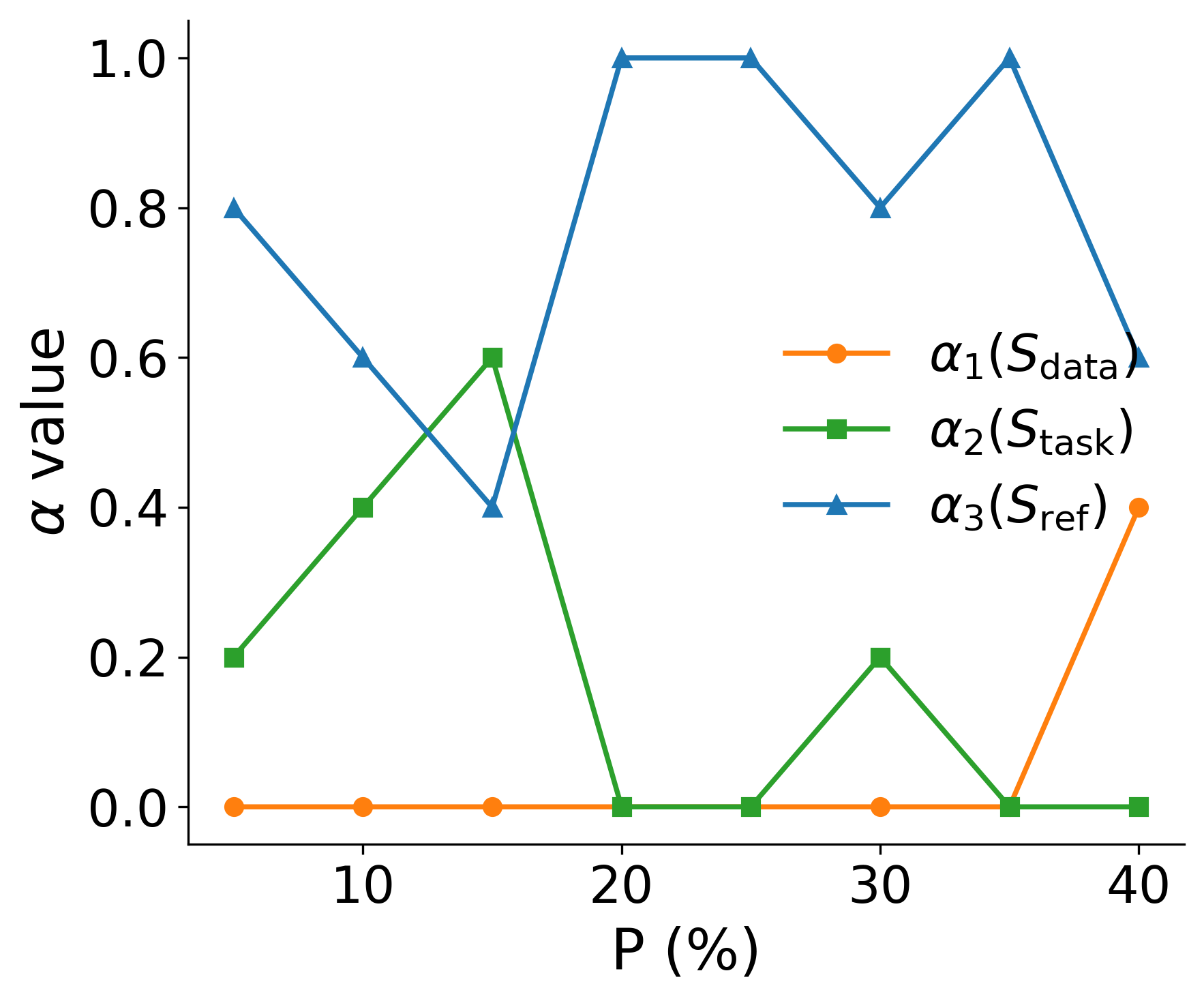}
        \caption{Optimal weights.}
        \label{fig:optimal_weight_wound}
    \end{subfigure}%
    \hfill
        \begin{subfigure}{0.32\textwidth} 
        \centering
        \includegraphics[width=\linewidth]{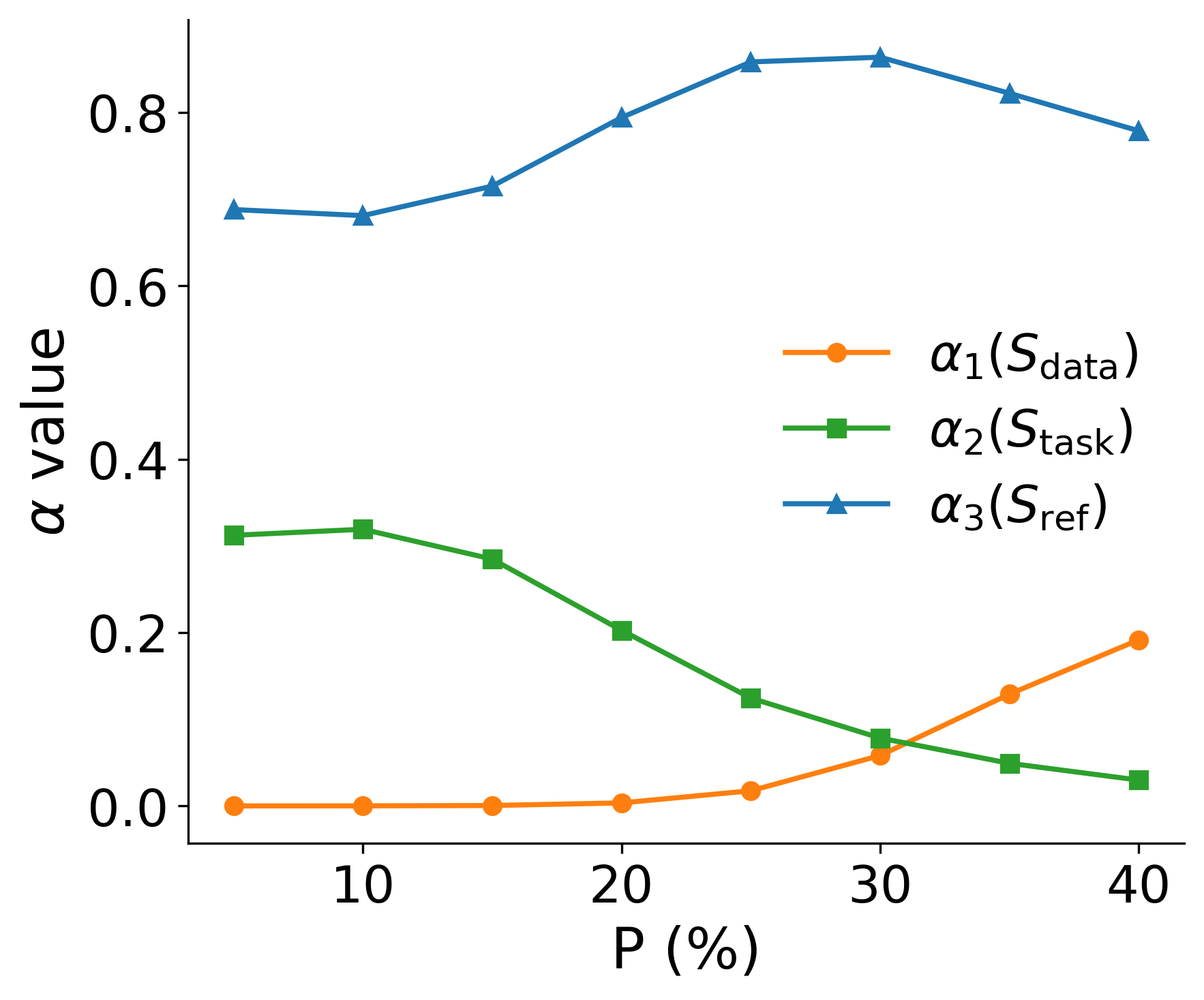}
        \caption{Smooth weights.}
        \label{fig:smooth_weight_wound}
    \end{subfigure}%
    \hfill
    \begin{subfigure}{0.32\textwidth} 
        \centering
        \includegraphics[width=\linewidth]{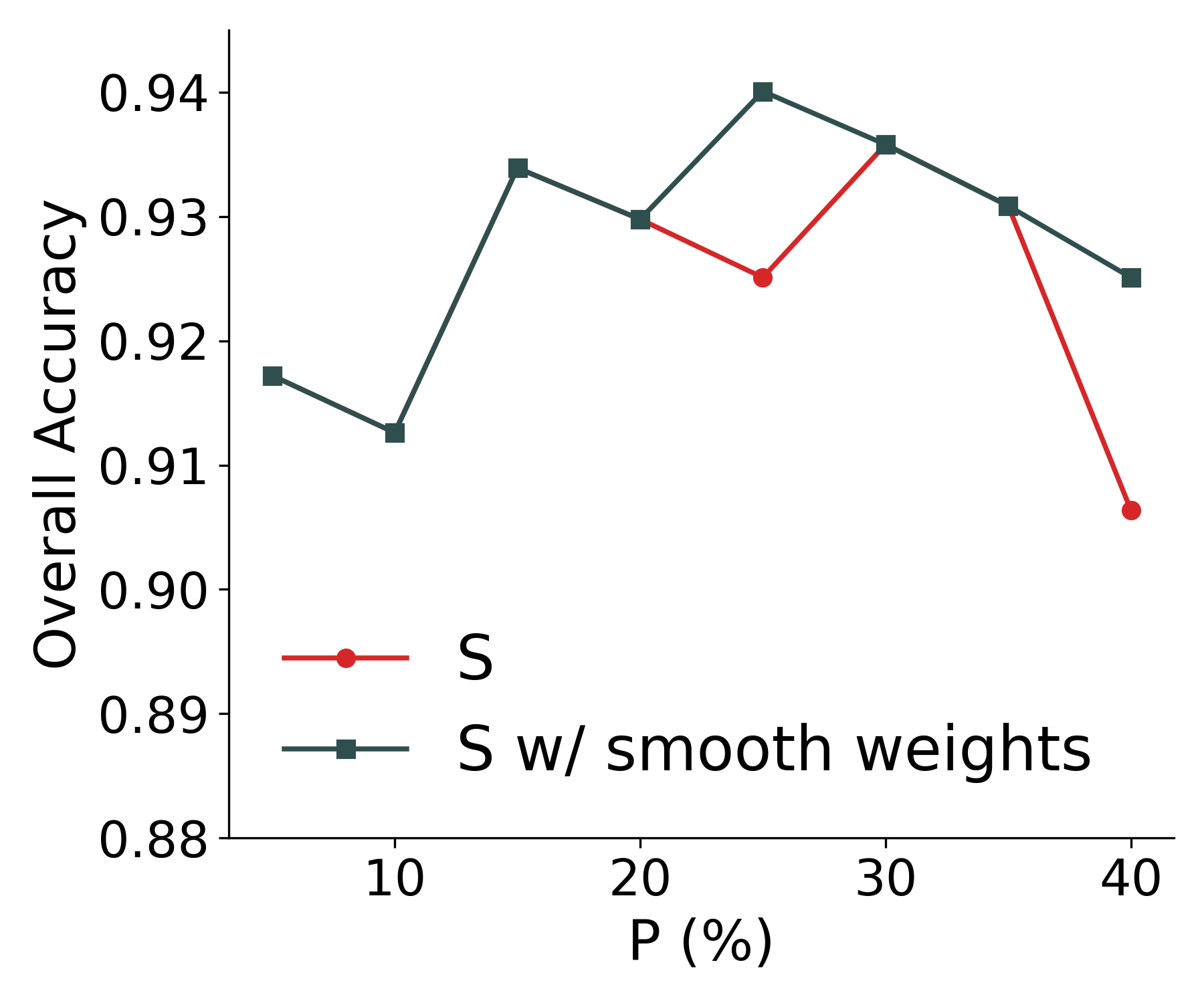}
        \caption{Overall accuracy.}
        \label{fig:acc_optimal_smooth_wound_comp}
    \end{subfigure}%
    \caption{Overall accuracy by integrating individual uncertainty scores using optimal weights and smooth weights in wound classification.}
    \label{fig:acc_optimal_smooth_wound}
\end{figure}

We repeat the smoothing strategy to generate the whole trajectory of the optimal weights. As shown in Figure \ref{fig:optimal_weight_wound}, the optimal weights fluctuate across $P$ values, reflecting sampling variation and optimization noise due to our cross-validation based optimization procedure to obtain the optimal weights for a given $P$. Figure \ref{fig:smooth_weight_wound} shows the smoothed trajectory, and Figure \ref{fig:acc_optimal_smooth_wound_comp} compares the overall accuracy using the unsmoothed weights versus the smooth weights. Across all $P$, the accuracy curves are closely aligned, indicating that smoothing does not degrade performance. In fact, smoothing slightly improves accuracy at certain $P$ values, likely due to its denoising capacity. These results are consistent with what we observed in the smart-home case and suggest that ALARM’s uncertainty integration is robust to minor perturbations in weight selection. It is interesting to see $\alpha_3$ (Reflection) dominates the other two weights at most levels of $P$, while $\alpha_2$ (Analytical Thinking) contributes more at low $P$, and $\alpha_1$ (Data Comprehension) plays a role mainly at the largest $P$. This is also consistent with our observation on Figure \ref{fig:acc_P_wound} that $S_{ref}$ is the most informative uncertainty score for this case. 

\subsubsection{The Impact of the Number of MLLMs \texorpdfstring{$M$}{M}}
For the wound classification, we evaluate MLLM ensembles of $M=2$ (Claude-3.5-haiku and GPT-4o-mini), $M=3$ (+Claude-3.5-sonnet), $M=4$ (+Claude-3.7-sonnet), and $M=5$ that includes all the 5 LLMs. Figure \mbox{\ref{fig:M_P_wound}} presents the relationship between MLLM ensemble size $M$ and overall accuracy across different $P$. Increasing the ensemble size $M$ consistently improves overall accuracy across nearly all rejection rate $P$. In wound classification, as $M$ grows, ALARM obtains a richer and more complete set of interpretations of the same image, reducing blind spots that any single model may have.

% Wound classification shows high overall accuracy ranging from 0.90 to 0.94 at $M=2$, followed by a gradual decline at $M=3$ and $M=4$, and then a subsequent recovery at $M=5$. This pattern suggests that the two base MLLMs (Claude-3.5-sonnet and GPT-4o) already provide a strong and complementary foundation for this task at $M=2$. When these two MLLMs align on clear-cut cases and appropriately flag ambiguous ones, adding additional MLLMs that introduce redundant or spurious disagreement can degrade UQ quality by injecting noise into the ensemble’s uncertainty signal. The configuration at $M=4$, dominated by three Claude MLLMs and a single GPT, illustrates this effect due to limited architectural diversity. When diversity is reintroduced, i.e., by adding GPT-4o-mini at $M=5$, ALARM’s performance recovers, demonstrating the framework’s ability to maintain robustness and recalibrate uncertainty estimation once sufficient heterogeneity is restored in the ensemble.}

% \subsubsection{Determine the Optimal $P$ and the Impact of Unit Cost $\lambda$}
\subsubsection{Determine the Optimal \texorpdfstring{$P$}{P} and the Impact of Unit Cost \texorpdfstring{$\lambda$}{lambda}}

Figure \ref{fig:P_cost_wound} shows the optimal solutions of Eq. \eqref{cost_opt} when given different hypothetic values of unit cost $\lambda$. In practice, this figure can help us decide what is the optimal value of $P$ (and correspondingly, leading to the optimal weights of $S$) for a given $\lambda$. As shown in Figure \ref{fig:P_cost_wound}, when unit cost $\lambda \leq 0.07$, the optimal strategy is to defer approximately 30\% of cases, maximizing accuracy gains from human review at minimal cost. As $\lambda$ increases to around 0.08, $P$ sharply decreases to 15\%, reflecting a more conservative allocation of manual labeling resources. A further rise triggers another abrupt drop to 5\%, where only the most uncertain cases are sent for expert evaluation. 

\section{Conclusion}
LLMs have given rise to the promising prospect of developing smart AI agents that can conduct open-ended anomaly detection in complex environments. The challenge is that in these complex environments, the anomalies are sometimes highly contextual and also ambiguous, and thereby, UQ is a crucial capacity for an MLLM-based VAD system to succeed. In this paper, we use the application of complex-environment monitoring as a setting to introduce our UQ-supported MLLM-based VAD framework called ALARM. ALARM integrates UQ with quality-assurance techniques like reasoning chain, self-reflection, and MLLM ensemble for robust and accurate performance and is designed based on a rigorous probabilistic inference pipeline and computational process. The integrated UQ scores $S$ consistently achieves the most stable improvement across datasets and rejection levels. This is because $S_{data}$, $S_{task}$, and $S_{ref}$ capture complementary uncertainty sources from Data Comprehension, Analytical Thinking, and Reflection. In general, in real-world deployment, the importance of a particular UQ source is unknown and can vary across tasks, datasets, or even implementation contexts; hence, relying on a single component may lead to unstable or biased uncertainty estimation. The optimized combination ensures a balanced UQ measure that generalizes across domains, providing reliable abstention behavior even when one component underperforms. For this reason, ALARM's generic UQ could be used for many other LLM-based decision-making frameworks which are multi-staged and consists of a sequential information processing pipeline. For example, in financial risk assessment, during the Data Comprehension stage, LLMs can summarize information from transaction logs, client profiles, and market reports. In the Analytical Thinking stage, the models reason over the summaries to infer credit risk, detect potential fraud, or identify investment anomalies based on historical patterns and statistical indicators. Finally, in the Reflection stage, domain-specific rules, such as regulatory thresholds or compliance constraints, serve as side information that allows the LLMs to refine their risk predictions. In future work, we will expand ALARM to a more comprehensive scope of complex-environment monitoring that covers not only video and image monitoring data but a whole range of other sensory data as well. MLLMs offer powerful new capabilities for interpreting and acting on these multimodal data. Recent analyses note that IoT sensors effectively serve as ``sensing body parts'' for LLMs, feeding real-time data that helps the model ground its outputs and generate more accurate, context-aware responses \citep{sun2024ai,chen2011web,bui2012home}. Monitoring the data from network traffic, temperature, humidity, audio, and motion together is beneficial for detecting anomalies and figure out the anomaly sources. Another direction is to expand dataset domains. Although SmartHome-Bench provides the benchmark for smart-home monitoring, its scale and diversity remain limited compared with large-scale VAD datasets in other domains. Future work could leverage transfer learning \citep{d2019transfer,pan2023transfer} by incorporating other VAD datasets beyond smart homes (e.g., surveillance datasets on public spaces \citep{liu2018future}). This would allow ALARM to capture broader anomaly patterns. In addition, synthetic data generation (e.g., via video diffusion models \citep{ho2022video}) could further expand the training pool, providing richer coverage of ambiguous or rare events. These approaches would help mitigate data scarcity and support more robust deployment of ALARM in diverse real-world environments. Last but not least, in this paper, we discuss the parameter configuration of ALARM within the scope of UQ-enabled quality assurance framework that assumes ALARM can defer some cases to human expert. In future work, we will also fully develop these promising prospects of ALARM in terms of human-AI collaboration. Different LLMs, shaped by their respective training data, may exhibit distinct strengths across various domains or categories. By systematically comparing human responses on data with the outputs of different LLMs, we can identify each model’s domain-specific expertise and selectively deploy them on tasks where they perform best.

%The common sense knowledge encoded in LLMs and their reasoning capability hold great promise in addressing these challenges.  (e.g. temperature, motion, audio)

%\THEEndNotes
% \begingroup \parindent 0pt \parskip 0.0ex \def\enotesize{\normalsize} \theendnotes \endgroup

% Appendix here
% Options are (1) APPENDIX (with or without general title) or
%             (2) APPENDICES (if it has more than one unrelated sections)
% Outcomment the appropriate case if necessary
%
% \begin{APPENDIX}{<Title of the Appendix>}
% \end{APPENDIX}
%
%   or
%
% \begin{APPENDICES}
% \section{<Title of Section A>}
% \section{<Title of Section B>}
% etc
% \end{APPENDICES}

% Acknowledgments here
% \ACKNOWLEDGMENT{We would like to express our sincere gratitude to [acknowledge individuals, organizations, or institutions] for their invaluable contributions to this research. We are also grateful to [mention any additional acknowledgements, such as technical assistance, data providers, or colleagues] for their support and assistance throughout the course of this work.}

% References here (outcomment the appropriate case)

% CASE 1: BiBTeX used to constantly update the references
%   (while the paper is being written).
\bibliographystyle{informs2014} % outcomment this and next line in Case 1
\bibliography{sample} % if more than one, comma separated

@inproceedings{geifman2019selectivenet,
  title={Selectivenet: A deep neural network with an integrated reject option},
  author={Geifman, Yonatan and El-Yaniv, Ran},
  booktitle={International conference on machine learning},
  pages={2151--2159},
  year={2019},
  organization={PMLR}
}

@article{alves2025benchmarking,
  title={A benchmarking framework and dataset for learning to defer in human-AI decision-making},
  author={Alves, Jean V and Leit{\~a}o, Diogo and Jesus, S{\'e}rgio and Sampaio, Marco OP and Li{\'e}bana, Javier and Saleiro, Pedro and Figueiredo, M{\'a}rio AT and Bizarro, Pedro},
  journal={Scientific data},
  volume={12},
  number={1},
  pages={506},
  year={2025},
  publisher={Nature Publishing Group UK London}
}

@article{bharadwaj2024vane,
  title={VANE-Bench: Video Anomaly Evaluation Benchmark for Conversational LMMs},
  author={Bharadwaj, Rohit and Gani, Hanan and Naseer, Muzammal and Khan, Fahad Shahbaz and Khan, Salman},
  journal={arXiv preprint arXiv:2406.10326},
  year={2024}
}

@article{mnih2007probabilistic,
  title={Probabilistic matrix factorization},
  author={Mnih, Andriy and Salakhutdinov, Russ R},
  journal={Advances in neural information processing systems},
  volume={20},
  year={2007}
}

@article{scikit-learn,
  title={Scikit-learn: Machine Learning in {P}ython},
  author={Pedregosa, F. and Varoquaux, G. and Gramfort, A. and Michel, V.
          and Thirion, B. and Grisel, O. and Blondel, M. and Prettenhofer, P.
          and Weiss, R. and Dubourg, V. and Vanderplas, J. and Passos, A. and
          Cournapeau, D. and Brucher, M. and Perrot, M. and Duchesnay, E.},
  journal={Journal of Machine Learning Research},
  volume={12},
  pages={2825--2830},
  year={2011}
}

@InProceedings{Zhao_2025_CVPR,
    author    = {Zhao, Xinyi and Zhang, Congjing and Guo, Pei and Li, Wei and Chen, Lin and Zhao, Chaoyue and Huang, Shuai},
    title     = {SmartHome-Bench: A Comprehensive Benchmark for Video Anomaly Detection in Smart Homes Using Multi-Modal Large Language Models},
    booktitle = {Proceedings of the Computer Vision and Pattern Recognition Conference (CVPR) Workshops},
    month     = {June},
    year      = {2025},
    pages     = {3975-3985}
}

@article{geifman2017selective,
  title={Selective classification for deep neural networks},
  author={Geifman, Yonatan and El-Yaniv, Ran},
  journal={Advances in neural information processing systems},
  volume={30},
  year={2017}
}

@article{sadinle2019least,
  title={Least ambiguous set-valued classifiers with bounded error levels},
  author={Sadinle, Mauricio and Lei, Jing and Wasserman, Larry},
  journal={Journal of the American Statistical Association},
  volume={114},
  number={525},
  pages={223--234},
  year={2019},
  publisher={Taylor \& Francis}
}

@article{madras2018predict,
  title={Predict responsibly: improving fairness and accuracy by learning to defer},
  author={Madras, David and Pitassi, Toni and Zemel, Richard},
  journal={Advances in neural information processing systems},
  volume={31},
  year={2018}
}

@article{vyas2025autonomous,
  title={Autonomous Control Leveraging LLMs: An Agentic Framework for Next-Generation Industrial Automation},
  author={Vyas, Javal and Mercangoz, Mehmet},
  journal={arXiv preprint arXiv:2507.07115},
  year={2025}
}

@article{yuan2024r,
  title={R-judge: Benchmarking safety risk awareness for llm agents},
  author={Yuan, Tongxin and He, Zhiwei and Dong, Lingzhong and Wang, Yiming and Zhao, Ruijie and Xia, Tian and Xu, Lizhen and Zhou, Binglin and Li, Fangqi and Zhang, Zhuosheng and others},
  journal={arXiv preprint arXiv:2401.10019},
  year={2024}
}

@article{mehandru2025er,
  title={ER-REASON: A Benchmark Dataset for LLM-Based Clinical Reasoning in the Emergency Room},
  author={Mehandru, Nikita and Golchini, Niloufar and Bamman, David and Zack, Travis and Molina, Melanie F and Alaa, Ahmed},
  journal={arXiv preprint arXiv:2505.22919},
  year={2025}
}

@inproceedings{zhu2024llms,
  title={Do LLMs Understand Visual Anomalies? Uncovering LLM's Capabilities in Zero-shot Anomaly Detection},
  author={Zhu, Jiaqi and Cai, Shaofeng and Deng, Fang and Ooi, Beng Chin and Wu, Junran},
  booktitle={Proceedings of the 32nd ACM International Conference on Multimedia},
  pages={48--57},
  year={2024}
}

@inproceedings{ionescu2019object,
  title={Object-centric auto-encoders and dummy anomalies for abnormal event detection in video},
  author={Ionescu, Radu Tudor and Khan, Fahad Shahbaz and Georgescu, Mariana-Iuliana and Shao, Ling},
  booktitle={Proceedings of the IEEE/CVF conference on computer vision and pattern recognition},
  pages={7842--7851},
  year={2019}
}

@inproceedings{hasan2016learning,
  title={Learning temporal regularity in video sequences},
  author={Hasan, Mahmudul and Choi, Jonghyun and Neumann, Jan and Roy-Chowdhury, Amit K and Davis, Larry S},
  booktitle={Proceedings of the IEEE conference on computer vision and pattern recognition},
  pages={733--742},
  year={2016}
}

@article{ling2024uncertainty,
  title={Uncertainty quantification for in-context learning of large language models},
  author={Ling, Chen and Zhao, Xujiang and Zhang, Xuchao and Cheng, Wei and Liu, Yanchi and Sun, Yiyou and Oishi, Mika and Osaki, Takao and Matsuda, Katsushi and Ji, Jie and others},
  journal={arXiv preprint arXiv:2402.10189},
  year={2024}
}

@article{romano2020classification,
  title={Classification with valid and adaptive coverage},
  author={Romano, Yaniv and Sesia, Matteo and Candes, Emmanuel},
  journal={Advances in neural information processing systems},
  volume={33},
  pages={3581--3591},
  year={2020}
}

@article{ye2024benchmarking,
  title={Benchmarking llms via uncertainty quantification},
  author={Ye, Fanghua and Yang, Mingming and Pang, Jianhui and Wang, Longyue and Wong, Derek and Yilmaz, Emine and Shi, Shuming and Tu, Zhaopeng},
  journal={Advances in Neural Information Processing Systems},
  volume={37},
  pages={15356--15385},
  year={2024}
}

@article{xu2024customizing,
  title={Customizing Visual-Language Foundation Models for Multi-modal Anomaly Detection and Reasoning},
  author={Xu, Xiaohao and Cao, Yunkang and Chen, Yongqi and Shen, Weiming and Huang, Xiaonan},
  journal={arXiv preprint arXiv:2403.11083},
  year={2024}
}

@inproceedings{liu2018future,
  title={Future frame prediction for anomaly detection--a new baseline},
  author={Liu, Wen and Luo, Weixin and Lian, Dongze and Gao, Shenghua},
  booktitle={Proceedings of the IEEE conference on computer vision and pattern recognition},
  pages={6536--6545},
  year={2018}
}

@article{lv2024video,
  title={Video anomaly detection and explanation via large language models},
  author={Lv, Hui and Sun, Qianru},
  journal={arXiv preprint arXiv:2401.05702},
  year={2024}
}

@article{zhang2024holmes,
  title={Holmes-VAD: Towards Unbiased and Explainable Video Anomaly Detection via Multi-modal LLM},
  author={Zhang, Huaxin and Xu, Xiaohao and Wang, Xiang and Zuo, Jialong and Han, Chuchu and Huang, Xiaonan and Gao, Changxin and Wang, Yuehuan and Sang, Nong},
  journal={arXiv preprint arXiv:2406.12235},
  year={2024}
}

@article{park2024artificial,
  title={Artificial intelligence-based video monitoring of movement disorders in the elderly: a review on current and future landscapes},
  author={Park, Kye Won and Mirian, Maryam S and McKeown, Martin J},
  journal={Singapore Medical Journal},
  volume={65},
  number={3},
  pages={141--149},
  year={2024},
  publisher={Medknow}
}

@inproceedings{lopes2021contactless,
  title={Contactless Smart Screening in Nursing Homes: an IoT-enabled solution for the COVID-19 era},
  author={Lopes, S{\'e}rgio Ivan and Pinho, Pedro and Marques, Paulo and Abreu, Carlos and Carvalho, Nuno B and Ferreira, Jos{\'e}},
  booktitle={2021 17th International Conference on Wireless and Mobile Computing, Networking and Communications (WiMob)},
  pages={145--150},
  year={2021},
  organization={IEEE}
}

@article{inan2023llama,
  title={Llama guard: Llm-based input-output safeguard for human-ai conversations},
  author={Inan, Hakan and Upasani, Kartikeya and Chi, Jianfeng and Rungta, Rashi and Iyer, Krithika and Mao, Yuning and Tontchev, Michael and Hu, Qing and Fuller, Brian and Testuggine, Davide and others},
  journal={arXiv preprint arXiv:2312.06674},
  year={2023}
}

@article{wei2022chain,
  title={Chain-of-thought prompting elicits reasoning in large language models},
  author={Wei, Jason and Wang, Xuezhi and Schuurmans, Dale and Bosma, Maarten and Xia, Fei and Chi, Ed and Le, Quoc V and Zhou, Denny and others},
  journal={Advances in neural information processing systems},
  volume={35},
  pages={24824--24837},
  year={2022}
}

@article{ho2022video,
  title={Video diffusion models},
  author={Ho, Jonathan and Salimans, Tim and Gritsenko, Alexey and Chan, William and Norouzi, Mohammad and Fleet, David J},
  journal={Advances in neural information processing systems},
  volume={35},
  pages={8633--8646},
  year={2022}
}

@article{pan2023transfer,
  title={Transfer learning-based data anomaly detection for structural health monitoring},
  author={Pan, Qiuyue and Bao, Yuequan and Li, Hui},
  journal={Structural Health Monitoring},
  volume={22},
  number={5},
  pages={3077--3091},
  year={2023},
  publisher={SAGE Publications Sage UK: London, England}
}

@article{d2019transfer,
  title={Transfer learning for non-intrusive load monitoring},
  author={D’Incecco, Michele and Squartini, Stefano and Zhong, Mingjun},
  journal={IEEE Transactions on Smart Grid},
  volume={11},
  number={2},
  pages={1419--1429},
  year={2019},
  publisher={IEEE}
}

@article{wang2022self,
  title={Self-consistency improves chain of thought reasoning in language models},
  author={Wang, Xuezhi and Wei, Jason and Schuurmans, Dale and Le, Quoc and Chi, Ed and Narang, Sharan and Chowdhery, Aakanksha and Zhou, Denny},
  journal={arXiv preprint arXiv:2203.11171},
  year={2022}
}

@article{liu2023pre,
  title={Pre-train, prompt, and predict: A systematic survey of prompting methods in natural language processing},
  author={Liu, Pengfei and Yuan, Weizhe and Fu, Jinlan and Jiang, Zhengbao and Hayashi, Hiroaki and Neubig, Graham},
  journal={ACM computing surveys},
  volume={55},
  number={9},
  pages={1--35},
  year={2023},
  publisher={ACM New York, NY}
}

@article{malone2022challenges,
  title={Challenges in the diagnosis and management of wound infection},
  author={Malone, Matthew and Schultz, Gregory},
  journal={British journal of dermatology},
  volume={187},
  number={2},
  pages={159--166},
  year={2022},
  publisher={Blackwell Publishing Ltd Oxford, UK}
}

@article{stojkoska2017review,
  title={A review of Internet of Things for smart home: Challenges and solutions},
  author={Stojkoska, Biljana L Risteska and Trivodaliev, Kire V},
  journal={Journal of cleaner production},
  volume={140},
  pages={1454--1464},
  year={2017},
  publisher={Elsevier}
}

@article{patel2009monitoring,
  title={Monitoring motor fluctuations in patients with Parkinson's disease using wearable sensors},
  author={Patel, Shyamal and Lorincz, Konrad and Hughes, Richard and Huggins, Nancy and Growdon, John and Standaert, David and Akay, Metin and Dy, Jennifer and Welsh, Matt and Bonato, Paolo},
  journal={IEEE transactions on information technology in biomedicine},
  volume={13},
  number={6},
  pages={864--873},
  year={2009},
  publisher={IEEE}
}

@article{chen2011web,
  title={Web-based remote human pulse monitoring system with intelligent data analysis for home health care},
  author={Chen, Chih-Ming},
  journal={Expert systems with Applications},
  volume={38},
  number={3},
  year={2011},
  publisher={Elsevier}
}

@article{bui2012home,
  title={Home monitoring for heart failure management},
  author={Bui, Anh L and Fonarow, Gregg C},
  journal={Journal of the American College of Cardiology},
  volume={59},
  number={2},
  pages={97--104},
  year={2012},
  publisher={American College of Cardiology Foundation Washington, DC}
}

@article{sun2024ai,
  title={An ai-based system utilizing iot-enabled ambient sensors and llms for complex activity tracking},
  author={Sun, Yuan and Ortiz, Jorge},
  journal={arXiv preprint arXiv:2407.02606},
  year={2024}
}

@article{song2025inv,
  title={Inv-Entropy: A Fully Probabilistic Framework for Uncertainty Quantification in Language Models},
  author={Song, Haoyi and Ji, Ruihan and Shi, Naichen and Lai, Fan and Kontar, Raed Al},
  journal={arXiv preprint arXiv:2506.09684},
  year={2025}
}

@article{nikitin2024kernel,
  title={Kernel language entropy: Fine-grained uncertainty quantification for llms from semantic similarities},
  author={Nikitin, Alexander and Kossen, Jannik and Gal, Yarin and Marttinen, Pekka},
  journal={Advances in Neural Information Processing Systems},
  volume={37},
  pages={8901--8929},
  year={2024}
}

@article{da2024llm,
  title={Llm uncertainty quantification through directional entailment graph and claim level response augmentation},
  author={Da, Longchao and Chen, Tiejin and Cheng, Lu and Wei, Hua},
  journal={arXiv preprint arXiv:2407.00994},
  year={2024}
}

@article{chen2023quantifying,
  title={Quantifying uncertainty in answers from any language model and enhancing their trustworthiness},
  author={Chen, Jiuhai and Mueller, Jonas},
  journal={arXiv preprint arXiv:2308.16175},
  year={2023}
}

@article{hou2023decomposing,
  title={Decomposing uncertainty for large language models through input clarification ensembling},
  author={Hou, Bairu and Liu, Yujian and Qian, Kaizhi and Andreas, Jacob and Chang, Shiyu and Zhang, Yang},
  journal={arXiv preprint arXiv:2311.08718},
  year={2023}
}

@article{kirchhof2025position,
  title={Position: Uncertainty quantification needs reassessment for large-language model agents},
  author={Kirchhof, Michael and Kasneci, Gjergji and Kasneci, Enkelejda},
  journal={arXiv preprint arXiv:2505.22655},
  year={2025}
}

@inproceedings{liu2025uncertainty,
  title={Uncertainty quantification and confidence calibration in large language models: A survey},
  author={Liu, Xiaoou and Chen, Tiejin and Da, Longchao and Chen, Chacha and Lin, Zhen and Wei, Hua},
  booktitle={Proceedings of the 31st ACM SIGKDD Conference on Knowledge Discovery and Data Mining V. 2},
  pages={6107--6117},
  year={2025}
}

@inproceedings{he2024llmelog,
  title={Llmelog: An approach for anomaly detection based on llm-enriched log events},
  author={He, Minghua and Jia, Tong and Duan, Chiming and Cai, Huaqian and Li, Ying and Huang, Gang},
  booktitle={2024 IEEE 35th International Symposium on Software Reliability Engineering (ISSRE)},
  pages={132--143},
  year={2024},
  organization={IEEE}
}

@article{zhang2024logicode,
  title={Logicode: an llm-driven framework for logical anomaly detection},
  author={Zhang, Yiheng and Cao, Yunkang and Xu, Xiaohao and Shen, Weiming},
  journal={IEEE Transactions on Automation Science and Engineering},
  year={2024},
  publisher={IEEE}
}

@article{park2024enhancing,
  title={Enhancing anomaly detection in financial markets with an llm-based multi-agent framework},
  author={Park, Taejin},
  journal={arXiv preprint arXiv:2403.19735},
  year={2024}
}

@article{gao2025vagu,
  title={VAGU \& GtS: LLM-Based Benchmark and Framework for Joint Video Anomaly Grounding and Understanding},
  author={Gao, Shibo and Yang, Peipei and Liu, Yangyang and Chen, Yi and Zhu, Han and Zhang, Xuyao and Huang, Linlin},
  journal={arXiv preprint arXiv:2507.21507},
  year={2025}
}

@article{wang2024visiongpt,
  title={Visiongpt: Llm-assisted real-time anomaly detection for safe visual navigation},
  author={Wang, Hao and Qin, Jiayou and Bastola, Ashish and Chen, Xiwen and Suchanek, John and Gong, Zihao and Razi, Abolfazl},
  journal={arXiv preprint arXiv:2403.12415},
  year={2024}
}

@inproceedings{yang2024follow,
  title={Follow the rules: Reasoning for video anomaly detection with large language models},
  author={Yang, Yuchen and Lee, Kwonjoon and Dariush, Behzad and Cao, Yinzhi and Lo, Shao-Yuan},
  booktitle={European Conference on Computer Vision},
  pages={304--322},
  year={2024},
  organization={Springer}
}

@inproceedings{duan2025home,
  title={A Home Broadband Maintenance and Installation Solution Leveraging LLM-Agent Technology},
  author={Duan, Hanting and Zhang, Jun and Zhang, Le and Wu, Yanqin and Lv, Tiantian and Zeng, Yu and Cheng, Xinna},
  booktitle={2025 IEEE 8th Information Technology and Mechatronics Engineering Conference (ITOEC)},
  volume={8},
  pages={1--6},
  year={2025},
  organization={IEEE}
}

@inproceedings{
bhat2025llm,
title={{LLM} Agents for Internet of Things (IoT) Applications},
author={Bhat, Akshat and Mondal, Aishee and Tripathy, Aniket},
booktitle={Submitted to CS598 LLM Agent 2025 Workshop},
year={2025},
url={https://openreview.net/forum?id=BikB3f8ByV},
note={under review}
}

@article{rivkin2024aiot,
  title={Aiot smart home via autonomous llm agents},
  author={Rivkin, Dmitriy and Hogan, Francois and Feriani, Amal and Konar, Abhisek and Sigal, Adam and Liu, Xue and Dudek, Gregory},
  journal={IEEE Internet of Things Journal},
  year={2024},
  publisher={IEEE}
}

@article{yang2024fasteval,
  title={FastEval Parkinsonism: an instant deep learning--assisted video-based online system for Parkinsonian motor symptom evaluation},
  author={Yang, Yu-Yuan and Ho, Ming-Yang and Tai, Chung-Hwei and Wu, Ruey-Meei and Kuo, Ming-Che and Tseng, Yufeng Jane},
  journal={Npj Digital Medicine},
  volume={7},
  number={1},
  pages={31},
  year={2024},
  publisher={Nature Publishing Group UK London}
}

@article{tian2024benefits,
  title={Benefits and barriers associated with the use of smart home health technologies in the care of older persons: a systematic review},
  author={Tian, Yi Jiao and Felber, Nadine Andrea and Pageau, F{\'e}lix and Schwab, Delphine Roulet and Wangmo, Tenzin},
  journal={BMC geriatrics},
  volume={24},
  number={1},
  pages={152},
  year={2024},
  publisher={Springer}
}

@article{lopes2024covis,
  title={CoViS: A contactless health monitoring system for the nursing home},
  author={Lopes, S{\'e}rgio Ivan and Silva, F{\'a}bio and Pinho, Pedro and Marques, Paulo and Abreu, Carlos and Milheiro, Jo{\~a}o and Braga, Bruno and Queir{\'o}s, Gabriel and Almeida, Rita and Carvalho, Nuno Borges},
  journal={IEEE Access},
  volume={12},
  pages={20802--20821},
  year={2024},
  publisher={IEEE}
}

@article{adhikari2024recent,
  title={Recent advances in anomaly detection in Internet of Things: Status, challenges, and perspectives},
  author={Adhikari, Deepak and Jiang, Wei and Zhan, Jinyu and Rawat, Danda B and Bhattarai, Asmita},
  journal={Computer Science Review},
  volume={54},
  pages={100665},
  year={2024},
  publisher={Elsevier}
}

@article{pang2021deep,
  title={Deep learning for anomaly detection: A review},
  author={Pang, Guansong and Shen, Chunhua and Cao, Longbing and Hengel, Anton Van Den},
  journal={ACM computing surveys (CSUR)},
  volume={54},
  number={2},
  pages={1--38},
  year={2021},
  publisher={ACM New York, NY, USA}
}

@article{kiran2018overview,
  title={An overview of deep learning based methods for unsupervised and semi-supervised anomaly detection in videos},
  author={Kiran, B Ravi and Thomas, Dilip Mathew and Parakkal, Ranjith},
  journal={Journal of Imaging},
  volume={4},
  number={2},
  pages={36},
  year={2018},
  publisher={MDPI}
}

@article{abshari2024llm,
  title={Llm-assisted physical invariant extraction for cyber-physical systems anomaly detection},
  author={Abshari, Danial and Fu, Chenglong and Sridhar, Meera},
  journal={arXiv preprint arXiv:2411.10918},
  volume={1},
  year={2024}
}

@article{nayak2021comprehensive,
  title={A comprehensive review on deep learning-based methods for video anomaly detection},
  author={Nayak, Rashmiranjan and Pati, Umesh Chandra and Das, Santos Kumar},
  journal={Image and Vision Computing},
  volume={106},
  pages={104078},
  year={2021},
  publisher={Elsevier}
}

@inproceedings{zanella2024harnessing,
  title={Harnessing Large Language Models for Training-free Video Anomaly Detection},
  author={Zanella, Luca and Menapace, Willi and Mancini, Massimiliano and Wang, Yiming and Ricci, Elisa},
  booktitle={Proceedings of the IEEE/CVF Conference on Computer Vision and Pattern Recognition},
  pages={18527--18536},
  year={2024}
}

@inproceedings{zaheer2022generative,
  title={Generative cooperative learning for unsupervised video anomaly detection},
  author={Zaheer, M Zaigham and Mahmood, Arif and Khan, M Haris and Segu, Mattia and Yu, Fisher and Lee, Seung-Ik},
  booktitle={Proceedings of the IEEE/CVF conference on computer vision and pattern recognition},
  pages={14744--14754},
  year={2022}
}

@inproceedings{lv2021learning,
  title={Learning normal dynamics in videos with meta prototype network},
  author={Lv, Hui and Chen, Chen and Cui, Zhen and Xu, Chunyan and Li, Yong and Yang, Jian},
  booktitle={Proceedings of the IEEE/CVF conference on computer vision and pattern recognition},
  pages={15425--15434},
  year={2021}
}

@article{wu2021learning,
  title={Learning causal temporal relation and feature discrimination for anomaly detection},
  author={Wu, Peng and Liu, Jing},
  journal={IEEE Transactions on Image Processing},
  volume={30},
  pages={3513--3527},
  year={2021},
  publisher={IEEE}
}

@inproceedings{tian2021weakly,
  title={Weakly-supervised video anomaly detection with robust temporal feature magnitude learning},
  author={Tian, Yu and Pang, Guansong and Chen, Yuanhong and Singh, Rajvinder and Verjans, Johan W and Carneiro, Gustavo},
  booktitle={Proceedings of the IEEE/CVF international conference on computer vision},
  pages={4975--4986},
  year={2021}
}

@inproceedings{li2022self,
  title={Self-training multi-sequence learning with transformer for weakly supervised video anomaly detection},
  author={Li, Shuo and Liu, Fang and Jiao, Licheng},
  booktitle={Proceedings of the AAAI Conference on Artificial Intelligence},
  volume={36},
  pages={1395--1403},
  year={2022}
}

@inproceedings{gong2019memorizing,
  title={Memorizing normality to detect anomaly: Memory-augmented deep autoencoder for unsupervised anomaly detection},
  author={Gong, Dong and Liu, Lingqiao and Le, Vuong and Saha, Budhaditya and Mansour, Moussa Reda and Venkatesh, Svetha and Hengel, Anton van den},
  booktitle={Proceedings of the IEEE/CVF international conference on computer vision},
  pages={1705--1714},
  year={2019}
}

@article{liu2021privacy,
  title={Privacy-preserving video fall detection using visual shielding information},
  author={Liu, Jixin and Xia, Yinyun and Tang, Zheng},
  journal={The Visual Computer},
  volume={37},
  number={2},
  pages={359--370},
  year={2021},
  publisher={Springer}
}

@article{withanage2016fall,
  title={Fall recovery subactivity recognition with RGB-D cameras},
  author={Withanage, Kalana Ishara and Lee, Ivan and Brinkworth, Russell and Mackintosh, Shylie and Thewlis, Dominic},
  journal={IEEE transactions on industrial informatics},
  volume={12},
  number={6},
  pages={2312--2320},
  year={2016},
  publisher={IEEE}
}

@inproceedings{ntelopoulos2024callm,
  title={CALLM: Cascading Autoencoder and Large Language Model for Video Anomaly Detection},
  author={Ntelopoulos, Apostolos and Nasrollahi, Kamal},
  booktitle={International Conference on Image Processing Theory, Tools and Applications},
  year={2024},
  organization={IEEE}
}

@article{ali2023real,
  title={Real-time video anomaly detection for smart surveillance},
  author={Ali, Manal Mostafa},
  journal={IET Image Processing},
  volume={17},
  number={5},
  pages={1375--1388},
  year={2023},
  publisher={Wiley Online Library}
}

@inproceedings{markovitz2020graph,
  title={Graph embedded pose clustering for anomaly detection},
  author={Markovitz, Amir and Sharir, Gilad and Friedman, Itamar and Zelnik-Manor, Lihi and Avidan, Shai},
  booktitle={Proceedings of the IEEE/CVF Conference on Computer Vision and Pattern Recognition},
  pages={10539--10547},
  year={2020}
}

@article{yahaya2021towards,
  title={Towards a data-driven adaptive anomaly detection system for human activity},
  author={Yahaya, Salisu Wada and Lotfi, Ahmad and Mahmud, Mufti},
  journal={Pattern Recognition Letters},
  volume={145},
  pages={200--207},
  year={2021},
  publisher={Elsevier}
}

@incollection{zhu2021video,
  title={Video anomaly detection for smart surveillance},
  author={Zhu, Sijie and Chen, Chen and Sultani, Waqas},
  booktitle={Computer Vision: A Reference Guide},
  pages={1315--1322},
  year={2021},
  publisher={Springer}
}

@inproceedings{ren2021deep,
  title={Deep video anomaly detection: Opportunities and challenges},
  author={Ren, Jing and Xia, Feng and Liu, Yemeng and Lee, Ivan},
  booktitle={2021 international conference on data mining workshops (ICDMW)},
  pages={959--966},
  year={2021},
  organization={IEEE}
}

@article{franc2023optimal,
  title={Optimal strategies for reject option classifiers},
  author={Franc, Vojtech and Prusa, Daniel and Voracek, Vaclav},
  journal={Journal of Machine Learning Research},
  volume={24},
  number={11},
  pages={1--49},
  year={2023}
}

@article{chow2003optimum,
  title={On optimum recognition error and reject tradeoff},
  author={Chow, C},
  journal={IEEE Transactions on information theory},
  volume={16},
  number={1},
  pages={41--46},
  year={2003},
  publisher={IEEE}
}

@article{lin2018selective,
  title={Selective sensing of a heterogeneous population of units with dynamic health conditions},
  author={Lin, Ying and Liu, Shan and Huang, Shuai},
  journal={IISE Transactions},
  volume={50},
  number={12},
  pages={1076--1088},
  year={2018},
  publisher={Taylor \& Francis}
}

@article{lingenerating,
  title={Generating with Confidence: Uncertainty Quantification for Black-box Large Language Models},
  author={Lin, Zhen and Trivedi, Shubhendu and Sun, Jimeng},
  journal={Transactions on Machine Learning Research},
  year={2024}
}

@article{liu2012robust,
  title={Robust recovery of subspace structures by low-rank representation},
  author={Liu, Guangcan and Lin, Zhouchen and Yan, Shuicheng and Sun, Ju and Yu, Yong and Ma, Yi},
  journal={IEEE transactions on pattern analysis and machine intelligence},
  volume={35},
  number={1},
  pages={171--184},
  year={2012},
  publisher={IEEE}
}

@article{chen2025uncertainty,
  title={Uncertainty Quantification of Large Language Models through Multi-Dimensional Responses},
  author={Chen, Tiejin and Liu, Xiaoou and Da, Longchao and Chen, Jia and Papalexakis, Vagelis and Wei, Hua},
  journal={arXiv preprint arXiv:2502.16820},
  year={2025}
}

%\bibliographystyle{informs2014} % outcomment this and next line in Case 1
%\bibliography{sample} % if more than one, comma separated

% CASE 2: BiBTeX used to generate mypaper.bbl (to be further fine tuned)
%\input{mypaper.bbl} % outcomment this line in Case 2

%If you don't use BiBTex, you can manually itemize references as shown below.

%\bibliographystyle{nonumber}

% \begin{thebibliography}{3}
% \providecommand{\natexlab}[1]{#1}
% \providecommand{\url}[1]{\texttt{#1}}
% \providecommand{\urlprefix}{URL }

% \bibitem[{Smith(2005)}]{smith2005}
% Smith J (2005) Optimal resource allocation in humanitarian logistics.
%   \emph{Journal of Operations Research} 30(2):123--135.
  
% \bibitem[{Jones(2010)}]{jones2010}
% Jones S (2010) Stochastic programming models for humanitarian logistics.
%   \emph{INFORMS Mathematics of Operations Research} 35(4):567--580.

% \bibitem[{Brown(2015)}]{brown2015}
% Brown D (2015) \emph{Introduction to Stochastic Programming} (Springer).

% \end{thebibliography}
\clearpage
\appendix

% \begin{APPENDIX}{} 
% \normalsize

\renewcommand{\thefigure}{S\arabic{figure}}
\setcounter{figure}{0} % Reset figure numbering
\setcounter{page}{1} % Reset figure numbering

\section{Theoretical proofs}

\subsection{Proof of Lemma 1}\label{proof_lemma1}
\begin{proof}
By noting $\boldsymbol{1}_{S(\boldsymbol{q})> \tau}=1-\boldsymbol{1}_{S(\boldsymbol{q})\leq \tau}$, we can observe that
\begin{align*}
\ell_g(\boldsymbol{q},y)&= \boldsymbol{1}_{f(\boldsymbol{q})\neq y}\boldsymbol{1}_{S(\boldsymbol{q})\leq \tau}+\boldsymbol{1}_{f^*(\boldsymbol{q})\neq y}\boldsymbol{1}_{S(\boldsymbol{q})>\tau}\\
&=\{\boldsymbol{1}_{f(\boldsymbol{q})\neq y}-\boldsymbol{1}_{f^*(\boldsymbol{q})\neq y}\}\boldsymbol{1}_{S(\boldsymbol{q})\leq \tau}+\boldsymbol{1}_{f^*(\boldsymbol{q})\neq y}   
\end{align*}   
It is easy to see that $\boldsymbol{1}_{S(\boldsymbol{q})\leq \tau}$ is a step function, which is non-increasing in $S(\boldsymbol{q})$. Thus, when $\boldsymbol{1}_{f(\boldsymbol{q})\neq y}>\boldsymbol{1}_{f^*(\boldsymbol{q})\neq y}$, $\ell_g(\boldsymbol{q},y)$ is monotonically non-decreasing in $\boldsymbol{1}_{S(\boldsymbol{q})\leq \tau}$; otherwise, it is monotonically non-increasing in $\boldsymbol{1}_{S(\boldsymbol{q})\leq \tau}$. By the composition rule, $\ell_g(\boldsymbol{q},y)$ is monotonic in $S(\boldsymbol{q})$.
\end{proof}

\subsection{Proof of Theorem 1}\label{proof_theorem1}
\begin{proof}
For the random strategy, since $S_r(\boldsymbol{q})\sim \text{Unif}(0,b)$, we have 
\begin{align*}
P(S_r(q)\leq \tau)=\frac{\tau}{b},    
\end{align*}
 which gives
 \begin{align*}
R_r&= \mathbb{E}_{(\boldsymbol{q},y)\sim\mathcal{F}}[\boldsymbol{1}_{f(\boldsymbol{q})\neq y}\boldsymbol{1}_{S_r(\boldsymbol{q})\leq \tau}] -\delta P(S_r(\boldsymbol{q})\leq \tau) +\delta\\
&=\mathbb{E}_{(\boldsymbol{q},y)\sim\mathcal{F}}[\boldsymbol{1}_{f(\boldsymbol{q})\neq y}\boldsymbol{1}_{S_r(\boldsymbol{q})\leq \tau}]+\frac{b-\tau}{b}\delta.
 \end{align*}
In random strategy, since $S_r(\boldsymbol{q})$ and $f(\boldsymbol{q})$ are independent, we have
\begin{align*}
R_r&=\mathbb{E}_{(\boldsymbol{q},y)\sim\mathcal{F}}[\boldsymbol{1}_{f(\boldsymbol{q})\neq y}]P(S_r(\boldsymbol{q})\leq \tau)+\frac{b-\tau}{b}\delta  \\
&=\frac{\tau}{b}\mathbb{E}_{(\boldsymbol{q},y)\sim\mathcal{F}}[\boldsymbol{1}_{f(\boldsymbol{q})\neq y}]+\frac{b-\tau}{b}\delta. 
\end{align*}
Then we have
\begin{align*}
\mathcal{R}_g-\mathcal{R}_r=& \mathbb{E}_{(\boldsymbol{q},y)\sim\mathcal{F}}[\boldsymbol{1}_{f(\boldsymbol{q})\neq y}(\boldsymbol{1}_{S(\boldsymbol{q})\leq \tau}-\frac{\tau}{b})]-\delta [P(S(\boldsymbol{q})\leq \tau)-\frac{\tau}{b}] \\
=&\mathbb{E}_{(\boldsymbol{q},y)\sim\mathcal{F}}[(\boldsymbol{1}_{f(\boldsymbol{q})\neq y}-\delta)(\boldsymbol{1}_{S(\boldsymbol{q})\leq \tau}-\frac{\tau}{b})]\\
=&\mathbb{E}_{\boldsymbol{q}}[\mathbb{E}_{y}[\boldsymbol{1}_{f(\boldsymbol{q})\neq y}-\delta|\boldsymbol{q}](\boldsymbol{1}_{S(\boldsymbol{q})\leq \tau}-\frac{\tau}{b})]\\
=&\mathbb{E}_{\boldsymbol{q}}[P(f(\boldsymbol{q})\neq y|\boldsymbol{q})-\delta)(\boldsymbol{1}_{S(\boldsymbol{q})\leq \tau}-\frac{\tau}{b})].
\end{align*}
% \end{proof}
Let $\tilde{p}(\boldsymbol{q})\triangleq P(f(\boldsymbol{q})\neq y|\boldsymbol{q})$, $\tilde{\delta}\triangleq\mathbb{E}_{\boldsymbol{q}}[P(f(\boldsymbol{q})\neq y|\boldsymbol{q})]$, and $\overline{s}=\mathbb{E}_{\boldsymbol{q}}[\boldsymbol{1}_{S(\boldsymbol{q})\leq\tau}]$. The risk gap can be rewritten as
\begin{align*}
 \mathcal{R}_g-\mathcal{R}_r=&\mathbb{E}_{\boldsymbol{q}}[(\tilde{p}(\boldsymbol{q})-\delta)(\boldsymbol{1}_{S(\boldsymbol{q})\leq \tau}-\frac{\tau}{b})]    \\
 =&\mathbb{E}_{\boldsymbol{q}}[(\tilde{p}(\boldsymbol{q})-\tilde{\delta}+\tilde{\delta}-\delta)(\boldsymbol{1}_{S(\boldsymbol{q})\leq \tau}-\frac{\tau}{b})] \\
 =&\mathbb{E}_{\boldsymbol{q}}[(\tilde{p}(\boldsymbol{q})-\tilde{\delta})(\boldsymbol{1}_{S(\boldsymbol{q})\leq \tau}-\overline{s})]+(\tilde{\delta}-\delta)(\overline{s}-\frac{\tau}{b}).
\end{align*}
Given that human experts perform better than the LLMs, i.e., $\tilde{\delta}\geq \delta$, and we can always adjust the parameters (e.g., $\tau$) to ensure the same coverage, we can ignore the second term. Thus, we have
\begin{align*}
\mathcal{R}_g-\mathcal{R}_r=&\mathbb{E}_{\boldsymbol{q}}[(\tilde{p}(\boldsymbol{q})-\tilde{\delta})(\boldsymbol{1}_{S(\boldsymbol{q})\leq \tau}-\overline{s})]\\
=&\text{Cov}(\tilde{p}(\boldsymbol{q}),\boldsymbol{1}_{S(\boldsymbol{q})\leq \tau}).
\end{align*}
We can see that when $\boldsymbol{1}_{S(\boldsymbol{q})\leq \tau}$ is negatively correlated with the wrong detection rates $\tilde{p}(\boldsymbol{q})$, we have $\mathcal{R}_g<\mathcal{R}_r$. The stronger this negative correlation, the greater the performance gain over random selection.   

\end{proof}

\subsection{Proof of Theorem 2}\label{proof_theorem2}
\begin{proof}
% First, by the law of total variance, we note that
% \begin{align}
% \hlm{\text{Var}[\boldsymbol{z}|\mathcal{T}]=
% \underbrace{\mathbb{E}_{\boldsymbol{x}}[\text{Var}[\boldsymbol{z}|\boldsymbol{x},\mathcal{T}]]}_{\substack{\text{expected variability}\\ \text{in analytic thinking}}}+\underbrace{\text{Var}_{\boldsymbol{x}}[\mathbb{E}[\boldsymbol{z}|\boldsymbol{x},\mathcal{T}]]}_{\text{variability from }\boldsymbol{x}}}
% \label{eq:z_T}
% \end{align}
While the second term in Eq. \eqref{eq:z_T} corresponds to the variability sourced from $\boldsymbol{x}$, we only concern about the first term which is the average variability in Analytic Thinking where the interferences of the uncertainty in $\boldsymbol{x}$ is averaged out. In other words, the first term gives us the estimate of the uncertainty inherent in Analytic Thinking regardless of the contribution of $\boldsymbol{x}$. Note that the variance $\text{Var}[\boldsymbol{z}|\boldsymbol{x},\mathcal{T}]$ can be further decomposed as
% \begin{align}
% \text{Var}[\boldsymbol{z}|\boldsymbol{x},\mathcal{T}]=\text{Var}_{\tilde{h}}[\mathbb{E}[\boldsymbol{z}|\boldsymbol{x},\mathcal{T},\tilde{h}]]+ \\
% \mathbb{E}_{\tilde{h}}[\text{Var}[\boldsymbol{z}|\boldsymbol{x},\mathcal{T},\tilde{h}]],
% \label{eq:z_x_T}
% \end{align}

\begin{align}
\text{Var}[\boldsymbol{z}|\boldsymbol{x},\mathcal{T}] 
&= \underbrace{\text{Var}_{\tilde{h}}\!\left[\mathbb{E}[\boldsymbol{z}|\boldsymbol{x},\mathcal{T},\tilde{h}]\right]}_{\text{variability in task-related reasoning}} + \underbrace{\mathbb{E}_{\tilde{h}}\!\left[\text{Var}[\boldsymbol{z}|\boldsymbol{x},\mathcal{T},\tilde{h}]\right]}_{\text{inherent variability in LLM}},
\label{eq:z_x_T}
\end{align}
\noindent where the two terms in Eq. \eqref{eq:z_x_T} represent the task-related reasoning variability (i.e., the part that $S_{task}$ concerns with) and the inherent variability in LLM reasoning ability (i.e., randomness caused by the probabilistic nature of the LLMs), respectively. To understand the inherent variability in LLM reasoning ability, one can think of the temperature parameter in LLM that could be set to be a large number to solicit more such uncertainty so the output of the LLM can be more random given the same input. And vice verse, if we set the temperature parameter as 0, we expect the LLM to generate the same output given the same input. In our work, we do set the temperatures of the LLMs to 0 and assume their stability in decision-making, since this randomness relies on the model designs and is irrelevant to our purpose of computing the task specific uncertainty score $S_{task}$. 

%  which represents the expected information gain from making the hypothesis $\tilde{h}$ in reasoning

Summarizing the aforementioned analysis, to compute $S_{task}$, we only need to focus on the first term in Eq. \eqref{eq:z_x_T}. We compute $S_{task}$ using $\Delta R=\mathbb{E}_{\boldsymbol{x}}\Bigl[\text{Var}_{\tilde{h}}[\mathbb{E} [\boldsymbol{z}|\boldsymbol{x},\mathcal{T},\tilde{h}]]\Bigr]$ which can be rewritten as
% \begin{align*}
% \Delta R=&\mathbb{E}_{\boldsymbol{x}}\Bigl[\mathbb{E}_{\tilde{h}}\bigl[\mathbb{E}[\boldsymbol{z}|\boldsymbol{x},\mathcal{T},\tilde{h}]-\mathbb{E}[\boldsymbol{z}|\boldsymbol{x},\mathcal{T}]\bigr]^2\Bigl]\\
% =&\mathbb{E}_{\tilde{h}}\Bigl[\mathbb{E}_{\boldsymbol{x}}\bigl[\mathbb{E}[\boldsymbol{z}|\boldsymbol{x},\mathcal{T},\tilde{h}]-\mathbb{E}[\boldsymbol{z}|\mathcal{T}] \\
% &+\mathbb{E}[\boldsymbol{z}|\mathcal{T}]-\mathbb{E}[\boldsymbol{z}|\boldsymbol{x},\mathcal{T}]\bigr]^2\Bigl]\\
% =&\mathbb{E}_{\tilde{h}}[R_{\tilde{h}}] - R. 
% \end{align*}
% \allowdisplaybreaks
\begin{align*}
\Delta R=&\mathbb{E}_{\boldsymbol{x}}\Bigl[\mathbb{E}_{\tilde{h}}\bigl[(\mathbb{E}[\boldsymbol{z}|\boldsymbol{x},\mathcal{T},\tilde{h}]-\mathbb{E}[\boldsymbol{z}|\boldsymbol{x},\mathcal{T}])^2\bigr]\Bigl]\\
=&\mathbb{E}_{\tilde{h}}\Bigl[\mathbb{E}_{\boldsymbol{x}}\bigl[\mathbb{E}[\boldsymbol{z}|\boldsymbol{x},\mathcal{T},\tilde{h}]-\mathbb{E}[\boldsymbol{z}|\mathcal{T}]+\mathbb{E}[\boldsymbol{z}|\mathcal{T}]-\mathbb{E}[\boldsymbol{z}|\boldsymbol{x},\mathcal{T}]\bigr]^2\Bigl].
\end{align*}
Let $A = \mathbb{E}[\boldsymbol{z}|\boldsymbol{x},\mathcal{T},\tilde{h}]-\mathbb{E}[\boldsymbol{z}|\mathcal{T}]$ and $B = \mathbb{E}[\boldsymbol{z}|\mathcal{T}]-\mathbb{E}[\boldsymbol{z}|\boldsymbol{x},\mathcal{T}]$. It is easy to see that
\begin{align*}
\Delta R=&\mathbb{E}_{\boldsymbol{x}}\Bigl[\mathbb{E}_{\tilde{h}}\bigl[A+B\bigr]^2\Bigl]=\mathbb{E}_{\tilde{h}}\Bigl[\mathbb{E}_{\boldsymbol{x}}\bigl[A +B\bigr]^2\Bigl]\\
=& \mathbb{E}_{\tilde{h}} \bigl[\mathbb{E}_{\boldsymbol{x}} [A^2]\bigl] + 2\mathbb{E}_{\boldsymbol{x}}\bigl[\mathbb{E}_{\tilde{h}}[AB]\bigl] + \mathbb{E}_{\boldsymbol{x}}[B^2] \\
=&\mathbb{E}_{\tilde{h}} \bigl[\mathbb{E}_{\boldsymbol{x}} [A^2]\bigl] -\mathbb{E}_{\boldsymbol{x}}[B^2]\\
=&\mathbb{E}_{\tilde{h}}[R_{\tilde{h}}] - R.
\end{align*}
% .
% A = \mathbb{E}[\boldsymbol{z}|\boldsymbol{x},\mathcal{T},\tilde{h}]-\mathbb{E}[\boldsymbol{z}|\mathcal{T}]
% B = \mathbb{E}[\boldsymbol{z}|\mathcal{T}]-\mathbb{E}[\boldsymbol{z}|\boldsymbol{x},\mathcal{T}

% \begin{align*}
% \Delta R=&\mathbb{E}_{\boldsymbol{x}}\Bigl[\mathbb{E}_{\tilde{h}}\bigl[(\mathbb{E}[\boldsymbol{z}|\boldsymbol{x},\mathcal{T},\tilde{h}]-\mathbb{E}[\boldsymbol{z}|\boldsymbol{x},\mathcal{T}])^2\bigr]\Bigl]\\
% =&\mathbb{E}_{\tilde{h}}\Bigl[\mathbb{E}_{\boldsymbol{x}}\bigl[\mathbb{E}[\boldsymbol{z}|\boldsymbol{x},\mathcal{T},\tilde{h}]-\mathbb{E}[\boldsymbol{z}|\mathcal{T}]\\
% &+\mathbb{E}[\boldsymbol{z}|\mathcal{T}]-\mathbb{E}[\boldsymbol{z}|\boldsymbol{x},\mathcal{T}]\bigr]^2\Bigl]\\
% =&\mathbb{E}_{\tilde{h}}[R_{\tilde{h}}] - R
% \end{align*}
Here, $R_{\tilde{h}}=\mathbb{E}_{\boldsymbol{x}} [A^2]=\mathbb{E}_{\boldsymbol{x}}\bigl[\mathbb{E}[\boldsymbol{z}|\boldsymbol{x},\mathcal{T},\tilde{h}]-\mathbb{E}[\boldsymbol{z}|\mathcal{T}]\bigr]^2$ and $R=\mathbb{E}_{\boldsymbol{x}} [B^2]=\mathbb{E}_{\boldsymbol{x}}\bigl[\mathbb{E}[\boldsymbol{z}|\boldsymbol{x},\mathcal{T}]-\mathbb{E}[\boldsymbol{z}|\mathcal{T}]\bigr]^2$. Thus, we can conclude that $S_{task}$ indicates the variations conditioned on $\boldsymbol{x}$, with and without the inclusion of the generated hypothesis $\tilde{h}$.
\end{proof}

\section{Additional Experiment Setup}\label{add_exp_setup}
For each case study, we split the dataset into training and testing sets at a 4:1 ratio, with stratification (i.e., based on video types and wound categories) to ensure a representative distribution (i.e., to ensure that the proportions of the video or wound types are the same between the training and testing datasets). All text components, e.g., data descriptions, reasoning, and side information, are embedded into vector representations using the Sentence-BERT encoder all-MiniLM-L6-v2 as the embedding function $e(\cdot)$. It produces 384-dimensional semantic embeddings optimized for measuring meaning similarity between sentences. For the PMF algorithms, we select the optimal latent dimension $K$ from the set $\{5,10,15\}$ using 5-fold cross-validation. In the Reflection stage, we use a standard logistic regression classifier to map the combined embeddings as features (i.e., the reasoning $\boldsymbol{z}$, the hypothesis $\tilde{h}$, and the side information $\boldsymbol{c}$) to anomaly probabilities. The classifier is implemented using sklearn \mbox{\citep{scikit-learn}} from Python with the LBFGS optimizer, a maximum of 1000 iterations, and an L2 regularization penalty. To reduce computational cost, we obtain the optimal weights $\boldsymbol{\alpha}^*$ by Eq. \mbox{\eqref{Kfold_opt_alpha}} through a grid search over the range [0,1] with a step size of 0.1, ensuring that the weights sum to 1.

\section{Justification for the Choice of Evaluation Metrics}
Our evaluation metrics are designed to comprehensively capture the performance and reliability of the proposed UQ decision system. They measure (1) Predictive accuracy under abstention. By evaluating overall accuracy as a function of rejection rate, the observed improvement in Figures 5(a) and 11 verifies that the combined UQ score behaves consistently with theoretical expectations. Recall evaluates the sensitivity to abnormal cases, which is essential for safety-critical applications, where missing true anomalies can lead to higher downstream cost. F1 score balances precision and recall, providing a robust summary when class distributions are imbalanced. Accuracy on $D_{Ambiguity}$ shows how the methods perform on  the most contextually ambiguous and hard-to-classify cases. (2) UQ. The ratio of true misclassification in rejected cases measures the precision of abstention, i.e., the proportion of rejected instances that are indeed misclassified by the LLM ensemble. A higher ratio empirically confirms that the learned UQ scores correctly identifies samples that the LLMs cannot make accurate predictions. (3) Robustness of the UQ methods. Overall accuracy using optimal and smooth weights empirically tests the robustness and stability of the UQ integration. Small performance differences confirm that the weighting scheme generalizes well and is not overfitted to noise. (4) Cost-aware decision efficiency. The trade-off curve (Figure \mbox{\ref{fig:P_cost_both}}) between optimal P values and different unit cost demonstrates that ALARM achieves accuracy-cost balance.

\section{Additional Results in Case Studies}
\subsection{VAD in Smart Homes}
\begin{figure}
\centering
\includegraphics[width=0.7\textwidth]{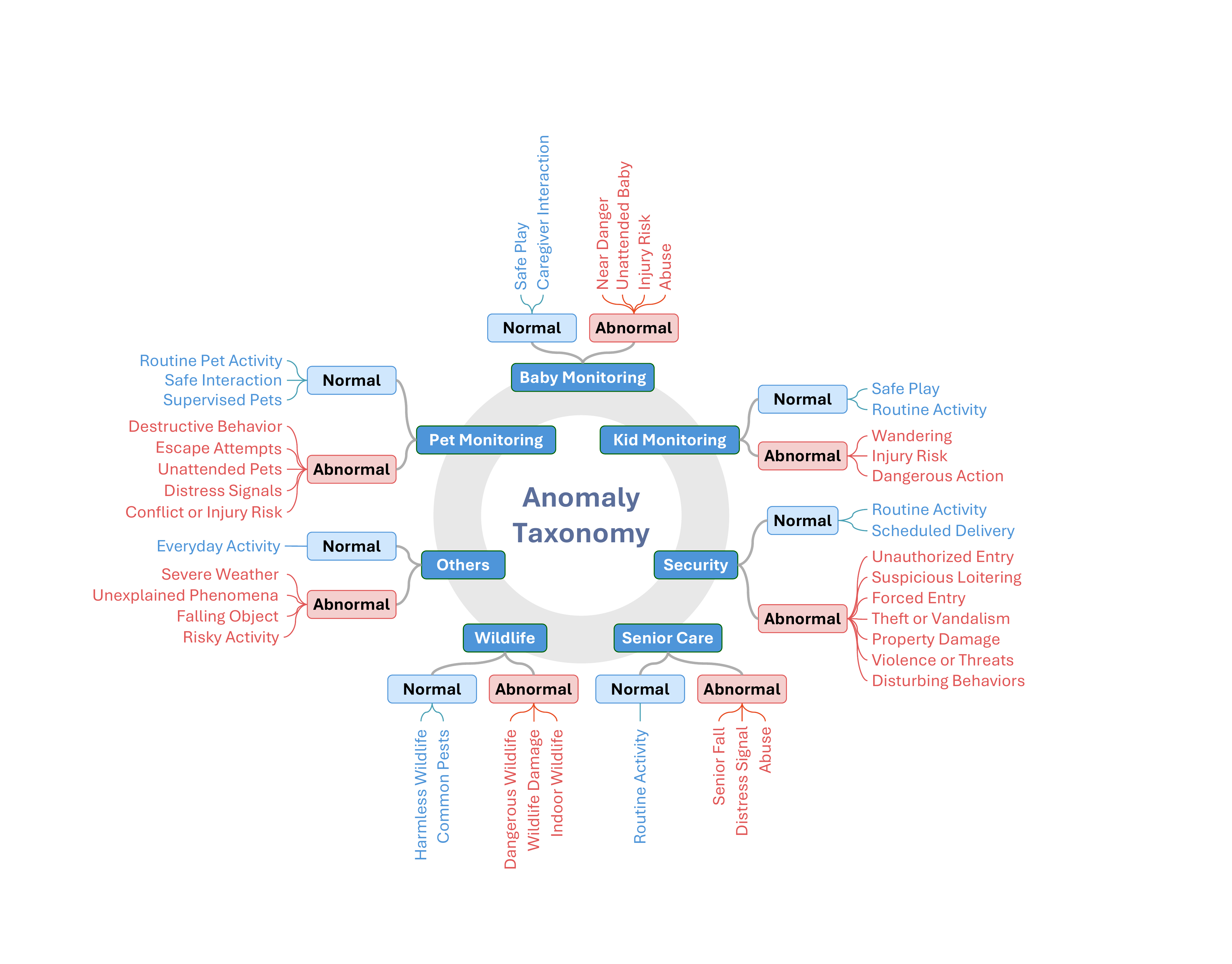}
\caption{Overview of the video anomaly taxonomy in smart homes.}\label{fig:taxonomy}
%\vspace{-1.5em}
\end{figure}

In the Reflection stage of the LLM reasoning chain, we incorporate anomaly rules as side information $\boldsymbol{c}$ based on an anomaly taxonomy shown in Figure \ref{fig:taxonomy} developed by domain experts. 

\begin{figure}
\centering
\includegraphics[width=0.5\textwidth]{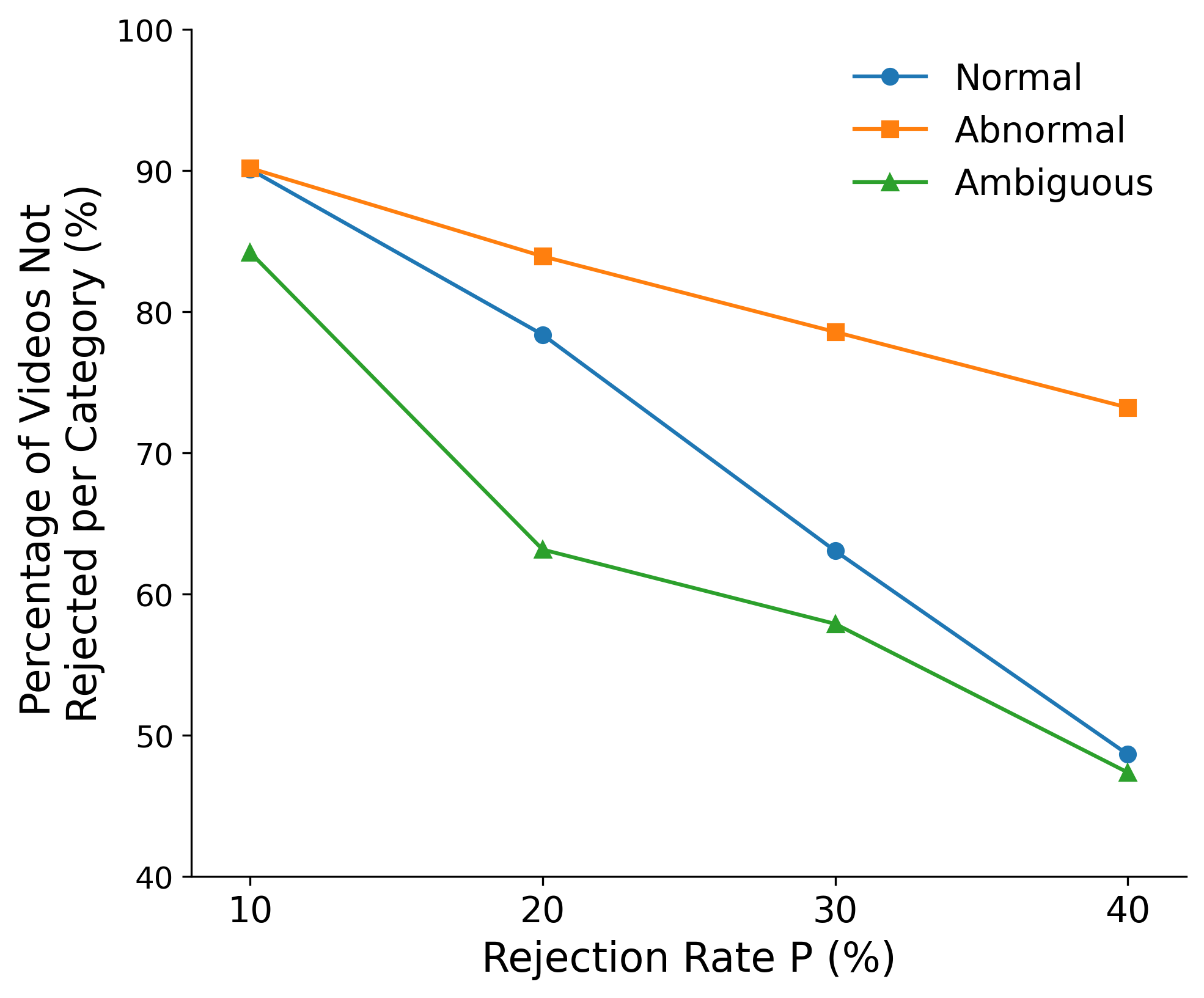}
%\vspace{-4mm}
\caption{The percentage of videos not rejected, i.e., videos identified as low-uncertainty by our ALARM UQ score, across different ground-truth anomaly categories under varying rejection rates $P$.}\label{fig:num_video_P}
%\vspace{-5mm}
\end{figure}  

\subsubsection{Which categories (normal, abnormal, and ambiguous) of cases are mostly impacted by our UQ score?}

It is of interest to examine the percentage of videos classified by MLLMs, meaning those identified as low uncertainty by our UQ score, in each of the three categories: normal, abnormal, and ambiguous (defined by ground truth). Online Figure \ref{fig:num_video_P} shows this information, while each curve corresponds to a category and shows how the percentage of low uncertainty videos changes according to the change of rejection rate $P$. Apparently, the category of abnormal videos is most ``resilient'' to the impact of $P$, indicating that if a video is anomalous, its meaning or signal is most certain. The category of ambiguous videos is the most impacted, which is consistent with the context, since this category contains the most ambiguous videos as determined by domain experts. Positioned between the ambiguous and abnormal categories lies the normal category, reflecting its intermediate level of uncertainty. This analysis not only confirms our guess but also sheds light on how UQ can really help to improve VAD in complex environments like smart homes. Some cases are apparently abnormal, like a senior fell down the floor or a cat threw up, but some cases are challenging since they could be normal in a different context, like the detection of a stranger (a friend or a thief?). These cases are conditionally normal and should be marked out by UQ and further evaluated by human stakeholders.

% Figure \ref{fig:num_video_P} shows the impact of $P$ on the percentage of videos not rejected across different ground-truth anomaly categories. Overall, the category of abnormal videos is most ``resilient'' to the impact of $P$, indicating that if a video is anomalous, its meaning or signal is most certain. The category of ambiguous videos is the most impacted, which is consistent with the context, since this category contains the most ambiguous videos as determined by domain experts. 

% \begin{figure}
% \centering
% \includegraphics[width=0.5\textwidth]{fig/percent_low.png}
% %\vspace{-4mm}
% \caption{The percentage of videos not rejected, i.e., videos identified as low-uncertainty by our ALARM UQ score, across different ground-truth anomaly categories under varying rejection rates $P$.}\label{fig:num_video_P}
% %\vspace{-5mm}
% \end{figure}  

\begin{figure}
\centering
\includegraphics[width=0.95\textwidth]{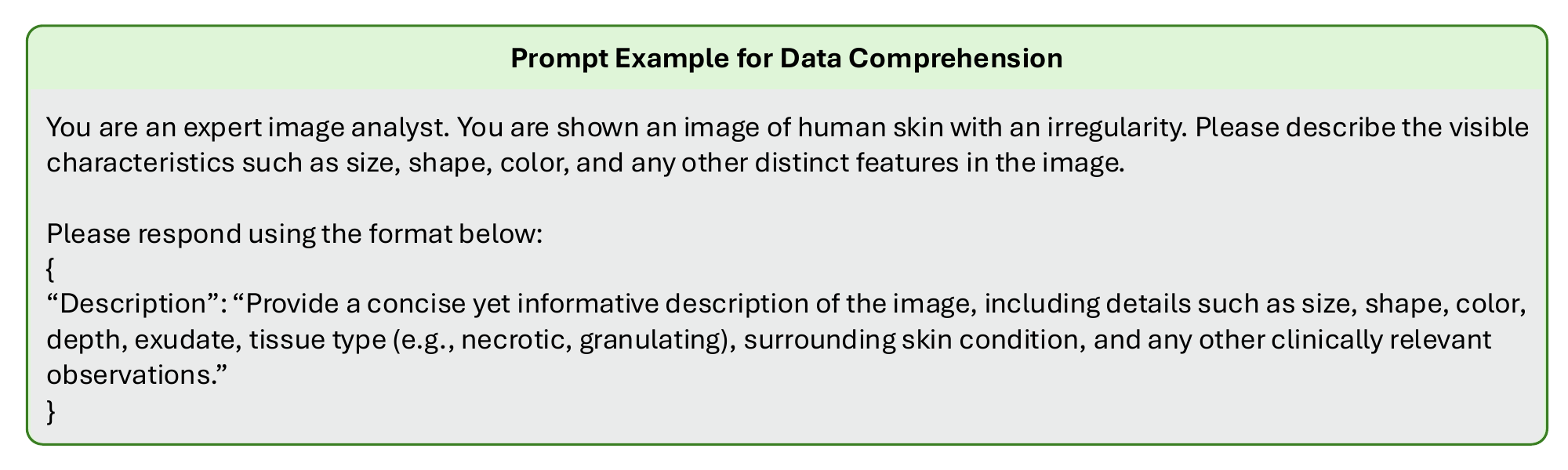}
%\vspace{-3mm}
\caption{Prompt example for Data Comprehension in wound classification.}\label{fig:prompt_data_wound}
%\vspace{-2mm}
\end{figure}

\begin{figure}
\centering
\includegraphics[width=\textwidth]{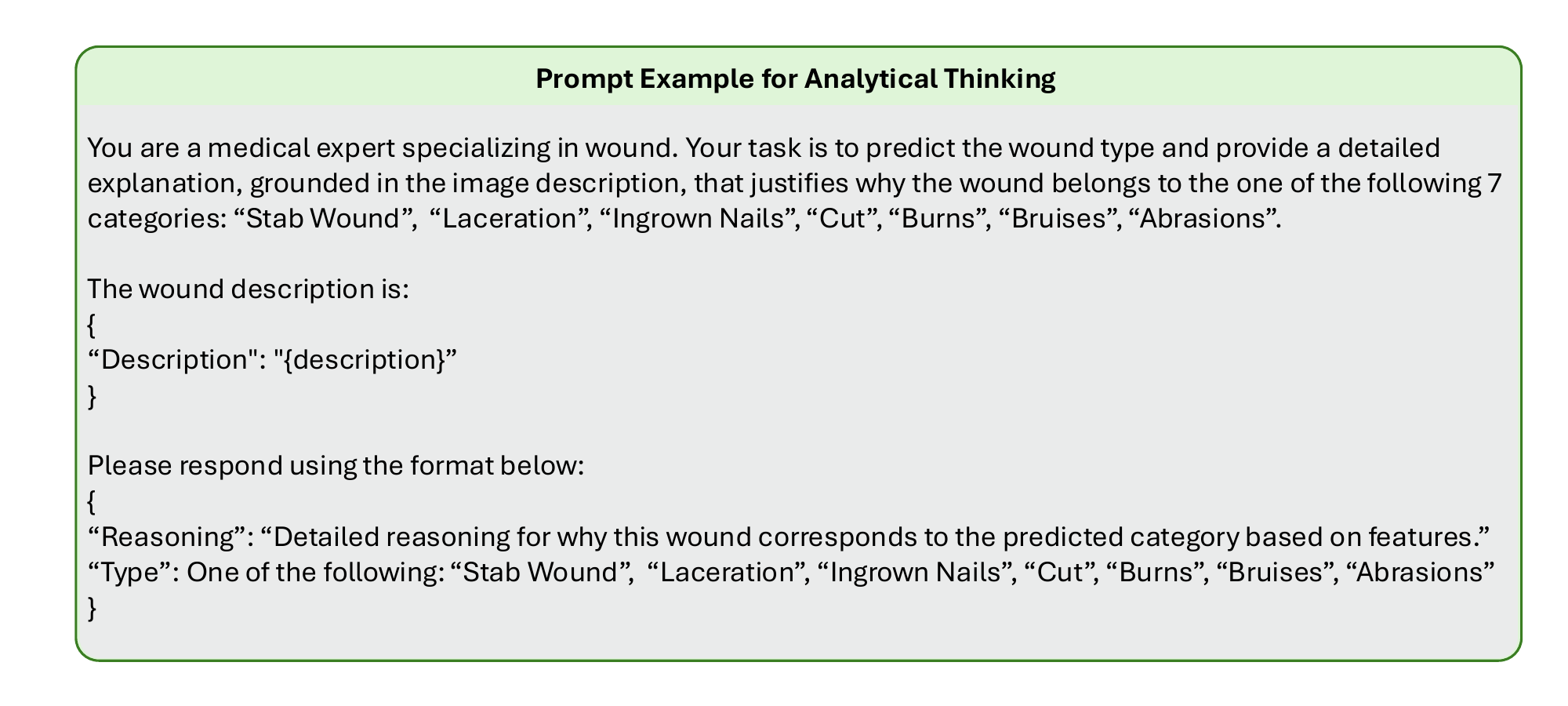}
%\vspace{-3mm}
\caption{Prompt example for Analytical Thinking in wound classification.}\label{fig:prompt_task_wound}
%\vspace{-2mm}
\end{figure}

\begin{figure}
\centering
\includegraphics[width=\textwidth]{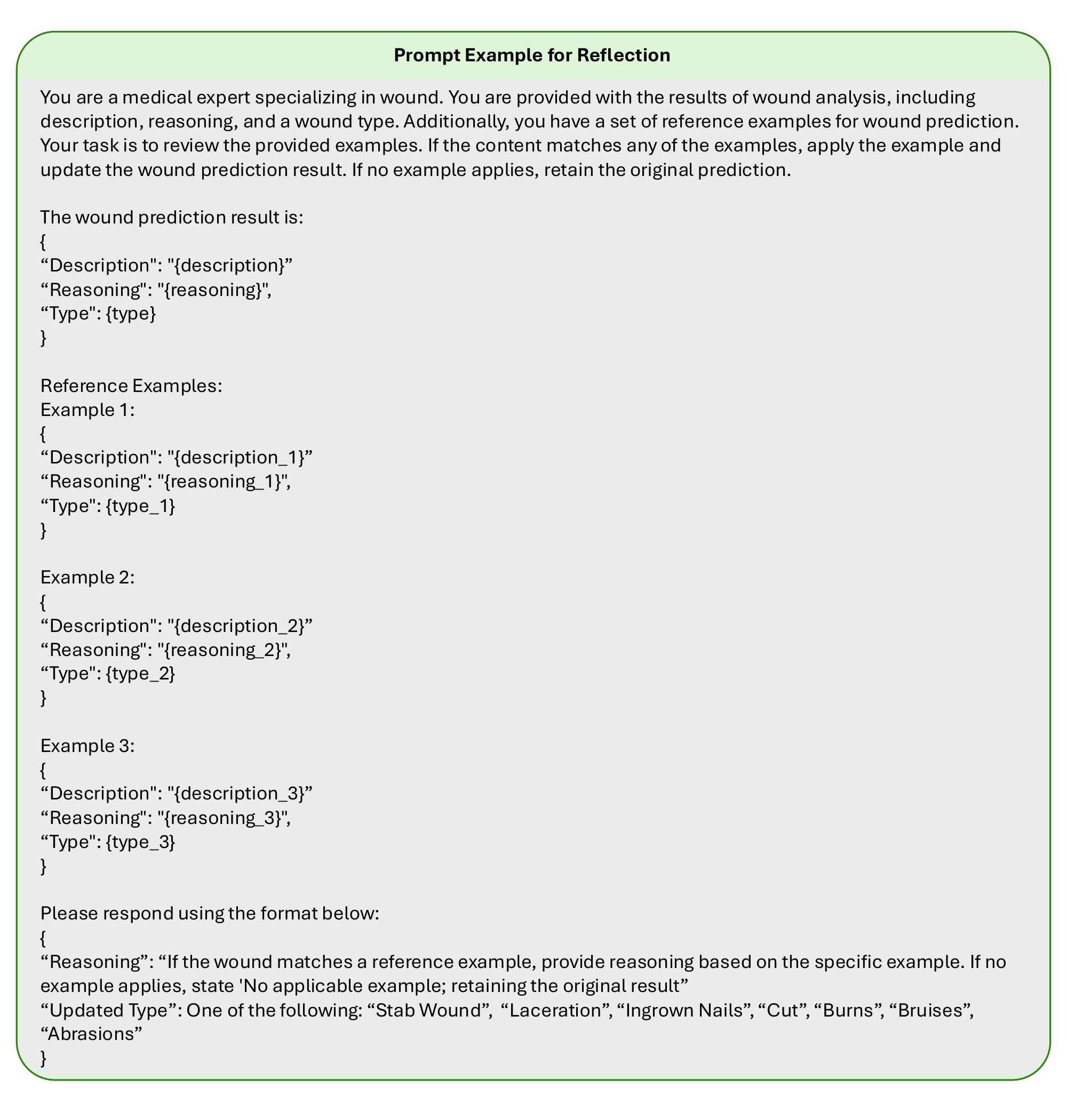}
%\vspace{-3mm}
\caption{Prompt example for Reflection in wound classification.}\label{fig:prompt_ref_wound}
%\vspace{-2mm}
\end{figure}  

\subsection{Wound Classification}

In Data Comprehension (prompt examples in Online Figure \ref{fig:prompt_data_wound}), MLLMs generate structured visual descriptions from wound images (e.g., color, shape, and size), capturing the essential observable features. MLLMs in Analytical Thinking (prompt examples in Online Figure \ref{fig:prompt_task_wound}) use these descriptions to predict the wound type and provide justifications grounded in visual features, producing an initial hypothesis (e.g., ``Bruise'' due to reddish-purplish discoloration and pooled blood beneath the skin). Reflection (prompt examples in Online Figure \ref{fig:prompt_ref_wound}) introduces domain-specific side information. Here, MLLMs use reference examples as side information with descriptions, reasoning, and known wound types to re-evaluate and, if needed, revise the initial hypothesis. In the example shown in Figure \ref{fig:overview_wound}, the comparison with an ``Abrasion'' reference led to a label change from bruise to abrasion. At each stage, ALARM computes stage-specific uncertainty scores $S_{data}$, $S_{task}$, and $S_{ref}$, which are optimally weighted ($\alpha_1, \alpha_2, \alpha_3$) to produce the final integrated uncertainty score $S$. This score determines whether a case is classified automatically via majority voting across MLLMs or deferred to human experts, enabling selective decision-making.

% \begin{figure}[H]
% \centering
% \includegraphics[width=\textwidth]{fig/wound_prompt_reas.pdf}
% %\vspace{-3mm}
% \caption{Prompt example for Analytical Thinking in wound classification.}\label{fig:prompt_task_wound}
% %\vspace{-2mm}
% \end{figure}  

% \begin{figure}[H]
% \centering
% \includegraphics[width=\textwidth]{fig/wound_prompt_ref.pdf}
% %\vspace{-3mm}
% \caption{Prompt example for Reflection in wound classification.}\label{fig:prompt_ref_wound}
% %\vspace{-2mm}
% \end{figure}  

% The prompts used in LLM reasoning chains are detailed as follows. Figure \ref{fig:prompt_data_wound} shows the prompt for Data Comprehension. We request LLMs to provide a description for the wound image. In Analytical Thinking, we prompt LLM as \ref{fig:prompt_task_wound} to use these descriptions to provide the reasoning and its prediction, i.e., hypothesis, on the wound type. Then shown as Figure \ref{fig:prompt_ref_wound}, we provide the LLM with some domain-specific side information, i.e., known wound description, reasoning and type, to reflect its previous results.

% \begin{figure}
% \centering
% \includegraphics[width=\textwidth]{fig/wound_prompt_ref.pdf}
% %\vspace{-3mm}
% \caption{Prompt example for Reflection in wound classification.}\label{fig:prompt_ref_wound}
% %\vspace{-2mm}
% \end{figure}  

% \end{APPENDIX}

%%%%%%%%%%%%%%%%%
\end{document}